\documentclass[manuscript,screen,authorversion,nonacm]{acmart}
\AtBeginDocument{%
  \providecommand\BibTeX{{%
    \normalfont B\kern-0.5em{\scshape i\kern-0.25em b}\kern-0.8em\TeX}}}

\usepackage{booktabs}						%
\usepackage{multirow}						%
\usepackage{amsfonts}						%
\usepackage{graphicx}						%
\usepackage{duckuments}						%

\usepackage[utf8]{inputenc} %
\usepackage[T1]{fontenc}    %
\usepackage{url}            %
\usepackage{booktabs}       %
\usepackage{amsfonts}       %
\usepackage{nicefrac}       %
\usepackage{microtype}      %
\usepackage{xcolor}         %
\usepackage{enumitem}
\usepackage{tikz}

\usepackage{graphics}
\usepackage{graphicx}
\usepackage{xcolor}
\usepackage{amsmath}
\usepackage{bbding}
\usepackage{enumitem}
\usepackage{wasysym}
\usepackage{mathtools}
\usepackage{booktabs}
\usepackage{subcaption}
\usepackage{array,multirow}
\usepackage{adjustbox}
\usepackage{longtable}
\usepackage{tikz-cd}
\usepackage{placeins}
\usepackage{float}
\usepackage{tablefootnote}
\usepackage{tocloft}

\usepackage{booktabs, makecell}
\usepackage{wrapfig,lipsum}

\usepackage{indentfirst}
\setlist[itemize]{leftmargin=5mm}

\everypar{\looseness=-1}

\newcommand{\survey}[1]{#1}
\newcommand{\answerYes}[1][]{\textcolor{blue}{[Yes] #1}}
\newcommand{\answerNo}[1][]{\textcolor{orange}{[No] #1}}
\newcommand{\answerNA}[1][]{\textcolor{gray}{[N/A] #1}}
\newcommand{\answerTODO}[1][]{\textcolor{red}{\bf [TODO]}}

\begin{document}

\title{Weisfeiler and Leman Go Measurement Modeling: \\
Probing the Validity of the WL Test}

\author{Arjun Subramonian}
\email{arjunsub@cs.ucla.edu}
\affiliation{%
  \institution{University of California, Los Angeles}
  \country{USA}
}

\author{Adina Williams}
\email{adinawilliams@meta.com}
\affiliation{%
  \institution{Meta AI}
  \country{USA}
}

\author{Maximilian Nickel}
\email{maxn@meta.com}
\affiliation{%
  \institution{Meta AI}
  \country{USA}
}

\author{Yizhou Sun}
\email{yzsun@cs.ucla.edu}
\affiliation{%
  \institution{University of California, Los Angeles}
  \country{USA}
}

\author{Levent Sagun}
\email{leventsagun@meta.com}
\affiliation{%
  \institution{Meta AI}
  \country{France}
}

\renewcommand{\shortauthors}{Subramonian, et al.}

\begin{abstract}
The expressive power of graph neural networks is usually measured by comparing how many pairs of graphs or nodes an architecture can possibly distinguish as non-isomorphic to those distinguishable by the $k$-dimensional Weisfeiler-Leman ($k$-WL) test. In this paper, we uncover misalignments between graph machine learning practitioners' conceptualizations of expressive power and $k$-WL through a systematic analysis of the reliability and validity of $k$-WL. We conduct a survey ($n = 18$) of practitioners to surface their conceptualizations of expressive power and their assumptions about $k$-WL. In contrast to practitioners' beliefs, our analysis (which draws from graph theory and benchmark auditing) reveals that $k$-WL does not guarantee isometry, can be irrelevant to real-world graph tasks, and may not promote generalization or trustworthiness. We argue for extensional definitions and measurement of expressive power based on benchmarks. We further contribute guiding questions for constructing such benchmarks, which is critical for graph machine learning practitioners to develop and transparently communicate our understandings of expressive power.
Our code can be found at: \url{https://github.com/ArjunSubramonian/wl-test-exploration}.
\end{abstract}



\maketitle

\section{Introduction}
Graph neural networks (GNNs) have been successfully applied to graph-structured data such as social networks, operations networks, and molecules \cite{Dwivedi2020BenchmarkingGN, Hu2020OpenGB} to aid in important tasks like content recommendation \cite{fan2019graph, Wu2020GraphNN}, routing \cite{Cappart2021CombinatorialOA}, and molecular property prediction \cite{zhang2021motifdriven, WIEDER20201}. However, researchers have raised concerns about the generalization limits of GNNs with respect to computing important graph properties (e.g., clique information), handling larger graphs, and counting substructures \cite{Garg2020Generalization, yehudai2021on, tahmasebi2022counting}. Moreover, GNNs exhibit serious fairness, privacy, and robustness concerns, including marginalizing stratified social groups and low-degree nodes \cite{Tang2020Degree, li2021on, Kang2022RawlsGCN, subramonian2023networked}, enabling the recovery of private links and node information \cite{Wu2021LINKTELLERRP, Duddu2021Leakage, HJBGZ21}, and being highly susceptible to adversarial node and link edits \cite{wang2019attack, ijcai2019p669, Wu2020Bottleneck}.

Towards improving GNN performance on various tasks, numerous GNN architectures have been proposed \cite{Kipf:2016tc, hamilton2017, Velickovic2018GraphAN, Zeng2020GraphSAINTGS}. In recent years, graph machine learning (ML) practitioners who contribute a new GNN architecture also often theoretically characterize its \textit{expressive power} \cite{xu2018how, Morris2019WeisfeilerAL, Maron2019InvariantAE, Li2020DistanceE, Bouritsas2022ImprovingGN, Kim2022PureTA, yao2023going}. Because expressive power is a theoretical construct that cannot be observed and measured directly, it must be inferred from measurements of observable properties. In the GNN literature, expressive power is usually measured by comparing how many pairs of graphs or nodes a GNN architecture can possibly distinguish as non-isomorphic to those distinguishable by the $k$-WL isomorphism test.
In particular, more distinguishable pairs of graphs or nodes indicates greater expressive power.

Poor expressive power has been postulated to limit model performance on real-world graph tasks \cite{xu2018how, Morris2019WeisfeilerAL};
however, GNN architectures that are proven to be more expressive often do not empirically improve performance on such tasks \cite{Morris2020TUDatasetAC, Zopf20221WLEI, fuchs2023universalityofneural}. Towards understanding this phenomenon, we adopt the measurement modeling framework from the social sciences \cite{Adcock2001MeasurementVA, jacobs2021}, which disentangles how graph ML practitioners \emph{conceptualize} expressive power from how we \emph{operationalize} its measurement.
Notably, $k$-WL operationalizes the measurement of a specific conceptualization of expressive power: the extent to which an architecture can possibly map non-isomorphic graphs or nodes to distinct representations \cite{xu2018how, Morris2019WeisfeilerAL}, or capture a mapping function that is injective up to isomorphism.
However, this could be misaligned with how graph ML practitioners conceptualize expressive power.
Even worse, practitioners' conceptualizations of expressive power may be unclear or inconsistent because of the unique requirements of different graph ML tasks. 

In this paper, we uncover misalignments between practitioners' conceptualizations of expressive power and $k$-WL (and the underlying assumptions that cause it) through a systematic analysis of the \textit{construct reliability} (can it be repeated?) and \textit{construct validity} (is it right?) of $k$-WL \cite{jacobs2021}. We conduct a survey ($n = \survey{18}$) of graph ML practitioners to surface their conceptualizations of expressive power and their assumptions about $k$-WL, which are often not stated in relevant literature.
In particular, our survey reveals that practitioners' conceptualizations of expressive power differ (e.g., some believe that expressive GNN architectures should induce isometry, while others do not). In addition, via our graph-theoretic analysis, we find that $k$-WL can be misaligned with accepted measurements for a task (e.g., graph edit distance), and does not guarantee isometry. Furthermore, we examine extrinsic limits of $k$-WL, such as decoder capacity and learning dynamics, and how $k$-WL can be antithetical to generalization.

Complementarily, our benchmark auditing reveals that: (1) 1-WL can distinguish effectively all the non-isomorphic graphs and nodes in many graph ML benchmarks; (2) 1-WL may not provide a useful upper bound on the accuracy of a GNN on every task \cite{Zopf20221WLEI}; and (3) \textbf{GNNs may learn representations that are more optimal with respect to the labels for a task than WL-aligned}. Moreover, we show how \textbf{$k$-WL can have negative implications for the fairness, robustness, and privacy of graph ML}. We compare and contrast the findings from our analysis with our survey results. \textbf{Ultimately, graph ML practitioners would benefit from recognizing that: (1) $k$-WL may not be aligned with our task, and we should devise other measurements of expressive power; or (2) in practice, $k$-WL does not limit GNN performance on our benchmarks, and we should construct more rigorous benchmarks for assessing expressive power.} Finally, we argue for extensional definitions and measurement of expressive power and contribute guiding questions for constructing benchmarks for such measurement.

In summary, a primary contribution of our work is disentangling how graph ML practitioners conceptualize expressive power from how we operationalize its measurement. Specifically, we apply measurement modeling to holistically show how $k$-WL operationalizes the measurement of a specific way of thinking about expressive power, and as such, does not always match our expectations of expressive power; we leave it to future work to explore specific misalignments more deeply. Through our investigation, we create room for alternative conceptualizations and operationalizations of expressive power: our adoption of measurement modeling to analyze the pitfalls of $k$-WL (\S\ref{sec:validity_analysis}), auditing of graph ML benchmarks (\S\ref{sec:benchmarking-expressive-power}, \S\ref{sec:consequential-validity}), and guiding questions to facilitate the creation of expressive power benchmarks (\S\ref{sec:better-measurements}, Table \ref{tab:guiding_questions}) constitute an \textbf{initial framework for practitioners to develop and transparently communicate our understandings of expressive power}. Moreover, the critical, sociotechnical nature of our examination of expressive power can inspire practitioners to \textbf{adopt conceptualizations and evaluations of expressive power that prioritize fairness, privacy, and robustness, among other aspects of trustworthiness}.

\section{Preliminaries}

\subsection{Notation} Suppose we have an undirected graph with node features ${\mathcal G} = ({\mathcal V}, {\mathcal E}, X)$. ${\mathcal V}$ is the set of nodes in ${\mathcal G}$. ${\mathcal E} \subseteq {\mathcal V} \times {\mathcal V}$ is the set of edges in ${\mathcal G}$. $X \in \mathbb{F}^{|{\mathcal V}| \times d_{\text{in}}}$ is a $|{\mathcal V}| \times d_{\text{in}}$-dimensional feature matrix in the field $\mathbb{F}$. Similarly, we have another graph ${\mathcal G}' = ({\mathcal V}', {\mathcal E}', X')$.

\subsection{Isomorphisms} We first distinguish between \textit{isomorphic graphs} and \textit{isomorphic nodes}.  An isomorphism is a permutation $\pi : {\mathcal V} \to {\mathcal V}'$ such that: (1) $\pi({\mathcal V}) = {\mathcal V}'$; (2) $\pi({\mathcal E}) = \{ (\pi(i), \pi(j)) | (i, j) \in {\mathcal E} \} = {\mathcal E}'$; and (3) $\pi(X)_{\pi (i)} = X_i'$ \cite{Wang2022HowPA}.
${\mathcal G}, {\mathcal G}'$ are isomorphic if there exists an isomorphism between them. In contrast, an automorphism is a permutation $\pi : {\mathcal V} \to {\mathcal V}$ such that: (1) $\pi({\mathcal V}) = {\mathcal V}$; (2) $\pi({\mathcal E}) = \{ (\pi(i), \pi(j)) | (i, j) \in {\mathcal E} \} = {\mathcal E}$; and (3) $\pi(X)_{\pi (i)} = X_i$ \cite{Wang2022HowPA}. Two nodes $i, j \in {\mathcal V}$ are isomorphic if there exists an automorphism under which $\pi(i) = j$.

\subsection{$k$-WL}
\label{sec:wl-test}
Isomorphism testing, which requires determining if two graphs or nodes are isomorphic, is an NP-intermediate problem \cite{SCHONING1988312}. $k$-WL ($k \geq 1$) is a hierarchy of deterministic, polynomial-time heuristics for isomorphism testing \cite{wlorig, Babai1980Talk}. In 1-WL, each node $i \in {\mathcal V}$ begins with a color $c_i^{0} := \textsc{Hash} (X_i)$, where $\textsc{Hash}$ is an injective hashing function that maps distinct inputs in an arbitrary field to distinct colors, and $X_i \in \mathbb{F}^{d_{in}}$ \cite{wlorig}.
(If the graph does not have node features, every node starts with the same color, or equivalently, modulo a single iteration of $k$-WL, $X_i$ is the degree of $i$.)
Each node then iteratively refines its color as:
\begin{align*}
    c_i^t = \textsc{Hash} \left(c_i^{t - 1}, \{ \{ c_j^{t - 1} | j \in \Gamma_{\mathcal G} (i) \} \} \right),
\end{align*}
where $\{ \{ \cdot \} \}$ is a multiset and $\Gamma_{\mathcal G} (i)$ is the set of neighbors of $i$ in $\mathcal G$. Refinement terminates when $\forall i \in {\mathcal V}, c_i^t = c_i^{t - 1}$, which we denote $c_i$.
Thus, the final colors produced by 1-WL are $\text{1-WL} ({\mathcal G}) = \left( c_i \right)_{i \in {\mathcal V}}$. Let $\textsc{SortedCount} (\cdot)$ be a function that takes as input a vector of colors and outputs a sorted vector of the counts of unique colors. ${\mathcal G}$ and ${\mathcal G}'$ are not isomorphic if $\textsc{SortedCount} (\text{1-WL} ({\mathcal G})) \neq \textsc{SortedCount} (\text{1-WL} ({\mathcal G}'))$; however, if $\textsc{SortedCount} (\text{1-WL} ({\mathcal G})) = \textsc{SortedCount} (\text{1-WL} ({\mathcal G}'))$, 1-WL is inconclusive and ${\mathcal G}$ and ${\mathcal G}'$ may or may not be isomorphic \cite{wlorig}. Similarly, two nodes $i, j \in {\mathcal V}$ are not isomorphic if $c_i \neq c_j$; but, if $c_i = c_j$, 1-WL is inconclusive and $i$ and $j$ may or may not be isomorphic \cite{Wang2022HowPA}.

$k$-WL augments 1-WL to refine colors over $k$-tuples of nodes rather than individual nodes.
The neighborhood of a $k$-tuple of nodes consists of other $k$-tuples that differ in only one node. While 1-WL is exactly as powerful as 2-WL \cite{Immerman1990, Cai1989AnOL}, $k$-WL ($k > 2$) has strictly more distinguishing power than $(k - 1)$-WL \cite{Cai1989AnOL}.
Furthermore, while the exact non-isomorphic graphs and nodes that $k$-WL cannot distinguish are unknown, \cite{Cai1989AnOL} shows that for $k \geq 2$, there exists a pair of non-isomorphic graphs ${\mathcal G}, {\mathcal G}'$ of size $O(k)$ nodes that $k$-WL cannot distinguish.

\subsection{Connecting $k$-WL to GNNs} A GNN ${\mathcal A} : \mathbb{G} \to \mathbb{R}^{d_\text{out}}$ is a neural network that maps a graph ${\mathcal G}$ in the space of all graphs $\mathbb{G}$ to $d_\text{out}$-dimensional real-valued node representations $\left( h_i \right)_{i \in {\mathcal V}}$ or a single whole-graph representation $h_{\mathcal G}$ (depending on whether the task is node or graph-level). Numerous GNN architectures have been proposed, including GCN \cite{Kipf:2016tc}, GIN \cite{xu2018how}, GraphSAGE \cite{hamilton2017}, and GAT \cite{Velickovic2018GraphAN}. The \textit{architecture} of a GNN comprises the operations in each layer and the types of activations between layers, but not the number of layers or parameter values. In contrast, an \textit{instantiation} of an architecture includes a specific number of layers and particular parameter values.

\cite{xu2018how} and \cite{Morris2019WeisfeilerAL} show that GCN can possibly distinguish at most as many pairs of graphs or nodes as non-isomorphic as 1-WL. (We note that \cite{xu2018how} and \cite{Morris2019WeisfeilerAL} assume that nodes do not have informative features \cite{fuchs2023universalityofneural}.) In particular, if 1-WL is inconclusive for ${\mathcal G}, {\mathcal G}'$, then $h_{\mathcal G} = h_{{\mathcal G}'}$. Similarly, if 1-WL is inconclusive for $i, j \in {\mathcal V}$, $h_i = h_j$ \cite{xu2018how}. \cite{xu2018how} proposes Graph Isomorphism Network (GIN), a GNN architecture that is provably as powerful as 1-WL. In other words, if 1-WL can distinguish ${\mathcal G}, {\mathcal G}'$ as non-isomorphic within $L$ iterations, an $L$-layer GIN can theoretically (i.e., with appropriate parameters) produce representations $h_{\mathcal G}$ and $h_{{\mathcal G}'}$ that are different (and similarly for node isomorphisms). \cite{Morris2019WeisfeilerAL} contributes $k$-GNNs, which are designed such that for $k \geq 2$, if $k$-WL can distinguish ${\mathcal G}, {\mathcal G}'$ as non-isomorphic, then a $k$-GNN can theoretically produce distinct representations $h_{\mathcal G}$ and $h_{{\mathcal G}'}$ (and similarly for node isomorphisms).
Maron et al.~\cite{Maron2019InvariantAE, Maron2019Provably} find that higher-order invariant and equivariant GNNs with $k$-th order tensors are as powerful as $k$-WL. %
For a detailed discussion of how expressive power has motivated new GNN architectures, refer to Section V of \cite{Zopf20221WLEI}. For an overview of related works on the representational limits of GNNs, consult \S\ref{sec:representational-limits}. A few papers have identified shortcomings of $k$-WL with respect to measuring the expressive power of GNNs and proposed fixes or alternative understandings of expressive power (cf. \S\ref{sec:beyond-wl}). However, we restrict the scope of our paper to $k$-WL to provide a sufficiently thorough treatment thereof within the page limit. We further justify our focus on $k$-WL in \S\ref{sec:content-validity}, based on our survey results and the citation count of \cite{xu2018how}.

\subsection{Graph tasks}
\label{sec:graph-tasks}
A graph task requires a certain graph-related skill to be demonstrated in the context of a particular input-output format \cite{Bowman2021WhatWI}. We consider \textit{graph-level} and \textit{node-level} tasks. (While we do not explicitly treat pairwise node or pairwise graph tasks, our definitions can be extended to these settings.) In a graph-level task $\tau_{\mathcal G} : \mathbb{G} \to {\mathcal Y}$, we are given as input a graph ${\mathcal G}$, and we must predict a value in the output space $\mathcal Y$ for $\tau_{\mathcal G}$. In a node-level task $\tau_{{\mathcal G}, i} : \mathbb{G} \times {\mathcal V} \to {\mathcal Y}$, we are given as input $\mathcal G$ and a node $i$, and we must predict a value in the output space $\mathcal Y$ for $\tau_{{\mathcal G}, i}$. The \textit{sample space} of a task is the set of all possible graph-label and node-label pairs on which a GNN may be evaluated. In contrast, the \textit{data distribution} of a task is the distribution over the task's sample space that captures the probability of encountering different graph-label and node-label pairs at test time. In this paper, we consider the following common workflow for solving graph tasks with GNNs: $\textsc{Input} \to \textsc{Encoder} \to \textsc{Decoder} \to \textsc{Output}$. \textsc{Encoder} is a GNN ${\mathcal A}$.
\textsc{Decoder} is a function $h$ (e.g., an MLP) that maps $h_{\mathcal G}$ or $h_i$ to a prediction in ${\mathcal Y}$.

\section{Related work}
\label{sec:related-work}

\subsection{Beyond k-WL}
\label{sec:beyond-wl}
A few papers have identified shortcomings of $k$-WL with respect to measuring the expressive power of GNNs and proposed fixes or alternative understandings of expressive power.
\cite{kanatsoulis2023graph} shows that, while $k$-WL would suggest that GNNs have poor discriminative power, GNNs can in fact distinguish any graphs whose spectra differ in at least one eigenvalue. \cite{joshi2023on} augments $k$-WL to be invariant to physical symmetries (e.g., rotation, reflection). \cite{wang2023mathscrnwl} argues that $k$-WL suffers from a ``lack of a natural interpretation and high computational costs,'' and proposes a novel hierarchy of isomorphism tests. \cite{balcilar2021analyzing} contends that $k$-WL neglects to capture the expressive power of GNNs in the spectral domain, which provides a complementary perspective to GNN capabilities in the spatial domain. \cite{zhang2023rethinking} introduces new expressive power metrics based on graph biconnectivity, and finds that existing GNNs are not expressive with respect to these metrics. \cite{pmlr-v202-zhang23k} analyzes the expressive power of subgraph GNNs from the perspective of Subgraph Weisfeiler-Leman Tests and their connection to $k$-WL. \cite{böker2023finegrained} considers more fine-grained, continuous extensions of 1-WL and message-passing neural networks to graphons.

\subsection{Representational limits}
\label{sec:representational-limits}
Numerous prior works have sought to characterize the complexity, hypothesis space, and representational limits of GNNs. 
\cite{Barceló2020The} studies the ability of certain message-passing GNNs to capture $\text{FOC}_2$ logical queries over nodes. \cite{Wang2022HowPA} shows that spectral GNNs can learn arbitrary spectral filters under certain conditions. \cite{Velivckovic2022MessagePA} argues that every permutation equivariant GNN can be expressed with pairwise message passing. \cite{fuchs2023universalityofneural} comments on connections between permutation-invariant universal function approximators on graphs and sets. 
\cite{Grohe2023TheDC} logically characterizes the queries that polynomial-size bounded-depth GNNs can theoretically answer, proving that they are in the circuit complexity class $TC^0$. \cite{AdamDay2023ZeroOneLO} studies the representation and extrapolation capacities of GNNs as their input size increases. \cite{dinverno2023generalization} investigates the ability of GNNs to determine whether an object comprises two identical components (e.g., a graph with two identical cycles).

\subsection{Auditing graph benchmarks}
\label{sec:auditing-benchmarks}
We build on prior and concurrent work that explores the extent to which $k$-WL expressive power is required to perform well on benchmarks (\S\ref{sec:benchmarking-expressive-power}). For instance, our experiments confirm prior findings that large fractions of graph ML benchmarks comprise isomorphic graphs \cite{Ivanov2019UnderstandingIB}. \cite{Zopf20221WLEI}, which was concurrent work, counts the fraction of 1-WL distinguishable graphs in benchmarks, which helps provide an upper bound on GNN performance, and finds this upper bound is often close to 100\%. This suggests that GNNs are not limited by expressive power in practice \cite{Zopf20221WLEI}. Notably, \cite{Zopf20221WLEI} only considers graph isomorphisms, while our work also encompasses node isomorphisms and further studies GIN-WL alignment and the ethical implications of 1-WL.

\section{Survey methodology}

We conduct a survey of graph ML practitioners who are familiar with expressive power to investigate: \textbf{(1)} how clearly and consistently graph ML practitioners conceptualize expressive power; \textbf{(2)} how practitioners perceive the validity of using $k$-WL to measure expressive power; and \textbf{(3)} how practitioners’ conceptualizations and measurements of expressive power are shaped by real-world graph tasks. We begin by asking participants about their background (e.g., occupation, experience with graph ML) and real-world graph tasks with which they are familiar (\hyperref[Q6]{Q6}). We do this to understand the demographics of our sample.
\survey{Participants indicated diverse graph ML experiences, including domains like molecules, knowledge graphs, and physics, and tasks like explainability, graph generation, and structured reasoning (\hyperref[Q5]{Q5}). Participants were also familiar with numerous real-world graph tasks, such as content recommendation, drug discovery, material design, knowledge graph reasoning, and fraud detection (\hyperref[Q4]{Q4}). 14 participants had experience proving expressive power, and 17 had experience evaluating GNNs (\hyperref[Q2]{Q2}, \hyperref[Q3]{Q3}).}

We then inquire in an open-ended manner into how participants conceptualize expressive power; we do this prior to asking participants to engage more critically with expressive power, so as not to sway their responses. Subsequently, we ask participants about their perceptions of the: \textbf{(a)} clarity and consistency with which expressive power is conceptualized; \textbf{(b)} relevance of expressive power to real-world graph tasks; \textbf{(c)} extent to which expressive power encompasses isometry, GNN architecture vs. instantiation, and the data distribution of tasks; and \textbf{(d)} ethical implications of expressive power. For all survey questions that ask participants to rate their perception, we provide them with a scale that ranges from 1 to 6, with articulations of what 1 and 6 mean in the context of the question. We provide additional details on our survey methodology in \S\ref{sec:additional-details-survey}, include the entirety of our survey and aggregate responses in \S\ref{sec:survey_questions}, and address the limitations of our survey and their implications in \S\ref{sec:limitations}.

\section{Reliability and validity analysis}
\label{sec:validity_analysis}

Following measurement modeling, we systematically analyze the reliability and validity of measuring the expressive power of a GNN architecture
via comparison to $k$-WL.

\subsection{Test-retest reliability}
\textit{To what extent do measurements of expressive power obtained via comparison to $k$-WL vary at different points in time?} Such measurements comprise the GNN architecture, which does not change over time, and $k$-WL, which is a deterministic algorithm. Hence, comparison to $k$-WL has test-retest reliability.

\subsection{Face validity}
\textit{To what extent do measurements of expressive power obtained via comparison to $k$-WL appear plausible?} Such measurements usually have face validity because of the step-by-step theoretical justifications in proofs thereof (e.g., \cite{xu2018how, Kim2022PureTA}). Survey respondents rated the apparent validity of proofs of $k$-WL expressive power as \survey{$4.067 \pm 1.438$} (\hyperref[Q21]{Q21}).

\subsection{Content validity (contestedness)}
\label{sec:content-validity}
\textit{To what extent does expressive power have ``multiple context-dependent, and sometimes even conflicting, theoretical understandings'' \cite{jacobs2021}}? This is an important question because the validity of comparing to $k$-WL is influenced by the clarity and consistency with which expressive power is conceptualized. The graph ML community has primarily focused on $k$-WL \survey{(\hyperref[Q16]{Q16}, \hyperref[Q17]{Q17})}. \cite{xu2018how}, one of the first papers to connect GNNs to the WL test, has been cited \survey{5319} times (as of the writing of this paper); however, only a few researchers have proposed alternative understandings of expressive power (cf. \S\ref{sec:beyond-wl}). This would suggest that expressive power is minimally contested. This hypothesis would appear to be confirmed by our survey results as well. On a scale of 1--6, participants rated the clarity with which expressive power is defined as \survey{$4.111 \pm 1.323$} (\hyperref[Q8]{Q8}).
In addition, participants rated the consistency with which expressive power is defined as \survey{$3.722 \pm 1.274$} (\hyperref[Q9]{Q9}).
Furthermore, many participants, when prompted in an open-ended matter, defined expressive power as \survey{the ability of a GNN to discriminate non-isomorphic graphs ($n = 13$) or approximate any function on graphs ($n = 6$)} (\hyperref[Q7]{Q7}). Here, \text{ability} is distinct from \text{capacity} (e.g., model capacity).

However, when prompted more specifically, participants revealed conflicting definitions of expressive power (\hyperref[Q11]{Q11}).
While \survey{8} participants said that expressive power does not involve isometry (i.e., mapping similar graphs to proportionately similar representations), \survey{10} participants said that it does. Furthermore,
while most participants agreed that architecture choice influences expressive power, there was disagreement about whether instantiation (e.g., learned parameters) and the sample space and data distribution of graph tasks have any bearing on expressive power.

\subsection{Content validity (substantative validity)} 
\textit{To what extent does comparison to $k$-WL include all but only those observable properties thought to be related to expressive power?} Expressive power, as it is most commonly conceptualized (\hyperref[Q11]{Q11}),
is solely an architectural property. Thus, because comparison to $k$-WL only involves the GNN architecture, and does not seem to consider extraneous observable properties, comparing to $k$-WL has substantive validity.

\subsection{Content validity (structural validity)}
\label{sec:structural-validity}
\textit{To what extent does comparison to $k$-WL ``capture the structure of the relationships'' between architecture and expressive power \cite{jacobs2021}?} In practice, comparing the pairs of non-isomorphic graphs or nodes that an architecture can theoretically distinguish to those distinguishable by $k$-WL is only useful if we assume that an instantiation of this architecture has parameter values that yield optimal discriminatory power. That is, $k$-WL does not consider the parameter values that an architecture instantiation learns in reality, which indisputably affects the instantiation's ability to distinguish non-isomorphic graphs or nodes. Moreover, comparing to $k$-WL often does not consider the number of layers in an architecture (or that this number is finite) or layer dimensions; these properties also undeniably affect expressive power measurement, as $k$-WL converges in a number of iterations that depends on both the number of nodes and edges in the input graph. Furthermore, expressive power can heavily depend on the input graph; for instance, per the oversquashing effect, certain graphs (e.g., planar graphs) can cause GNNs to exhibit pathological behaviors regardless of which architecture is chosen \cite{Giovanni2023OnOI}. In addition, two GNN architectures in the same $k$-WL realm may perform differently in practice on the same task. Finally, in the real world, graph features and structure are not always fully observable, but comparing to $k$-WL relies on them. As such, comparing to $k$-WL has poor structural validity.

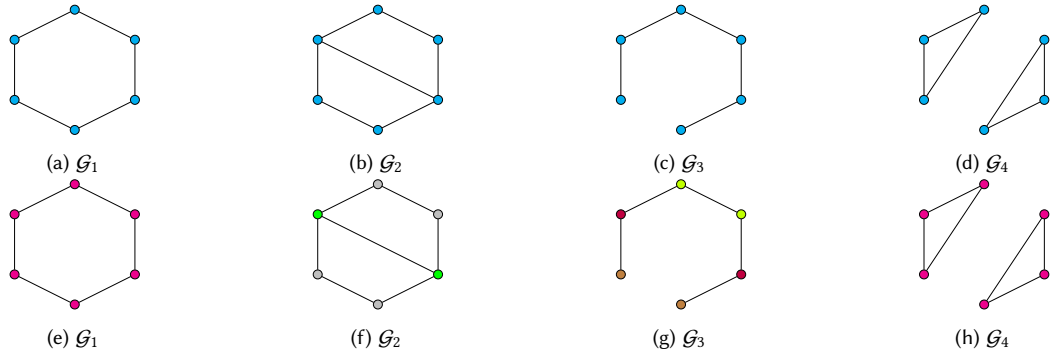
\begin{figure}
\centering
\begin{subfigure}[b]{0.2\textwidth}
\centering
\scalebox{0.4}{
\begin{tikzpicture}
    \node[shape=circle,draw=black,fill=cyan] (A) at (0,0) {};
    \node[shape=circle,draw=black,fill=cyan] (B) at (0,2) {};
    \node[shape=circle,draw=black,fill=cyan] (C) at (2,3) {};
    \node[shape=circle,draw=black,fill=cyan] (D) at (4,2) {};
    \node[shape=circle,draw=black,fill=cyan] (E) at (4,0) {};
    \node[shape=circle,draw=black,fill=cyan] (F) at (2,-1) {} ;

    \path [-] (A) edge (B);
    \path [-](B) edge (C);
    \path [-](C) edge (D);
    \path [-](D) edge (E);
    \path [-](E) edge (F);
    \path [-](F) edge (A);  
\end{tikzpicture}}
\caption{${\mathcal G}_1$}
\end{subfigure}
\hfill
\begin{subfigure}[b]{0.2\textwidth}
\centering
\scalebox{0.4}{
\begin{tikzpicture}
    \node[shape=circle,draw=black,fill=cyan] (A) at (0,0) {};
    \node[shape=circle,draw=black,fill=cyan] (B) at (0,2) {};
    \node[shape=circle,draw=black,fill=cyan] (C) at (2,3) {};
    \node[shape=circle,draw=black,fill=cyan] (D) at (4,2) {};
    \node[shape=circle,draw=black,fill=cyan] (E) at (4,0) {};
    \node[shape=circle,draw=black,fill=cyan] (F) at (2,-1) {} ;

    \path [-] (A) edge (B);
    \path [-](B) edge (C);
    \path [-](C) edge (D);
    \path [-](D) edge (E);
    \path [-](E) edge (F);
    \path [-](F) edge (A);
    \path [-](B) edge (E);
\end{tikzpicture}}
\caption{${\mathcal G}_2$}
\end{subfigure}
\hfill
\begin{subfigure}[b]{0.2\textwidth}
\centering
\scalebox{0.4}{
\begin{tikzpicture}
    \node[shape=circle,draw=black,fill=cyan] (A) at (0,0) {};
    \node[shape=circle,draw=black,fill=cyan] (B) at (0,2) {};
    \node[shape=circle,draw=black,fill=cyan] (C) at (2,3) {};
    \node[shape=circle,draw=black,fill=cyan] (D) at (4,2) {};
    \node[shape=circle,draw=black,fill=cyan] (E) at (4,0) {};
    \node[shape=circle,draw=black,fill=cyan] (F) at (2,-1) {} ;

    \path [-] (A) edge (B);
    \path [-](B) edge (C);
    \path [-](C) edge (D);
    \path [-](D) edge (E);
    \path [-](E) edge (F);
\end{tikzpicture}}
\caption{${\mathcal G}_3$}
\end{subfigure}
\hfill
\begin{subfigure}[b]{0.2\textwidth}
\centering
\scalebox{0.4}{
\begin{tikzpicture}
    \node[shape=circle,draw=black,fill=cyan] (A) at (0,0) {};
    \node[shape=circle,draw=black,fill=cyan] (B) at (0,2) {};
    \node[shape=circle,draw=black,fill=cyan] (C) at (2,3) {};
    \node[shape=circle,draw=black,fill=cyan] (D) at (4,2) {};
    \node[shape=circle,draw=black,fill=cyan] (E) at (4,0) {};
    \node[shape=circle,draw=black,fill=cyan] (F) at (2,-1) {} ;

    \path [-] (A) edge (B);
    \path [-](B) edge (C);
    \path [-](A) edge (C);
    \path [-](D) edge (E);
    \path [-](E) edge (F);
    \path [-](F) edge (D);
\end{tikzpicture}}
\caption{${\mathcal G}_4$}
\end{subfigure}
\bigskip
\begin{subfigure}[b]{0.2\textwidth}
\centering
\scalebox{0.4}{
\begin{tikzpicture}
    \node[shape=circle,draw=black,fill=magenta] (A) at (0,0) {};
    \node[shape=circle,draw=black,fill=magenta] (B) at (0,2) {};
    \node[shape=circle,draw=black,fill=magenta] (C) at (2,3) {};
    \node[shape=circle,draw=black,fill=magenta] (D) at (4,2) {};
    \node[shape=circle,draw=black,fill=magenta] (E) at (4,0) {};
    \node[shape=circle,draw=black,fill=magenta] (F) at (2,-1) {} ;

    \path [-] (A) edge (B);
    \path [-](B) edge (C);
    \path [-](C) edge (D);
    \path [-](D) edge (E);
    \path [-](E) edge (F);
    \path [-](F) edge (A);  
\end{tikzpicture}}
\caption{${\mathcal G}_1$}
\end{subfigure}
\hfill
\begin{subfigure}[b]{0.2\textwidth}
\centering
\scalebox{0.4}{
\begin{tikzpicture}
    \node[shape=circle,draw=black,fill=lightgray] (A) at (0,0) {};
    \node[shape=circle,draw=black,fill=green] (B) at (0,2) {};
    \node[shape=circle,draw=black,fill=lightgray] (C) at (2,3) {};
    \node[shape=circle,draw=black,fill=lightgray] (D) at (4,2) {};
    \node[shape=circle,draw=black,fill=green] (E) at (4,0) {};
    \node[shape=circle,draw=black,fill=lightgray] (F) at (2,-1) {} ;

    \path [-] (A) edge (B);
    \path [-](B) edge (C);
    \path [-](C) edge (D);
    \path [-](D) edge (E);
    \path [-](E) edge (F);
    \path [-](F) edge (A);
    \path [-](B) edge (E);
\end{tikzpicture}}
\caption{${\mathcal G}_2$}
\end{subfigure}
\hfill
\begin{subfigure}[b]{0.2\textwidth}
\centering
\scalebox{0.4}{
\begin{tikzpicture}
    \node[shape=circle,draw=black,fill=brown] (A) at (0,0) {};
    \node[shape=circle,draw=black,fill=purple] (B) at (0,2) {};
    \node[shape=circle,draw=black,fill=lime] (C) at (2,3) {};
    \node[shape=circle,draw=black,fill=lime] (D) at (4,2) {};
    \node[shape=circle,draw=black,fill=purple] (E) at (4,0) {};
    \node[shape=circle,draw=black,fill=brown] (F) at (2,-1) {} ;

    \path [-] (A) edge (B);
    \path [-](B) edge (C);
    \path [-](C) edge (D);
    \path [-](D) edge (E);
    \path [-](E) edge (F);
\end{tikzpicture}}
\caption{${\mathcal G}_3$}
\end{subfigure}
\hfill
\begin{subfigure}[b]{0.2\textwidth}
\centering
\scalebox{0.4}{
\begin{tikzpicture}
    \node[shape=circle,draw=black,fill=magenta] (A) at (0,0) {};
    \node[shape=circle,draw=black,fill=magenta] (B) at (0,2) {};
    \node[shape=circle,draw=black,fill=magenta] (C) at (2,3) {};
    \node[shape=circle,draw=black,fill=magenta] (D) at (4,2) {};
    \node[shape=circle,draw=black,fill=magenta] (E) at (4,0) {};
    \node[shape=circle,draw=black,fill=magenta] (F) at (2,-1) {} ;

    \path [-] (A) edge (B);
    \path [-](B) edge (C);
    \path [-](A) edge (C);
    \path [-](D) edge (E);
    \path [-](E) edge (F);
    \path [-](F) edge (D);
\end{tikzpicture}}
\caption{${\mathcal G}_4$}
\end{subfigure}
\caption{
The top row depicts graphs prior to running 1-WL, and the bottom row depicts the graphs and their colors after running 1-WL till convergence. For simplicity, all nodes have the same initial features.}
\label{fig:example}
\end{figure}

\subsection{Convergent validity}
\label{sec:convergent-validity}
\textit{To what extent does comparing to $k$-WL correlate with accepted measurements of expressive power?} The acceptability of other measurements of expressive power is task-dependent (e.g., graph edit distance, random walk kernel). Consider the task of predicting the edit distance between pairs of graphs. For this task, expressive power can be conceptualized as the ability of an architecture to map pairs of graphs with a large edit distance to proportionately distinct representations, and pairs with a small edit distance to proportionately similar representations. Suppose we have an architecture that is exactly as powerful as 1-WL (i.e., an instantiation of this architecture can theoretically distinguish exactly the same pairs of non-isomorphic graphs as 1-WL). Further, consider the graphs in Figure \ref{fig:example}.
\begin{itemize}[noitemsep,topsep=0pt,parsep=0pt,partopsep=0pt]
    \item ${\mathcal G}_1$ and ${\mathcal G}_2$, which have an edit distance of 1, may have more distinct representations than ${\mathcal G}_1$ and ${\mathcal G}_4$, which have a larger edit distance. This is unintuitive with respect to metric space axioms, and demonstrates that $k$-WL is misaligned with how expressive power may be conceptualized for edit distance prediction.
    \item ${\mathcal G}_3$ and ${\mathcal G}_1$, which have an edit distance of 1, may have distinct representations. Furthermore, ${\mathcal G}_3$ and ${\mathcal G}_4$, which have a larger edit distance, may have distinct representations. However, ${\mathcal G}_1$ and ${\mathcal G}_4$ will have the same representation. This is once again unintuitive and misaligned.
\end{itemize}

These observations threaten the convergent validity of comparing to $k$-WL for edit distance prediction. However, consider a task that requires counting specific substructures of at most size $k$ in graphs. This is possible with an architecture that is at least as expressive as $k$-WL \cite{Chen2020Substructures}. Hence, for substructure counting, comparing to $k$-WL has convergent validity. At the same time, \cite{Loukas2020WhatGN} and \cite{Garg2020GeneralizationAR} show impossibility results for subgraph detection and verification by proving that there exists a graph on which a GNN would require a width or depth dependent on the number of nodes to perform correctly. This highlights a tension between the content validity (cf. \S\ref{sec:structural-validity}) and convergent validity of comparing to $k$-WL for substructure counting. \textbf{Ultimately, defining expressive power and operationalizing its measurement in the context of a particular task, as well as grappling with the tensions in the validity of this measurement, is important.}

\subsection{Discriminant validity}
\textit{To what extent is comparing to $k$-WL capturing aspects of constructs besides expressive power?} Because $k$-WL is tightly coupled with how graph ML practitioners predominantly conceptualize expressive power, comparing to $k$-WL appears to have discriminant validity.

\subsection{Hypothesis validity}
\textit{To what extent do measurements of expressive power obtained by comparing to $k$-WL support hypotheses about expressive power?} Graph ML practitioners often posit that poor expressive power limits
GNN performance on real-world graph tasks. This is confirmed by our survey. Participants rated the relevance of expressive power to real-world graph tasks as \survey{$3.611 \pm 1.461$} (\hyperref[Q10]{Q10}).
Furthermore, participants rated the informativeness of expressive power for real-world graph task performance as \survey{$3.389 \pm
1.195$} (\hyperref[Q18]{Q18}).
However, there are intrinsic limits to $k$-WL, as well as two other regimes in which comparing to $k$-WL is suboptimal.

\subsubsection{Intrinsic limits of k-WL} Similar graphs may not have similar labels (e.g., mismatched graphs \cite{Ivanov2019UnderstandingIB}). For example, consider the extreme case of a task where for some $k$, for each set of non-isomorphic $n$-node graphs ${\mathcal G}_1, \ldots, {\mathcal G}_m$ (where $m \leq 2^{{n \choose 2}}$ and is thus finite) that cannot be distinguished by $k$-WL, we assign them different labels $1, \ldots, m$.
Suppose we have a GNN architecture that is at most as expressive as $k$-WL . Because $k$-WL cannot distinguish these graphs, neither can an instantiation of the architecture, and hence any GNN with this architecture will necessarily score 0\% accuracy on this task. It is also critical to account for discriminatory power over the data distributions for particular tasks. For instance, consider the extreme case of a task whose sample space is exactly the set of non-isomorphic graphs that are not distinguishable by $k$-WL, for some $k$. Suppose we have a GNN architecture that is at most as expressive as $k'$-WL, for any $k' \leq k$. $k'$-WL may be reasonably powerful over all possible graphs \cite{Babai1980RandomGI}; but, because $k'$-WL cannot distinguish any of the graphs in the task's sample space, neither can an instantiation of the architecture, and hence any GNN with this architecture will necessarily score 0\% accuracy on this task. \textbf{The intrinsic limitations of $k$-WL threaten the hypothesis validity of comparing to $k$-WL, and further motivate the need to consider task-specific requirements.}

\subsubsection{Extrinsic limits of $k$-WL (decoder capacity)} Suppose that for some $k$, we have a GNN that is exactly as powerful as $k$-WL, and can produce a unique representation of each non-isomorphic graph in the sample space of a task. This would suggest that the accuracy of a GNN on the task is upper-bounded by 100\%. However, this bound only becomes an equality if: (1) every graph in the task's sample space has a different label, or (2) \textsc{Decoder} has sufficient complexity and capacity to translate representations to correct predictions. In particular, for an arbitrary task, \textsc{Decoder} would need to have been trained on the task's entire sample space and have as high a complexity and capacity as a majority-vote lookup table (or an infinitely wide MLP), which is usually intractable.

For example, consider a task whose sample space is ${\mathcal G}_1, {\mathcal G}_2, {\mathcal G}_3$ in Figure \ref{fig:example}, with labels: $y_{{\mathcal G}_1} = 0, y_{{\mathcal G}_2} = 1, y_{{\mathcal G}_3} = 0$. ${\mathcal G}_1, {\mathcal G}_2, {\mathcal G}_3$ would have distinct representations, but because their labels are not distinct, \textsc{Decoder} will need to have a complex decision boundary. This occurs in part because $k$-WL does not preserve isometry, and hence \textsc{Decoder} cannot exploit the geometry of the GNN's representation space to generalize to graphs in the task's sample space that may not be seen during training or to out-of-distribution graphs. In this way, \textbf{$k$-WL can be antithetical to generalization}, even if it is \textbf{sample-efficient}. This observation is important because our survey participants generally indicated that they prioritize generalization (\survey{$5.0 \pm 1.372$}) over expressive power  (\survey{$4.167 \pm 1.098$}) (\hyperref[Q27]{Q27}).

\subsubsection{Extrinsic limits of $k$-WL (learning dynamics)} In practice, GNNs may not learn representations that align with the colorings produced by $k$-WL, or even be able to (e.g., due to inductive biases of gradient descent, data sampling). However, survey participants indicated that they believe that GNNs can \survey{almost always ($3.714 \pm 0.825$)}
learn such representations (\hyperref[Q25]{Q25}). Furthermore, depending on the task, it may be more optimal for GNNs to learn representations that are not aligned with $k$-WL colorings. In other words, $k$-WL-aligned representations may not be optimal for a specific task, even if they are optimal for arbitrary tasks. For example, consider a task whose sample space is the ${\mathcal G}_1, {\mathcal G}_2, {\mathcal G}_4$ in Figure \ref{fig:example}, with labels: $y_{{\mathcal G}_1} = 0, y_{{\mathcal G}_2} = 0, y_{{\mathcal G}_4} = 1$. The 1-WL colorings for ${\mathcal G}_1, {\mathcal G}_2$ would be distinct, but since these graphs share the same label, it would be more optimal for a GNN to learn identical or similar representations for both, so that \textsc{Decoder} requires a less complex decision boundary.

\subsection{Benchmarking expressive power}
\label{sec:benchmarking-expressive-power}
We investigate the extent to which $k$-WL colorings are relevant to and 1-WL colorings empirically align with GNN representations on real-world graph tasks, and compare our findings with our survey results. We run experiments with 1-WL and GIN (which is exactly as powerful as 1-WL in theory) on popular graph-level and node-level benchmarks (Table \ref{tab:benchmarks}). We audit these benchmarks because they have been used by seminal papers that connect $k$-WL to GNNs to validate the effectiveness of more expressive architectures  \cite{xu2018how, Morris2019WeisfeilerAL, Zopf20221WLEI}. The benchmarks are from common graph domains (e.g., bioinformatics, social networks, citation networks), and some have node features while others do not.
Herein, we assume these benchmarks are reasonable measurements of real-world task performance; we critically expand on this assumption in \S\ref{sec:graph-tasks-benchmarks}.

\subsubsection{Relevance of 1-WL to benchmarks} We partition benchmark instances (i.e., graphs for graph-level benchmarks, and nodes for node-level benchmarks) into three sets of equivalence classes: (1) instances with the same label ${\mathcal E}_y$; (2) instances that are isomorphic ${\mathcal E}_\pi$; and (3) instances with the same 1-WL colorings after $t$ iterations of refinement ${\mathcal E}_{\text{WL}}^{t}$. We denote the cardinality of a set of equivalence classes using $| \cdot |$, and the number of singleton equivalence classes in a set using $\#_1 (\cdot)$. For example, $\#_1 ({\mathcal E}_\pi)$ signifies the number of unique instances in a benchmark, and $\frac{\#_1 ({\mathcal E}_{\text{WL}}^3)}{\#_1 ({\mathcal E}_\pi)}$ captures the proportion of 1-WL distinguishable instances to unique instances after three iterations of refinement.

As shown in Table \ref{tab:equivalence-class-statistics}, $| {\mathcal E}_\pi |$ is often lower than the number of instances, indicating that the benchmarks (e.g., IMDB-BINARY, IMDB-MULTI, PTC\_MR) contain numerous isomorphic instances; upon further inspection, the isomorphic nodes are all duplicates. This finding would suggest that it is advantageous for a GNN to be able to distinguish non-isomorphic graphs. However, for IMDB-BINARY, IMDB-MULTI, REDDIT-BINARY\footnote{The REDDIT-BINARY dataset was previously compiled and made publicly available by a third party.}, PROTEINS, CiteSeer, and PubMed, $| {\mathcal E}_\pi | = |{\mathcal E}_{\text{WL}}^{1}|$, which evidences that 1-WL is able to distinguish all the non-isomorphic instances in these benchmarks after only one iteration. For the remaining benchmarks, 1-WL can distinguish all or almost all non-isomorphic instances within three iterations. We provide example 1-WL non-distinguishable graph pairs from MUTAG in Figure \ref{fig:non-distinguishable-mutag}. Our findings suggest that, in theory, a GNN that is as expressive as 1-WL would be able to perform close to perfectly on these benchmarks.
To confirm that ${\mathcal E}_\pi$ and ${\mathcal E}_{\text{WL}}^{3}$ are indeed the same (or similar) partitions, we compute the symmetric adjusted mutual information (AMI) between these sets.
We see in Figure
\ref{fig:other-benchmarks-ami} that, for all graph-level benchmarks, ${\mathcal E}_\pi$ and ${\mathcal E}_{\text{WL}}^{3}$ have an AMI of 1 or close to 1, indicating these partitions are identical or nearly identical.

\begin{table*}
  \small
  \caption{Graph ML benchmarks with size and equivalence class statistics.}
  \label{tab:equivalence-class-statistics}
  \centering
  \begin{adjustbox}{max width=\textwidth}
  \begin{tabular}{l|rr|rr|rrrrrr}
\toprule
{} &  \# graphs &  Avg \# nodes/graph &  $|{\mathcal E}_\pi|$ &  $\#_1 ({\mathcal E}_\pi)$ &  $|{\mathcal E}_{\text{WL}}^1|$ &  $\#_1 ({\mathcal E}_{\text{WL}}^1)$ &  $|{\mathcal E}_{\text{WL}}^2|$ &  $\#_1 ({\mathcal E}_{\text{WL}}^2)$ &  $|{\mathcal E}_{\text{WL}}^3|$ &  $\#_1 ({\mathcal E}_{\text{WL}}^3)$ \\
\midrule
IMDB-BINARY \cite{Yanardag2015DeepGK}   &       1000 &               19.77 &               537 &                    421 &                         537 &                              421 &                         537 &                              421 &                         537 &                              421 \\
IMDB-MULTI \cite{Yanardag2015DeepGK}    &       1500 &               13.00 &               387 &                    288 &                         387 &                              288 &                         387 &                              288 &                         387 &                              288 \\
REDDIT-BINARY \cite{Yanardag2015DeepGK} &       2000 &              429.63 &              1998 &                   1996 &                        1998 &                             1996 &                        1998 &                             1996 &                        1998 &                             1996 \\
PROTEINS \cite{Borgwardt2005ProteinFP}      &       1113 &               39.06 &              1069 &                   1050 &                        1069 &                             1050 &                        1069 &                             1050 &                        1069 &                             1050 \\
PTC\_MR \cite{Helma2001ThePT}        &        344 &               14.29 &               328 &                    313 &                         315 &                              291 &                         328 &                              313 &                         328 &                              313 \\
MUTAG \cite{Debnath1991StructureactivityRO}         &        188 &               17.93 &               175 &                    164 &                          90 &                               53 &                         167 &                              152 &                         171 &                              158 \\
\midrule
Cora \cite{Sen2008CollectiveCI}          &          1 &             2708.00 &              2693 &                   2683 &                        2693 &                             2683 &                        2693 &                             2683 &                        2693 &                             2683 \\
CiteSeer \cite{Sen2008CollectiveCI}      &          1 &             3327.00 &              3319 &                   3311 &                        3319 &                             3311 &                        3319 &                             3311 &                        3319 &                             3311 \\
PubMed \cite{Sen2008CollectiveCI}        &          1 &            19717.00 &             19717 &                  19717 &                       19717 &                            19717 &                       19717 &                            19717 &                       19717 &                            19717 \\
\bottomrule
\end{tabular}
\end{adjustbox}
\end{table*}

\subsubsection{Predictive power of isomorphisms and 1-WL}

\begin{wraptable}{r}{6cm}
\caption{Accuracy of majority-vote classifier on predicting labels from isomorphic graphs and 1-WL distinguishable graphs.}
\label{tab:majority-classifier-performance}
\scalebox{0.75}{%
\begin{tabular}{lrr}
\toprule
{} &  $h({\mathcal E}_\pi \to {\mathcal E}_y)$ &  $h({\mathcal E}_{\mathrm{WL}}^3 \to {\mathcal E}_y)$ \\
\midrule
IMDB-BINARY   &                          88.60 &                                      88.60 \\
IMDB-MULTI    &                          63.27 &                                      63.27 \\
REDDIT-BINARY &                         100.00 &                                     100.00 \\
PROTEINS      &                          99.73 &                                      99.73 \\
PTC\_MR        &                          99.13 &                                      99.13 \\
MUTAG         &                         100.00 &                                     100.00 \\
Cora          &                         100.00 &                                     100.00 \\
CiteSeer      &                          99.97 &                                      99.97 \\
PubMed        &                         100.00 &                                     100.00 \\
\bottomrule
\end{tabular}}
\end{wraptable}

Figure
\ref{fig:other-benchmarks-ami} reveals that ${\mathcal E}_\pi$ and ${\mathcal E}_{\text{WL}}^{3}$ share an AMI of close to 0 with ${\mathcal E}_y$, which suggests that these partitions are close to independent. That is, distinguishing non-isomorphic graphs provides effectively no information about the labels of these graphs. As discussed previously, this necessitates that \textsc{Decoder} has a high capacity and complex decision boundary. Interestingly, the AMI of ${\mathcal E}_y$ with ${\mathcal E}_{\text{WL}}^{3}$ can be higher than that with ${\mathcal E}_\pi$ because non-isomorphic graphs often have the same label. To further investigate the label predictive power that ${\mathcal E}_\pi$ and ${\mathcal E}_{\text{WL}}^{3}$ grant, we evaluate the accuracy of the best possible deterministic graph classifier $h$ \cite{Zopf20221WLEI}: a lookup table based on majority vote.

In particular, for an input graph ${\mathcal G}$, let $e_{\mathcal G} \in {\mathcal E}_\pi$ be the isomorphism equivalence class to which ${\mathcal G}$ belongs; then, the prediction $\hat{y}_{\mathcal G} = \text{mode} \{ y_{{\mathcal G}'} | {{\mathcal G}'} \in e_{\mathcal G} \}$ (and similarly for ${\mathcal E}_{\text{WL}}^{3}$). The accuracy of $h$ (displayed in Table \ref{tab:majority-classifier-performance}) provides an upper bound on the performance of any deterministic GNN. We see that the accuracy is or is close to 100\% for almost all the benchmarks. This demonstrates that 1-WL can produce colorings that are sufficient to solve common graph ML tasks; however, survey participants opined that 1-WL does so only \survey{sometimes ($3.067 \pm 0.594$)} (\hyperref[Q23]{Q23}). Furthermore, since GNNs do not achieve near perfect performance on these benchmarks in practice, our results suggest that neither isomorphism testing nor 1-WL expressive power, and rather decoder complexity \& capacity and generalization, limit GNN performance in practice. Moreover, 1-WL does not help provide a useful upper bound on the accuracy of a GNN on every task. Along this vein, survey participants indicated that 1-WL admits a useful upper bound only \survey{sometimes ($3.133 \pm 0.640$)} (\hyperref[Q22]{Q22}).  

We note that our observation that even with the discriminative power of 1-WL, effectively all the non-isomorphic graphs and nodes in the benchmarks are distinguishable, is not expected. Although \cite{Babai1980RandomGI} shows that 1-WL can distinguish almost all graphs (in a probabilistic sense), graph ML benchmarks do \textit{not} comprise random graphs. Moreover, more powerful tests (e.g., 3-WL, 4-WL, etc.) will produce the same color partitions of graphs and nodes as 1-WL for the benchmarks that we audit and necessarily cannot provide a better upper bound on performance; hence, we do not provide experimental results for these more powerful tests. Our aforementioned observation also suggests that the benchmarks are not good assessments of harder tasks like triangle counting \cite{Barceló2020The}.

\begin{figure}[!ht]
    \centering
    \includegraphics[width=0.75\textwidth]{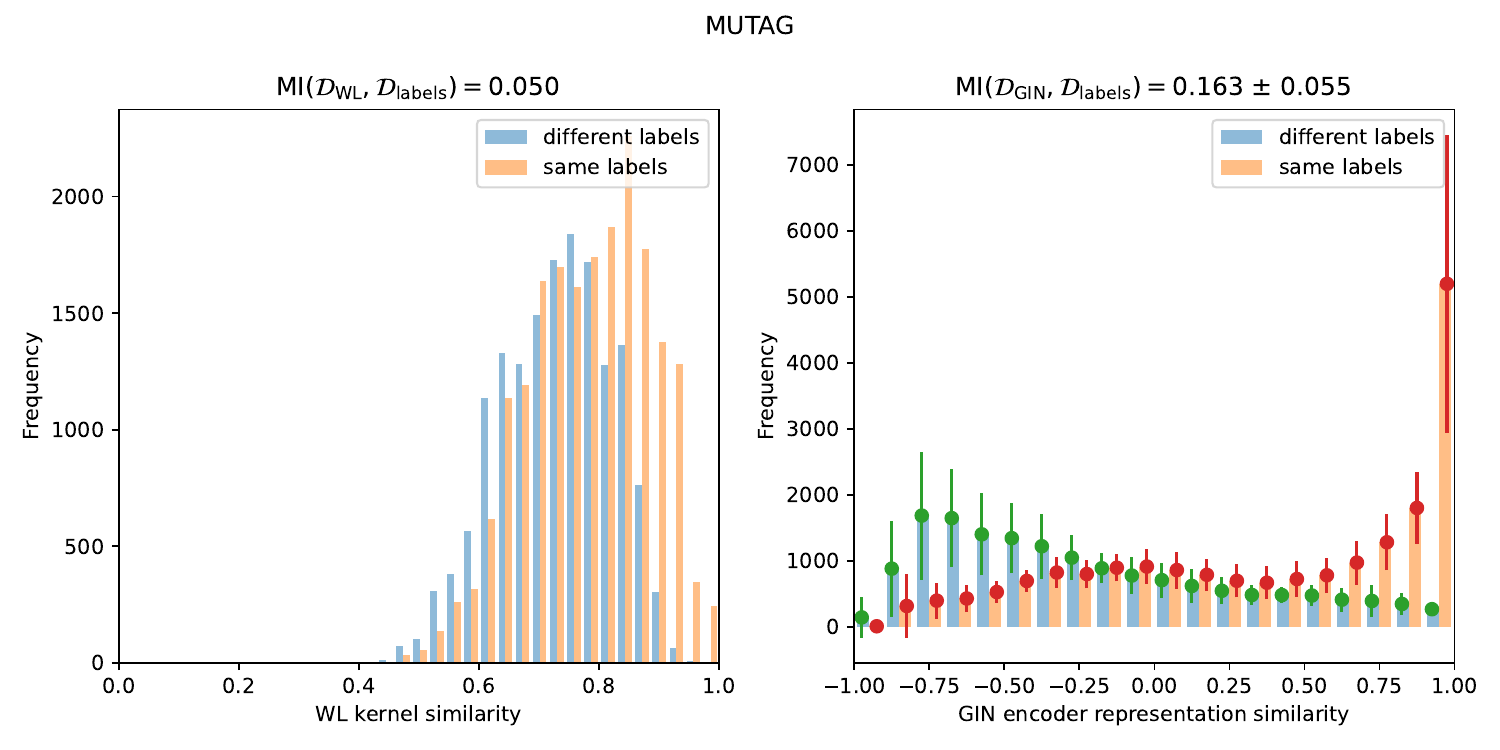}
    \includegraphics[width=0.75\textwidth]{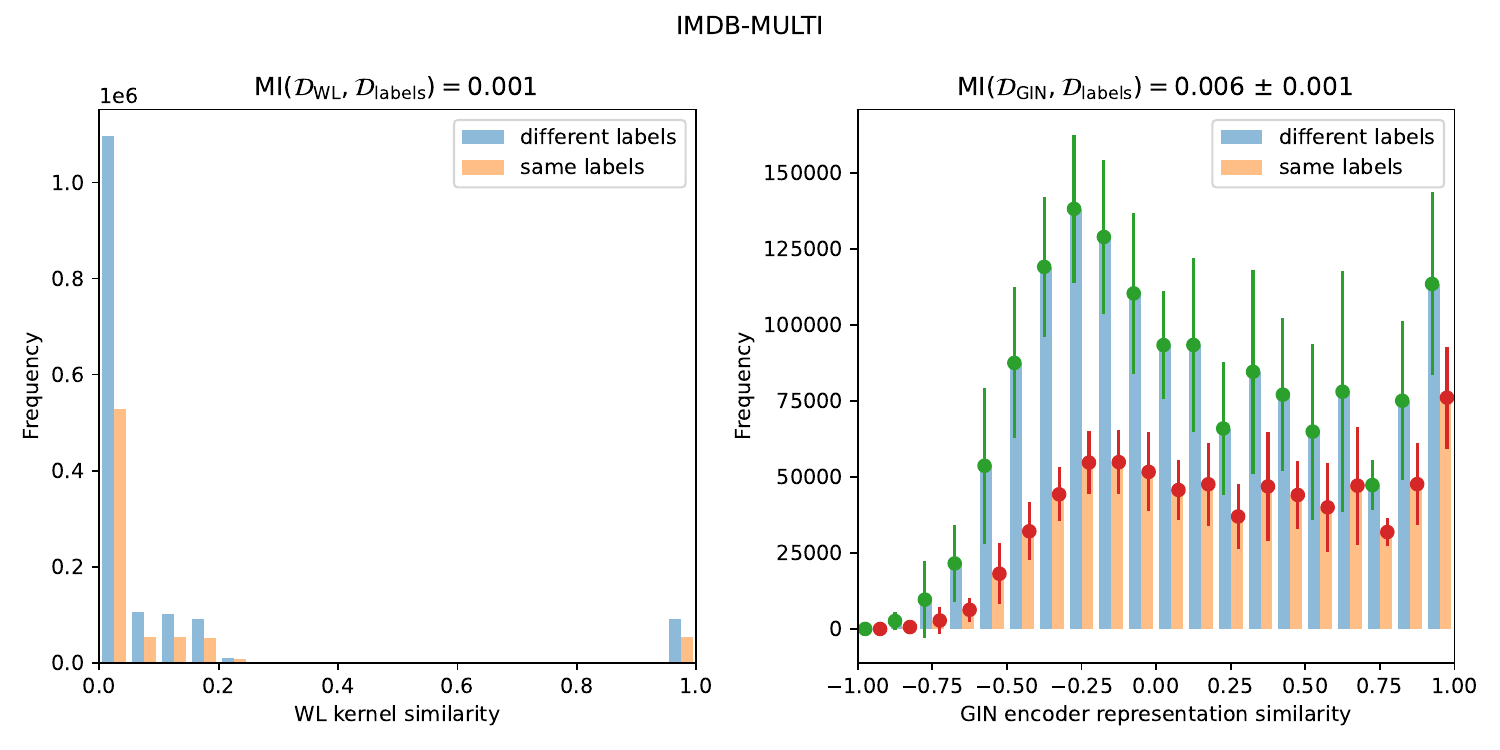}
    \includegraphics[width=0.75\textwidth]{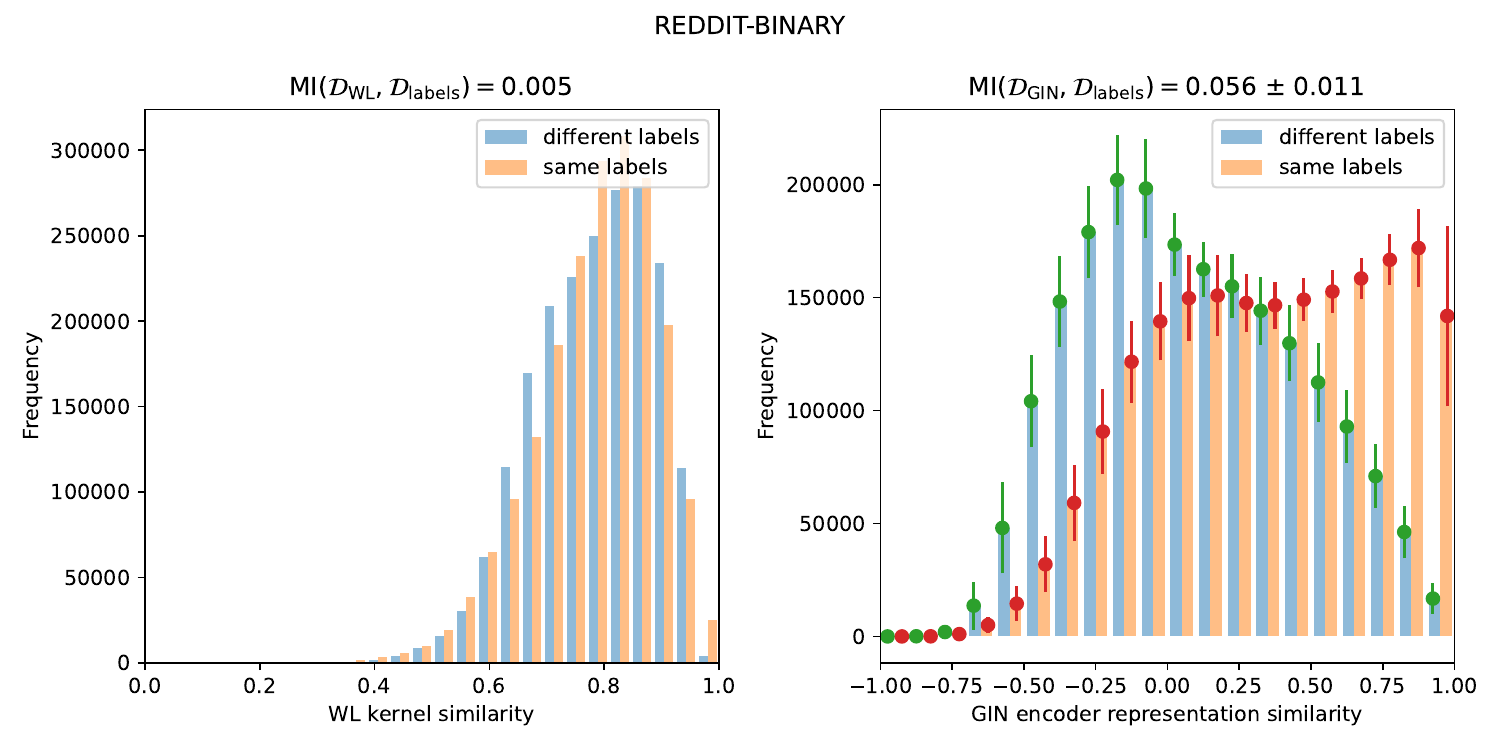}
    \caption{Distributions of WL kernel similarities and GIN encoder representation similarities of graph pairs with different vs. the same labels.}
    \label{fig:mutag-alignment}
\end{figure}

\subsubsection{GIN alignment with 1-WL} 
We investigate how well GIN representations align with 1-WL colorings in practice. We choose GIN  because, by construction, it is theoretically as powerful as 1-WL \cite{xu2018how}. We use the default experimental settings from \cite{xu2018how} (cf. \S\ref{sec:experimental-settings} for more details). For each graph-level benchmark, we jointly train a GIN encoder and MLP decoder \cite{Paszke_PyTorch_An_Imperative_2019, Fey/Lenssen/2019}. For graph pairs ${\mathcal G}, {\mathcal G}'$, we compute the cosine similarity ($\in [-1, 1]$) of their GIN representations $h_{\mathcal G}, h_{{\mathcal G}'}$ (i.e., the output of the encoder). We also compute the WL subtree kernel cosine similarity ($\in [0, 1]$) of ${\mathcal G}, {\mathcal G}'$ \cite{Shervashidze2011WeisfeilerLehmanGK, Siglidis2018GraKeLAG}, after four iterations of color refinement (to match the number of GIN layers). The WL kernel measures similarities in node colorings after 1-WL refinement, and upon convergence, is equivalent to comparing the number of shared subtrees between graphs \cite{Shervashidze2011WeisfeilerLehmanGK}.

Figures \ref{fig:mutag-alignment} and \ref{fig:other-benchmarks-alignment} show the distributions ${\mathcal D}_{\text{WL}}^{\text{same}}$ and ${\mathcal D}_{\text{WL}}^{\text{different}}$ of WL kernel similarities of graph pairs with different vs. the same labels, and similarly, the distributions ${\mathcal D}_{\text{GIN}}^{\text{same}}$ and ${\mathcal D}_{\text{GIN}}^{\text{different}}$ of GIN encoder representation similarities. For almost all the benchmarks (except PTC\_MR), ${\mathcal D}_{\text{WL}}^{\text{same}}$ and ${\mathcal D}_{\text{WL}}^{\text{different}}$ are much less divergent than ${\mathcal D}_{\text{GIN}}^{\text{same}}$ and ${\mathcal D}_{\text{GIN}}^{\text{different}}$, which means that the mutual information (MI) of the GIN representation similarities with label matches is higher than the MI of the WL kernel similarities with the label matches. This is confirmed by the MI calculations (discretized over 20 bins) displayed in the figures. Ultimately, the figures show that GIN learns representations that are more optimal with respect to the labels than WL-aligned. This could be because node features are significantly more informative for label prediction than graph structure for many tasks \cite{fuchs2023universalityofneural}. Furthermore, this ensures that \textsc{Decoder} need not be overly complex to predict the labels. Complementing this empirical finding, survey participants indicated that they believe that GNNs only \survey{sometimes ($3.200 \pm 0.676$)} learn 1-WL aligned representations (\hyperref[Q24]{Q24}).

\paragraph{Takeaway} \textbf{Based on our analysis, graph ML practitioners would benefit from recognizing that: (1) $k$-WL may not be aligned with our task, and we should devise other measurements of expressive power; or (2) in practice, $k$-WL does not limit GNN performance on our benchmarks, and we should construct more rigorous benchmarks for assessing $k$-WL expressive power.}

\subsection{Consequential validity}
\label{sec:consequential-validity}
\textit{What are the social consequences of measuring expressive power by comparing to $k$-WL?} Such measurements have likely contributed to a snowball effect of graph ML research universalizing the specific conceptualization of expressive power encoded by $k$-WL, which
has potentially led them to: (1) neglect task-driven formulations of expressive power (\survey{$3.444 \pm 1.042$} (\hyperref[Q17]{Q17})); and (2) prioritize expressive power (\survey{$4.167 \pm
1.098$}) over ethical aspects of functionality like efficiency (\survey{$4.556 \pm
1.149$}), robustness (\survey{$4.556 \pm 1.097$}), fairness (\survey{$3.167 \pm
1.543$}), and privacy (\survey{$2.667 \pm 1.815$}) (\hyperref[Q27]{Q27}). In fact, survey participants rated the ethical implications of expressive power as \survey{$2.444 \pm
1.423$} (\hyperref[Q12]{Q12}).

However, $k$-WL can have implications for individual fairness and adversarial robustness. With these concerns in mind, we desire that our GNN ${\mathcal A}$ produces representations that are not overly sensitive to (small) changes in the input (i.e., is Lipschitz continuous) \cite{Dwork2012Awareness}. More formally, for graph-level tasks, we would like that $\forall {\mathcal G}, {\mathcal G}' \in {\mathcal S}, \| h_{\mathcal G} - h_{{\mathcal G}'} \|_p \leq L \cdot d({\mathcal G}, {\mathcal G}')$, where $\mathcal S$ is our task sample space, $\| \cdot \|_p$ is the $L_p$-norm, $L$ is a global Lipschitz constant for ${\mathcal A}$, and $d$ is a distance metric over graphs (e.g., edit distance, random walk kernel similarity). Similarly, for node-level tasks, we would like that $\forall i, j \in {\mathcal V}, \| h_i - h_j \|_p \leq L_{\mathcal G}\cdot d_{\mathcal G} (i, j)$, where $d_{\mathcal G}$ is a distance metric over nodes in ${\mathcal G}$ (e.g., shortest path). However, as discussed in \S\ref{sec:convergent-validity}, it is unknown if such global Lipschitz constants exist when ${\mathcal S}$ and ${\mathcal G}$ are not finite. Furthermore, $d({\mathcal G}, {\mathcal G}')$ is not always intuitively predictive of $\| h_{\mathcal G} - h_{{\mathcal G}'} \|_p$ (for graph-level tasks) and $d_{\mathcal G} (i, j)$ of $\| h_i - h_j \|_p$ (for node-level tasks).

$k$-WL can also have implications for privacy. Consider the graphs from Figure \ref{fig:example}. $({\mathcal G}_1, {\mathcal G}_2)$ and $({\mathcal G}_1, {\mathcal G}_3)$ are pairs of neighboring graphs because they differ in one edge; in contrast, $({\mathcal G}_1, {\mathcal G}_4)$ are not neighboring graphs. Let ${\mathcal N} ({\mathcal S})$ be the set of all neighboring graphs in a task's sample space ${\mathcal S}$. Then, for graph-level tasks, we define the sensitivity $\Delta ({\mathcal A}, {\mathcal S})$ of ${\mathcal A}$ over ${\mathcal S}$ as $\max_{({\mathcal G}, {\mathcal G}') \in {\mathcal N} ({\mathcal S})} \| h_{{\mathcal G}} - h_{{\mathcal G}'} \|_1$. Often, to obtain $\epsilon$-node differential privacy, practitioners add $\text{Lap}(\Delta({\mathcal A}, {\mathcal S}) / \epsilon)^{d_\text{out}}$ noise to $h_{\mathcal G}$ \cite{Kasiviswanathan2013AnalyzingGW}. As discussed in \S\ref{sec:convergent-validity}, $\Delta ({\mathcal A}, {\mathcal S})$ can be large: considering $({\mathcal G}_1, {\mathcal G}_2)$ and $({\mathcal G}_1, {\mathcal G}_3)$, adding or removing an edge can greatly change the representations produced by a $k$-WL-powerful GNN; intuitively, this is problematic if these graphs were ego social networks, as changes in graph representations could undesirably reveal the existence of high-degree nodes or symmetries in users' social circles to an adversary. As such, practitioners may need to add significant noise to achieve private representations.

\begin{wraptable}{r}{5cm}
  \small
  \caption{Benchmarks with fraction of 1-WL identifiable graphs.}
  \label{tab:ethics-equivalence-class-statistics}
  \centering
  \scalebox{0.75}{%
\begin{tabular}{lrl}
\toprule
{} &  \# nodes & $\#_1 ({\mathcal E}_{\text{WL}}^3)$ \\
\midrule
\textsc{Credit}   &     30000 & 29367 (97.89\%) \\
\textsc{Credit} (Age $\leq 25$) &     27315 & 26720 (97.82\%) \\
\textsc{Credit} (Age $> 25$) &      2685 & 2649 (98.66\%) \\
\textsc{German}   &      1000 & 1000 (100\%) \\
\bottomrule
\end{tabular}}
\end{wraptable}

To empirically examine the ethical implications of 1-WL, we use the \textsc{Credit} and \textsc{German} networks from \cite{agarwal2021unified}. (For more information on these benchmarks, consult \S\ref{sec:ethics-benchmarks-overview}.) Table \ref{tab:ethics-equivalence-class-statistics} shows that all or almost all nodes in \textsc{Credit} and \textsc{German} are uniquely identifiable after just three iterations of 1-WL, which poses a privacy risk. This also presents individual fairness and robustness risks because it suggests that 1-WL colorings are sensitive to small feature and structural differences between nodes. Furthermore, Table \ref{tab:ethics-equivalence-class-statistics} suggests group fairness concerns with respect to age for \textsc{Credit}: (1) nodes with age $> 25$ are uniquely identifiable at a higher rate, which means that a majority-vote lookup table classifier will perform better on them; (2) however, they also have greater privacy risks.

\section{Towards task-driven, practical, and trustworthy\\ definitions and measurements of expressive power}
\label{sec:better-measurements}

Addressing the validity issues in \S\ref{sec:validity_analysis}, we advocate for \textit{extensional} definitions and measurement of expressive power to \textit{complement} intensional ones \cite{Schlangen2020TargetingTB}. An intensional definition (e.g., $k$-WL) comprises theoretical properties of GNNs, and is often stated in terms of logic; intensional definitions have helped develop mathematical thinking, guided research into poorly-understood properties of GNNs, and aided in data modeling \cite{fuchs2023universalityofneural}. An extensional definition of expressive power is grounded in dataset and metric-based evaluation: a GNN is considered ``expressive'' in the context of a task if it performs well on a benchmark for the task. \textbf{We explicitly do not propose an intensional alternative to $k$-WL, as in the spirit of the No Free Lunch Theorem, the pitfalls of $k$-WL will not be magically cured by another universal criterion.} We argue that extensional expressive power is:

\begin{itemize}[noitemsep,topsep=0pt,parsep=0pt,partopsep=0pt]
    \item \textbf{Task-driven:} Extensional definitions reduce the contestedness of expressive power, as expressive power will be defined in the context of a specific task, thus drawing from domain knowledge. Furthermore, expressive power will be measured based on task-specific (i.e., accepted) metrics, so extensional measurement will improve convergent validity.
    \item \textbf{Practical:} Extensional measurement encompasses the complex relationships between the innumerable elements of graph ML (e.g., architecture, training, data structure), thereby improving content validity. Additionally, extensional measurement ensures that higher measurements of expressive power correspond to improved task performance, and promotes generalization.
    \item \textbf{Trustworthy:} Extensional measurement encourages empirically examining potential tradeoffs between expressive power and ethical aspects of graph ML (e.g., robustness, fairness, privacy). However, there may be concerns about its reliability and face validity (cf. \S\ref{sec:graph-tasks-benchmarks}).
\end{itemize}

\begin{table*}
\centering
\begin{adjustbox}{max width=0.9\textwidth}
\small
\begin{tabular}{|p{0.95\textwidth}|}
\midrule
\textbf{\textsc{Conceptualization}}\\
\midrule
\textbf{Capabilities:} Which GNN capabilities related to expressive power do you believe the task involves? \\
\textbf{Correctness:} How may the ground truth or acceptable/unacceptable methods for performing the task be contested? How would you accurately characterize ``solving'' the task? \\
\textbf{Community:} How may the benchmark limit ``progress'' to only working on one conceptualization of expressive power? Do you hold space for others to propose alternative understandings of expressive power? \\
\midrule
\textbf{\textsc{Operationalization}} \\
\midrule
\textbf{Validity:} What kinds of validity may the benchmark lack and why? If the benchmark were to indicate that a GNN performs exceptionally well on it, what can the graph ML community realistically conclude about the expressive power of the GNN? \\
\hline
\end{tabular}
\end{adjustbox}
\caption{Guiding questions to facilitate the creation of expressive power benchmarks.}
\label{tab:guiding_questions}
\end{table*}

Importantly, with extensional expressive power, \textit{conceptualizations} of expressive power and \textit{operationalizations} of its measurement are more closely linked. We draw from \cite{subramonian-etal-2023-tango} to provide guiding questions to facilitate the creation of GNN expressive power benchmarks (Table \ref{tab:guiding_questions}).
In our survey, participants rated the use of expressive power to inform architecture design as \survey{$4.333 \pm 1.085$} (\hyperref[Q28]{Q28}), but benchmark design as \survey{$3.833 \pm 1.249$} (\hyperref[Q29]{Q29}); we hope extensional measurement shifts these findings. Extensional measurement is already being practiced; for example, \cite{Velivckovic2022TheCA} carefully designs the CLRS benchmark to assess GNN expressive power in the context of algorithmic reasoning tasks (e.g., shortest path).

\section{Conclusion}

We uncover misalignments between practitioners' conceptualizations of expressive power and $k$-WL through a systematic analysis of the reliability and validity of $k$-WL. Our survey ($n = \survey{18}$) reveals that practitioners' conceptualizations of expressive power differ. In addition, our graph-theoretic analysis shows that $k$-WL can be misaligned with accepted measurements for a task, does not guarantee isometry, and can be antithetical to generalization.
Complementarily, our benchmark auditing reveals that $k$-WL can be irrelevant to solving many real-world graph tasks, and detrimental to trustworthiness.
Finally, we argue for extensional measurements of expressive power and propose guiding questions for constructing benchmarks for such measurements. Ultimately, our adoption of measurement modeling to analyze the pitfalls of $k$-WL (\S\ref{sec:validity_analysis}), auditing of graph ML benchmarks (\S\ref{sec:benchmarking-expressive-power}, \S\ref{sec:consequential-validity}), and guiding questions to facilitate the creation of expressive power benchmarks (\S\ref{sec:better-measurements}, Table \ref{tab:guiding_questions}) constitute an initial framework for graph ML practitioners to develop and transparently communicate our understandings of expressive power. This framework will further inspire practitioners to adopt understandings of expressive power that center trustworthiness (e.g., fairness audits, usable privacy assessments, robustness monitoring). We detail limitations and future work in \S\ref{sec:limitations}, and include a responsible research checklist in \S\ref{sec:responsible-research-checklist}.

\section*{Acknowledgments}
We thank Shichang Zhang and Soledad Villar for their insightful and valuable feedback on this paper. We are further grateful to Hannes Stärk, Chaitanya Joshi, and Xavier Bresson for helping share our survey with a wider audience.

\bibliographystyle{ACM-Reference-Format}
\bibliography{sample-base}

\clearpage

\appendix

\begin{center}
    \Large \textbf{Appendices}
\end{center}

\textbf{\large{Contents in Appendices:}}
\begin{itemize}
    \item In Appendix~\ref{sec:additional-details-survey}, we provide additional details on our survey methodology.
    \item In Appendix~\ref{sec:survey_questions}, we provide all our survey questions and aggregate responses.
    \item In Appendix~\ref{sec:graph-tasks-benchmarks}, we critically expand upon the distinction between graph ML tasks and benchmarks.
    \item In Appendix~\ref{sec:benchmarks-overview}, we overview the benchmarks we audit.
    \item In Appendix~\ref{sec:non-distinguishable-mutag}, we include pairs of graphs from MUTAG that 1-WL cannot distinguish.
    \item In Appendix~\ref{sec:app-ami}, we provide the adjusted mutual information (AMI) scores for all the benchmarks we audit.
    \item In Appendix~\ref{sec:experimental-settings}, we discuss the settings we used to train and evaluate GIN.
    \item In Appendix~\ref{sec:app-alignment}, we provide our remaining GIN and 1-WL alignment results for the benchmarks we audit.
    \item In Appendix~\ref{sec:limitations}, we detail the limitations of our work and future work.
    \item In Appendix~\ref{sec:responsible-research-checklist}, we include a responsible research checklist, based on the NeurIPS checklist.
\end{itemize}

\clearpage

\section{Additional details on survey methodology}
\label{sec:additional-details-survey}

\paragraph{Participant guidance} We ask participants to list real-world applications of graph ML with which they are familiar to ground their responses to questions about the relevance of expressive power to real-world graph tasks. Furthermore, we use a scale of 1--6 (with articulations of what 1 and 6 mean) to: a) capture the distribution of participants' responses with sufficient granularity, b) impel participants to lean towards one side of the scale, and c) improve the consistency of how participants interpret answer choices.

\paragraph{Quality control} We piloted our survey with a couple of graph ML practitioners in order to identify potential problems with the clarity of our questions. We further provided participants with the opportunity to optionally justify their responses or indicate disagreement or a lack of clarity with any definitions or questions; \survey{only a couple of participants did so}. We intentionally included a few free-response questions to deter and remove spammers from our sample.

\paragraph{Participant recruitment and IRB}
\label{sec:survey_participant_recruitment_irb}
Following \cite{Zhou2022DeconstructingNE}, by ``practitioners,'' we refer to academic and industry researchers, applied scientists, and engineers. Survey participants must have experience theoretically characterizing the expressive power of a GNN, or empirically evaluating GNNs on real-world tasks. We shared our survey as a Google form multiple times on Twitter, graph ML-focused Slack workspaces, and mailing lists and Slack channels for artificial intelligence (AI) affinity groups like Queer in AI \cite{ovalle2023queer} and Women in Machine Learning\footnote{\url{https://wimlworkshop.org/}}. We additionally shared the survey at a tech company via an internal communication platform. In all cases, we requested participants to share the survey with other relevant groups in order to perform snowball sampling \cite{Naderifar2017SnowballSA}. Participants were not compensated due to internal bureaucratic hurdles. We obtained informed consent (refer to \S\ref{sec:survey_consent_form}) from all participants, no personally identifiable or sensitive information was collected, and the survey was reviewed by an internal privacy team at a tech company.

\clearpage

\section{Survey questions and responses}
\label{sec:survey_questions}

* Indicates required questions.

\textbf{Probing the Validity of Measurements of Graph Neural Network Expressive Power}

The goal(s) of this project is to:
\begin{enumerate}
    \item investigate how clearly and consistently graph machine learning practitioners conceptualize graph neural network expressive power;
    \item uncover how practitioners perceive the validity of common measurement models for expressive power;
    \item inquire into the extent to which practitioners’ conceptualizations of expressive power and measurements thereof are driven by and impact real-world applications.
\end{enumerate}
Survey results will be presented in aggregate in the form of a research paper, which will be submitted to a conference in 2023. 

This survey will take approximately 5-10 minutes. We thank you for your participation.

Researchers involved: [REDACTED FOR ANONYMITY]

If you have any questions or concerns, please contact [REDACTED FOR ANONYMITY].

\subsection{Consent}
\label{sec:survey_consent_form}

\phantomsection\label{Q1}[Q1] Do you consent to the terms of this survey? *

\textit{Participants must be at least 18 years of age and will not be compensated for their participation. No personally identifiable or sensitive information will be collected. This survey has been reviewed by an internal privacy review team.}
\begin{itemize}[noitemsep,topsep=0pt,parsep=0pt,partopsep=0pt]
    \item[$\ocircle$] Yes
    \item[$\ocircle$] No \textcolor{red}{[if selected, survey branches to final section]}
\end{itemize}

\clearpage

\subsection{Background}

\phantomsection\label{Q2}[Q2] Do you have any experience theoretically characterizing the expressive power of a graph neural network? *
\begin{itemize}[noitemsep,topsep=0pt,parsep=0pt,partopsep=0pt]
    \item[$\ocircle$] Yes
    \item[$\ocircle$] No
\end{itemize}

\begin{figure}[!ht]
    \centering
    \includegraphics[width=0.45\textwidth]{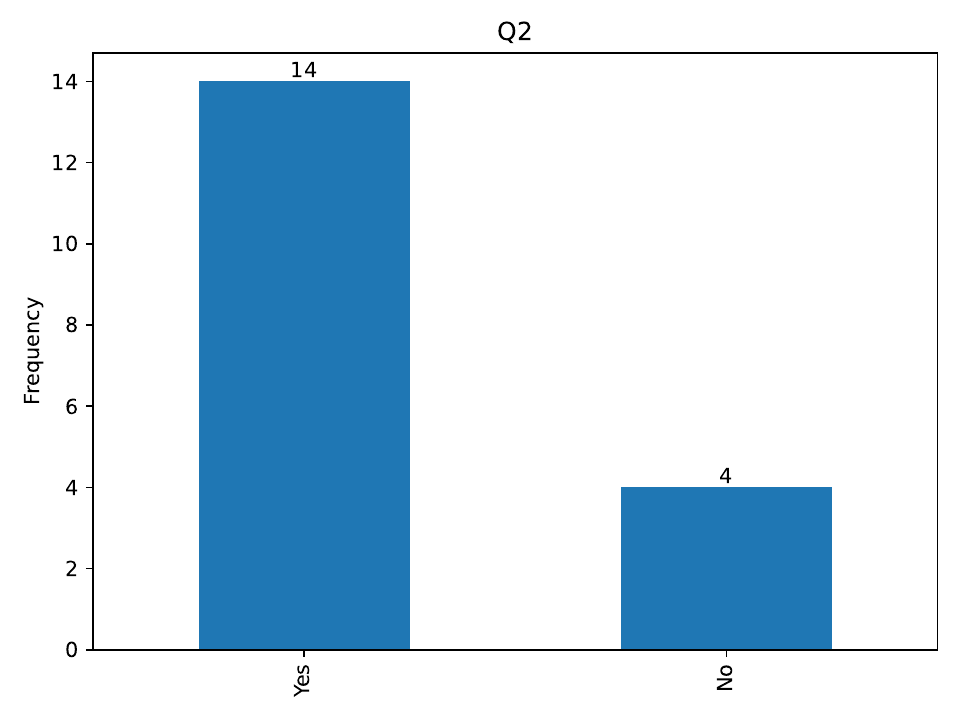}
\end{figure}
\FloatBarrier

\phantomsection\label{Q3}[Q3] Do you have any experience empirically evaluating graph neural networks on real-world tasks or datasets? *
\begin{itemize}[noitemsep,topsep=0pt,parsep=0pt,partopsep=0pt]
    \item[$\ocircle$] Yes
    \item[$\ocircle$] No
\end{itemize}

\begin{figure}[!ht]
    \centering
    \includegraphics[width=0.45\textwidth]{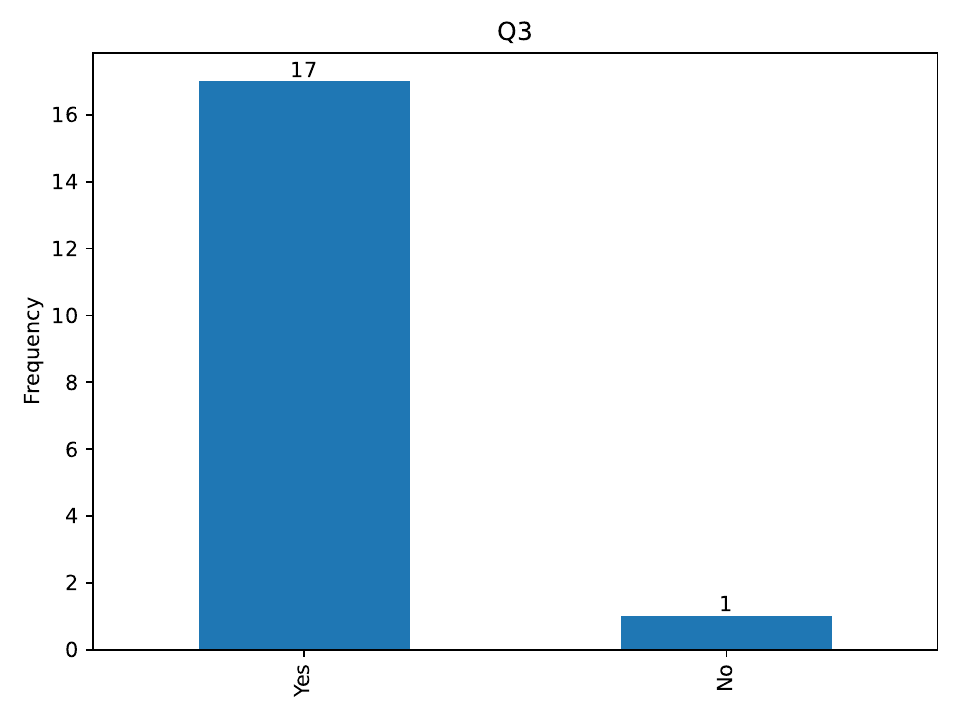}
\end{figure}
\FloatBarrier

\clearpage

\phantomsection\label{Q4}[Q4] List some real-world applications of graph machine learning with which you are familiar. *
\begin{itemize}[noitemsep,topsep=0pt,parsep=0pt,partopsep=0pt]
    \item Open text field
\end{itemize}

\begin{figure}[!ht]
    \centering
    \includegraphics[width=0.45\textwidth]{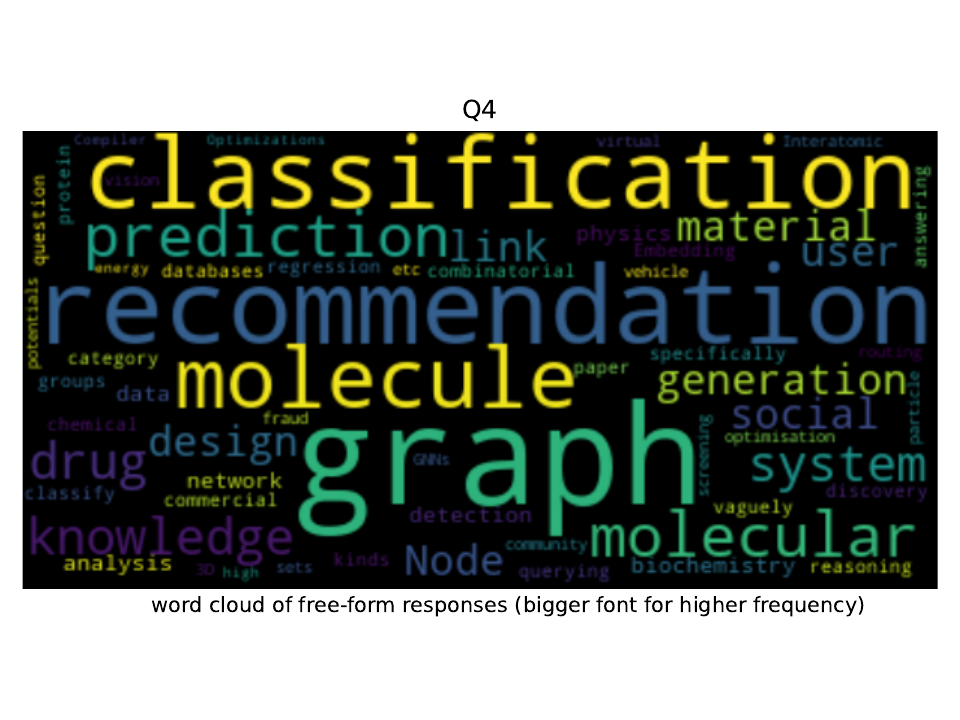}
\end{figure}
\FloatBarrier

\phantomsection\label{Q5}[Q5] Briefly describe the type of graph machine learning work that you do. *

\textit{Please be mindful not to bring up any identifying or sensitive information about yourself or third-parties.}
\begin{itemize}[noitemsep,topsep=0pt,parsep=0pt,partopsep=0pt]
    \item Open text field
\end{itemize}

\begin{figure}[!ht]
    \centering
    \includegraphics[width=0.45\textwidth]{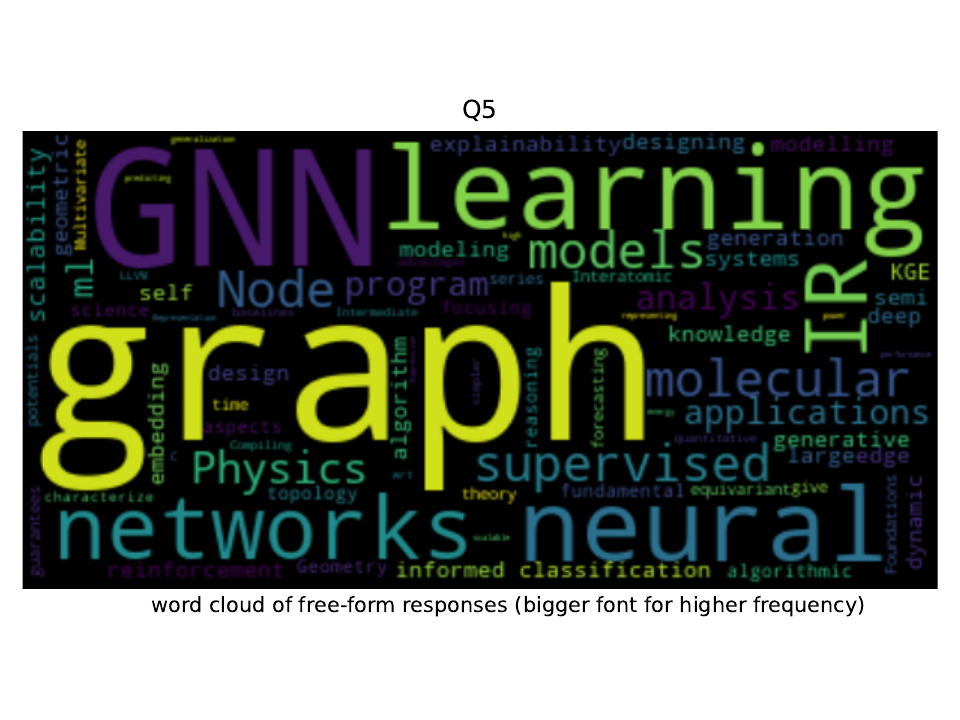}
\end{figure}
\FloatBarrier

\clearpage

\phantomsection\label{Q6}[Q6] Please select all the options that apply to you. *
\begin{itemize}[noitemsep,topsep=0pt,parsep=0pt,partopsep=0pt]
    \item[$\square$] I work on deployed systems
    \item[$\square$] I am an industry practitioner (not researcher)
    \item[$\square$] I am an industry researcher
    \item[$\square$] I am an academic researcher
    \item[$\square$] I am a student researcher
\end{itemize}

\begin{figure}[!ht]
    \centering
    \includegraphics[width=0.45\textwidth]{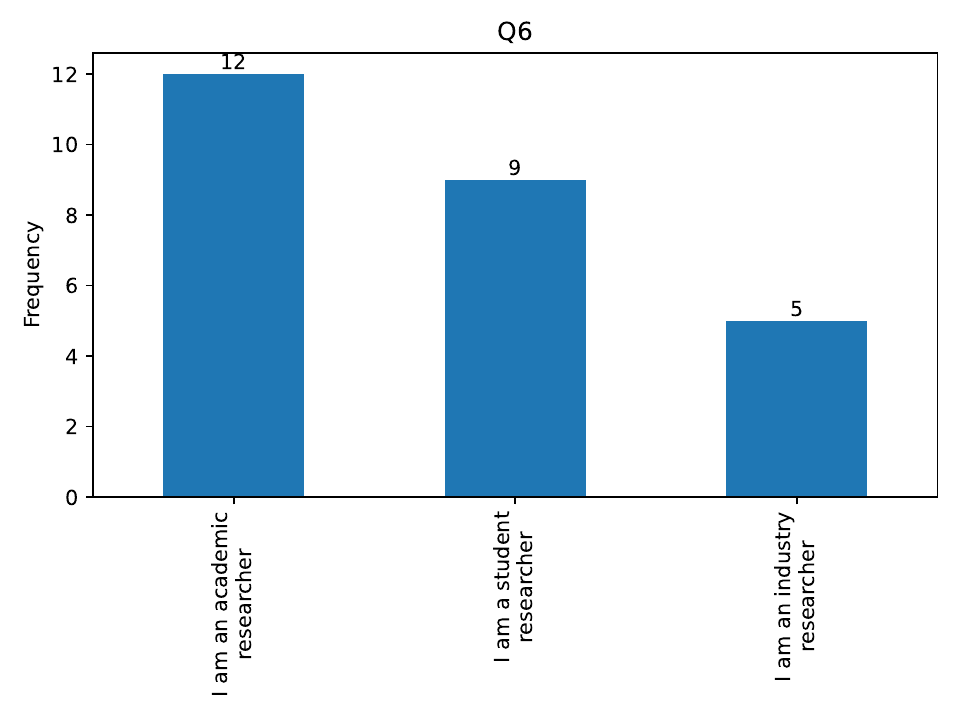}
\end{figure}
\FloatBarrier

\subsection{Conceptualization Questions}

We consider the following system setup:

Input Graph(s) $\to$ Graph Neural Network (GNN) Encoder $\to$ Graph Representation(s) $\to$ Head (e.g., linear model, MLP) $\to$ Prediction

Furthermore, we only consider graph-level prediction tasks (e.g., molecular property prediction, graph similarity prediction), wherein the input is a whole graph(s). To be clear, we do not consider node-level prediction tasks.

\phantomsection\label{Q7}[Q7] How do you define expressive power? *

\textit{Consider what properties you would like an expressive GNN encoder to satisfy. Try to avoid using heuristics.}
\begin{itemize}[noitemsep,topsep=0pt,parsep=0pt,partopsep=0pt]
    \item Open text field
\end{itemize}

\begin{figure}[!ht]
    \centering
    \includegraphics[width=0.45\textwidth]{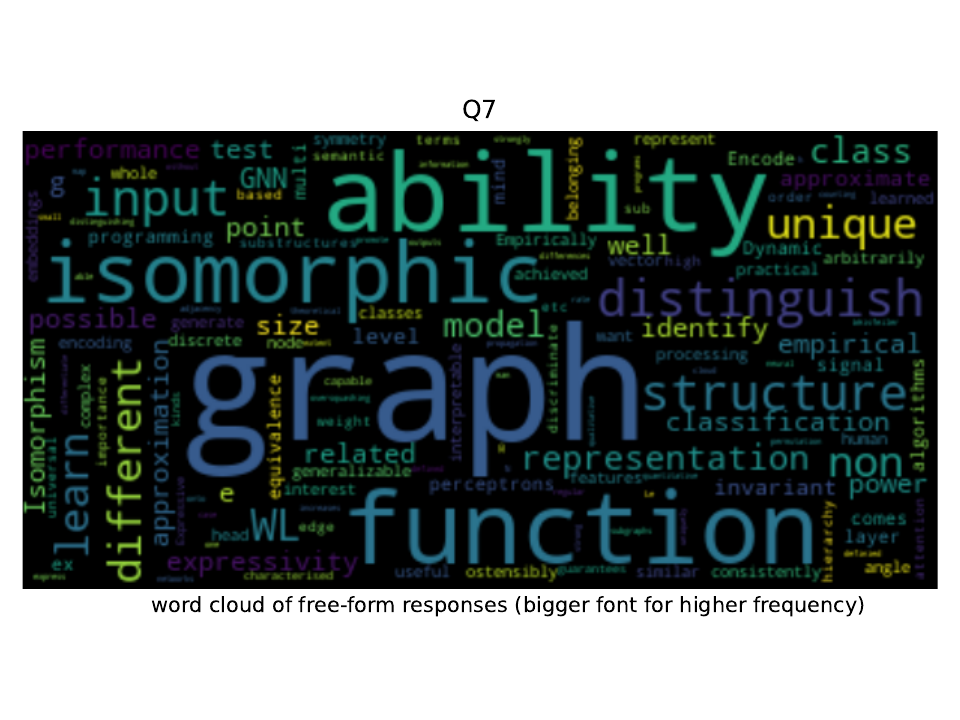}
\end{figure}
\FloatBarrier

\clearpage

\phantomsection\label{Q8}[Q8] How clearly defined is the expressive power of graph neural networks? *

\textit{6 (clearly defined) means that expressive power has definitions that are clearly articulated and understood by the graph ML community (even if these definitions are not consistent). In contrast, 1 (unclearly defined) means that expressive power is unclear to most people in the community.}
\begin{tabular}{ccccccc}

1 & 2 & 3 & 4 & 5 & 6  \\
$\ocircle$ & $\ocircle$ & $\ocircle$ & $\ocircle$ & $\ocircle$ & $\ocircle$
\end{tabular}

\begin{figure}[!ht]
    \centering
    \includegraphics[width=0.45\textwidth]{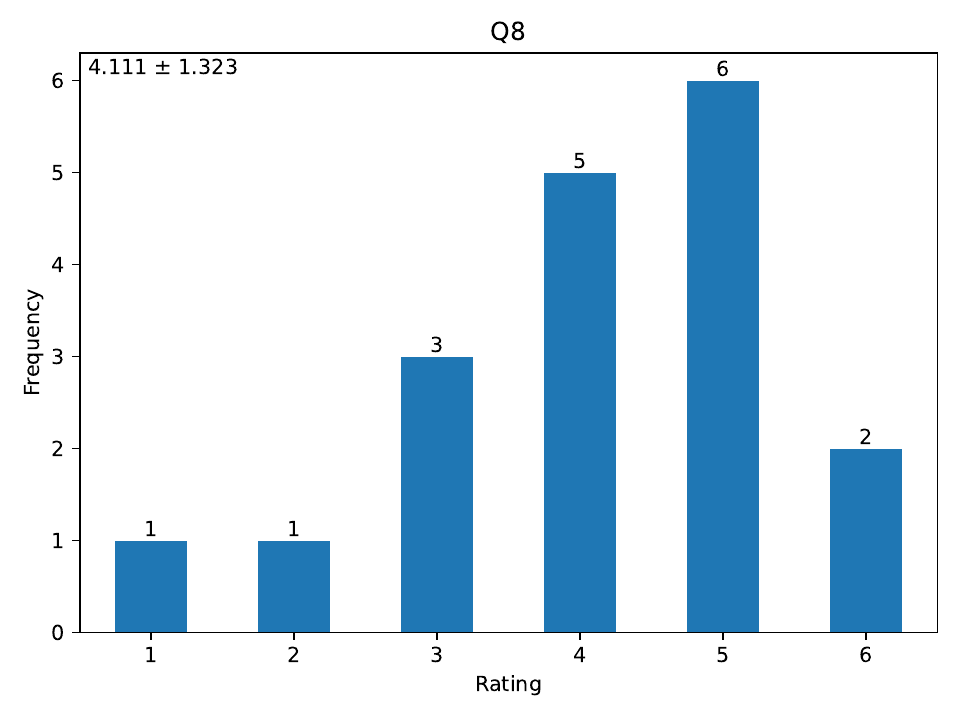}
\end{figure}
\FloatBarrier

\phantomsection\label{Q9}[Q9] How consistently defined is the expressive power of graph neural networks? *

\textit{6 (consistently defined) means that expressive power is consistently defined by the graph ML community. In contrast, 1 (inconsistently defined) means that expressive power is understood differently from person to person in the community.}

\begin{tabular}{ccccccc}
1 & 2 & 3 & 4 & 5 & 6  \\
$\ocircle$ & $\ocircle$ & $\ocircle$ & $\ocircle$ & $\ocircle$ & $\ocircle$
\end{tabular}

\begin{figure}[!ht]
    \centering
    \includegraphics[width=0.45\textwidth]{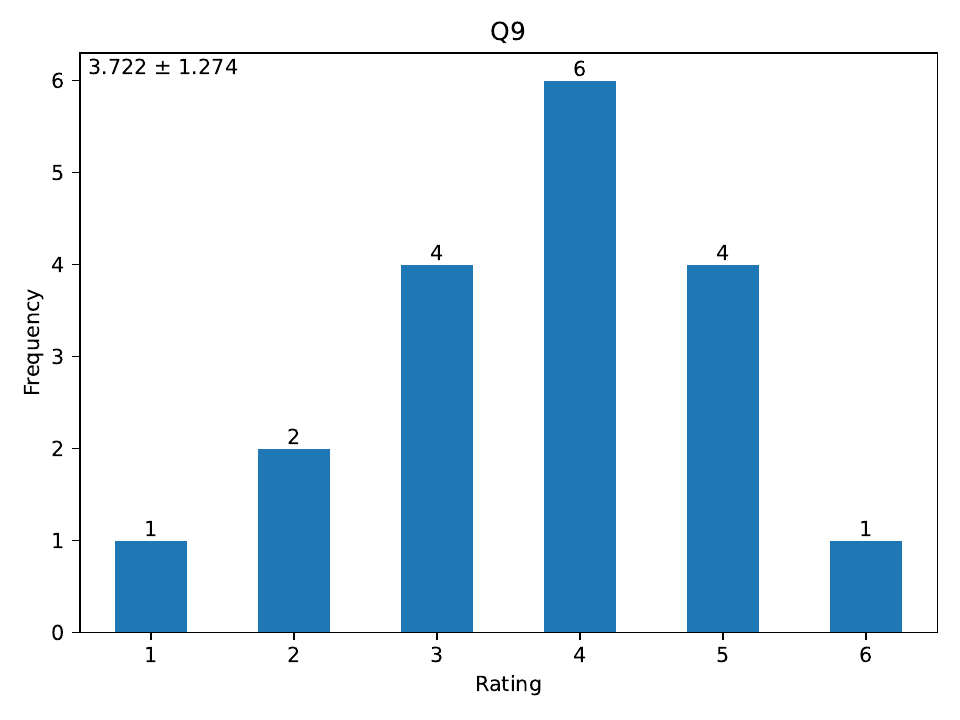}
\end{figure}
\FloatBarrier

\clearpage

\phantomsection\label{Q10}[Q10] How relevant are common conceptualizations of expressive power to the real-world applications of graph machine learning that you listed in the Background Questions section? *

\textit{6 (directly relevant) means that a graph neural network with higher expressive power is guaranteed to improve system performance on most real-world applications. 1 (not relevant) means that a graph neural network with higher expressive power has a low chance of increasing system performance on real-world applications.}

\begin{tabular}{ccccccc}
1 & 2 & 3 & 4 & 5 & 6  \\
$\ocircle$ & $\ocircle$ & $\ocircle$ & $\ocircle$ & $\ocircle$ & $\ocircle$
\end{tabular}

\begin{figure}[!ht]
    \centering
    \includegraphics[width=0.45\textwidth]{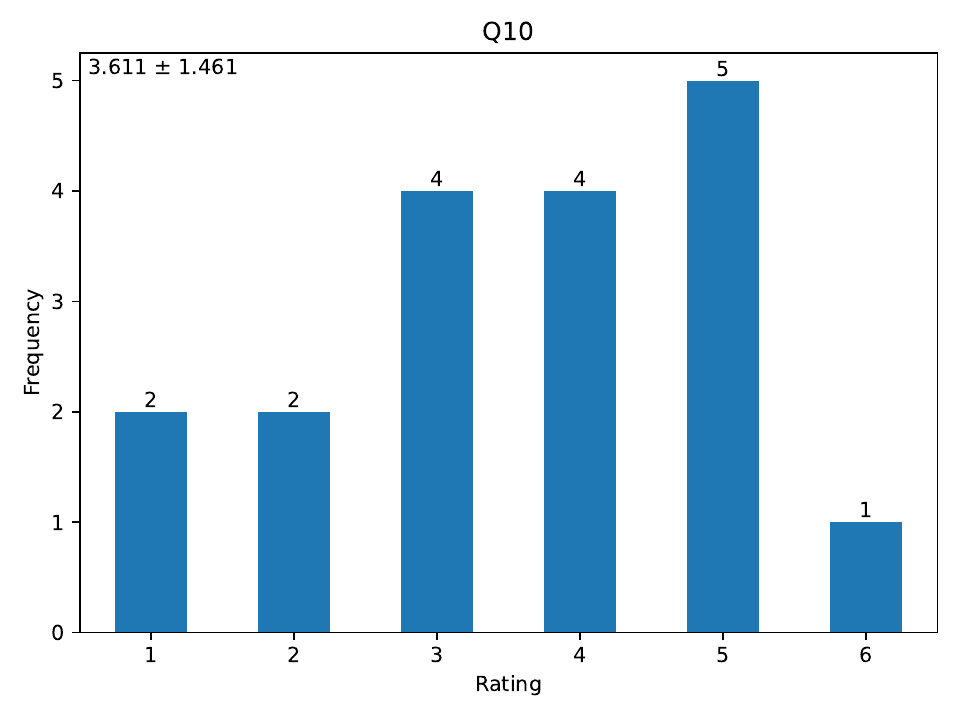}
\end{figure}
\FloatBarrier

\clearpage

\phantomsection\label{Q11}[Q11] Please select all of the following options that describe how you conceptualize expressive power: *

\textit{Definitions:}

\textit{The ARCHITECTURE of a graph neural network encoder comprises the operations in each layer and the types of activations between layers, but not the number of layers or parameter values.}

\textit{An INSTANTIATION of an ARCHITECTURE comprises a specific number of layers and particular parameter values.}

\textit{The SAMPLE SPACE of a task refers to the set of all possible graph-label pairs on which a MODEL may be evaluated.}

\textit{The DATA DISTRIBUTION of a task refers to the distribution over the task's SAMPLE SPACE that characterizes the probability of encountering different graph-label pairs.}

\begin{itemize}[noitemsep,topsep=0pt,parsep=0pt,partopsep=0pt]
    \item[$\square$] Different (i.e., non-isomorphic) graphs can be mapped to uniquely identifiable representations
    \item[$\square$] Similar graphs can be mapped to proportionately similar (i.e., isometric) representations
    \item[$\square$] Expressive power is affected by ARCHITECTURE
    \item[$\square$] Expressive power is affected by an INSTANTIATION
    \item[$\square$] Expressive power is affected by the SAMPLE SPACE of a task
    \item[$\square$] Expressive power is affected by the DATA DISTRIBUTION of a task
\end{itemize}

\begin{figure}[!ht]
    \centering
    \includegraphics[width=0.45\textwidth]{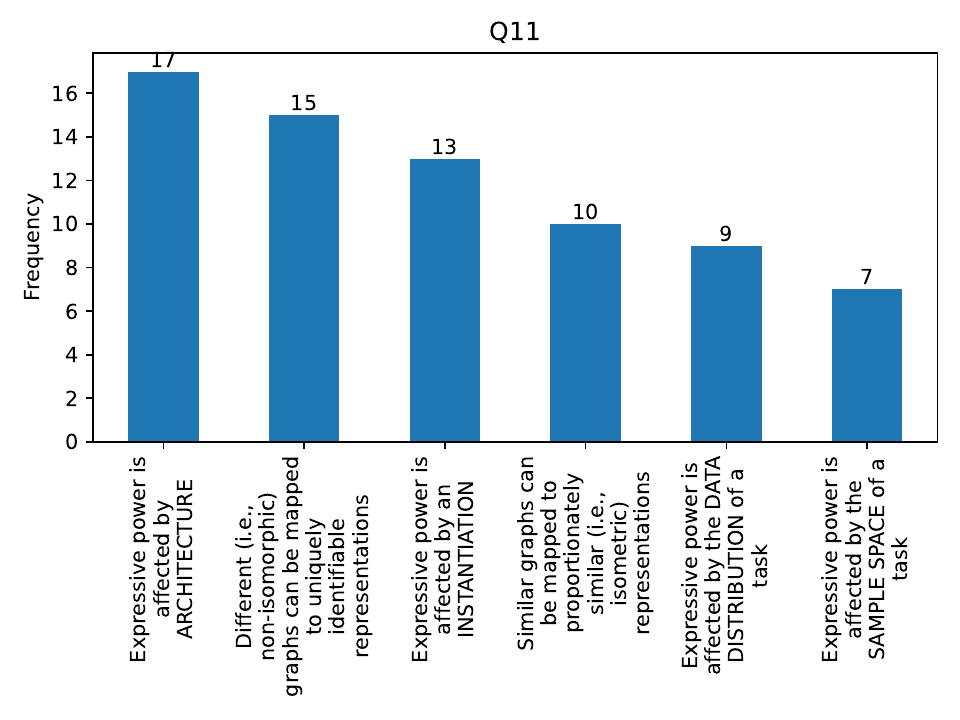}
\end{figure}
\FloatBarrier

\clearpage

\phantomsection\label{Q12}[Q12] How much does expressive power have ethical implications? *

\textit{6 (large extent) means that a graph neural network with higher expressive power has direct and critical ethical implications for graph learning tasks. 1 (small extent) means that expressive power has no clear connection to ethics.}

\begin{tabular}{ccccccc}
1 & 2 & 3 & 4 & 5 & 6  \\
$\ocircle$ & $\ocircle$ & $\ocircle$ & $\ocircle$ & $\ocircle$ & $\ocircle$
\end{tabular}

\begin{figure}[!ht]
    \centering
    \includegraphics[width=0.45\textwidth]{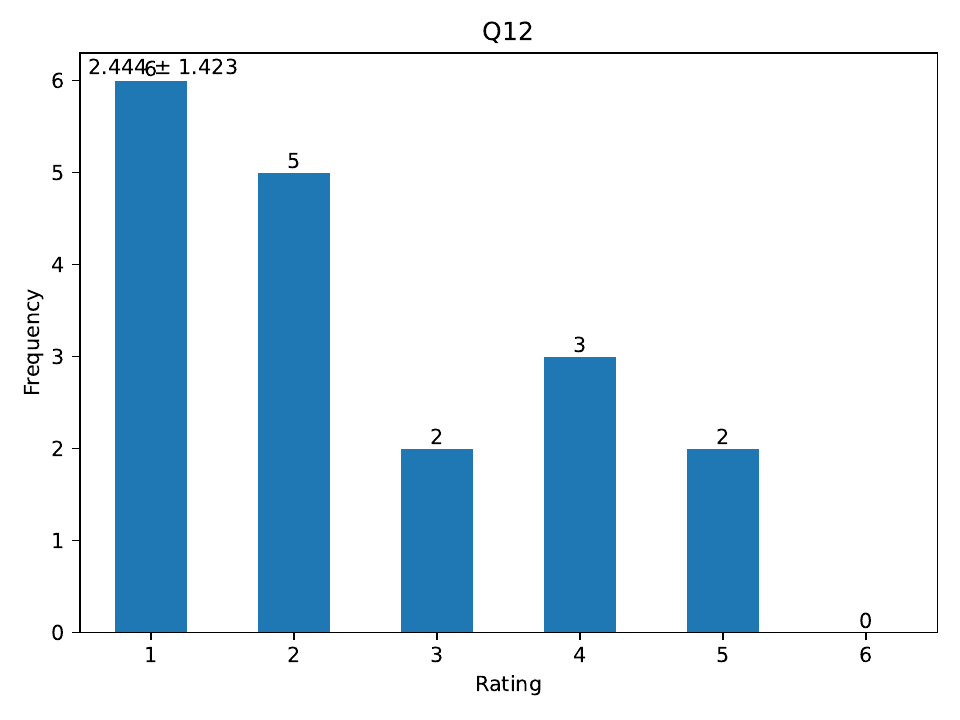}
\end{figure}
\FloatBarrier

\phantomsection\label{Q13}[Q13] Without changing your previous answer, how do you now define expressive power? *

\begin{itemize}[noitemsep,topsep=0pt,parsep=0pt,partopsep=0pt]
    \item Open text field
\end{itemize}

\begin{figure}[!ht]
    \centering
    \includegraphics[width=0.45\textwidth]{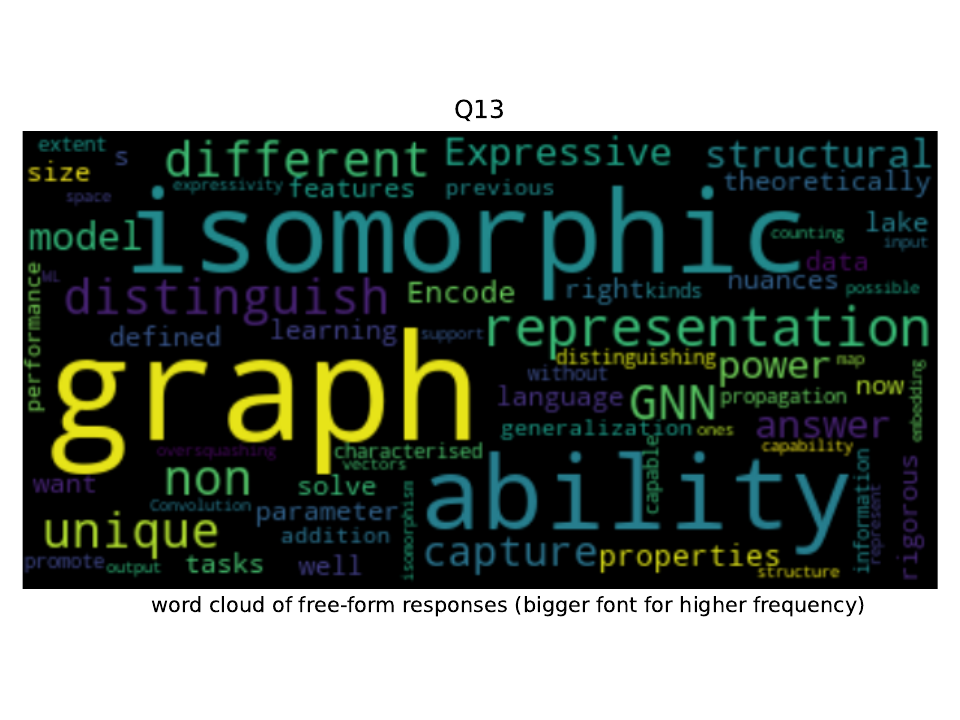}
\end{figure}
\FloatBarrier

\phantomsection\label{Q14}[Q14] Do you have additional thoughts in response to the questions in this section?

\textit{This can include justifications of your responses or a lack of clarity on any questions. Please be mindful not to bring up any identifying or sensitive information about yourself or third-parties.}

\begin{itemize}[noitemsep,topsep=0pt,parsep=0pt,partopsep=0pt]
    \item Open text field
\end{itemize}

\clearpage

\subsection{Operationalization Questions}

\phantomsection\label{Q15}[Q15] Please rank the following ways of measuring expressive power in order of how commonly you think they are employed in relevant literature: *

\textit{6 means most commonly, 1 means least commonly.}

\begin{enumerate}[label=\alph*,noitemsep,topsep=0pt,parsep=0pt,partopsep=0pt]
    \item Comparing the non-isomorphic graphs that an architecture can theoretically distinguish to those that the WL test (or higher-order variants) can distinguish
    \item Characterizing the ability of an architecture to theoretically solve various combinatorial problems (e.g., subgraph detection, graph coloring, minimum vertex cover)
    \item Describing the complexity of logical queries that an architecture can theoretically represent
    \item Describing the invariance and equivariance properties of an architecture
    \item Awareness of an architecture of simplicial and cell complexes
    \item Comparing the test performance of an instantiation of an architecture on a benchmark to the performance of models with previously-proposed architectures
\end{enumerate}

\clearpage

\begin{figure}[!ht]
    \centering
    \includegraphics[width=0.45\textwidth]{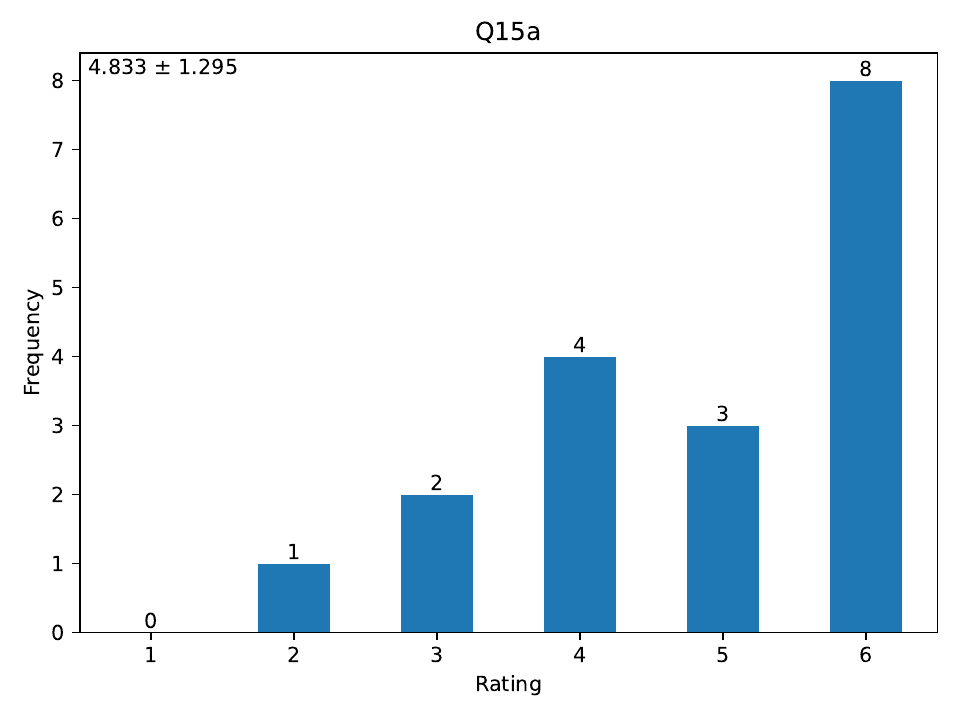}
    \includegraphics[width=0.45\textwidth]{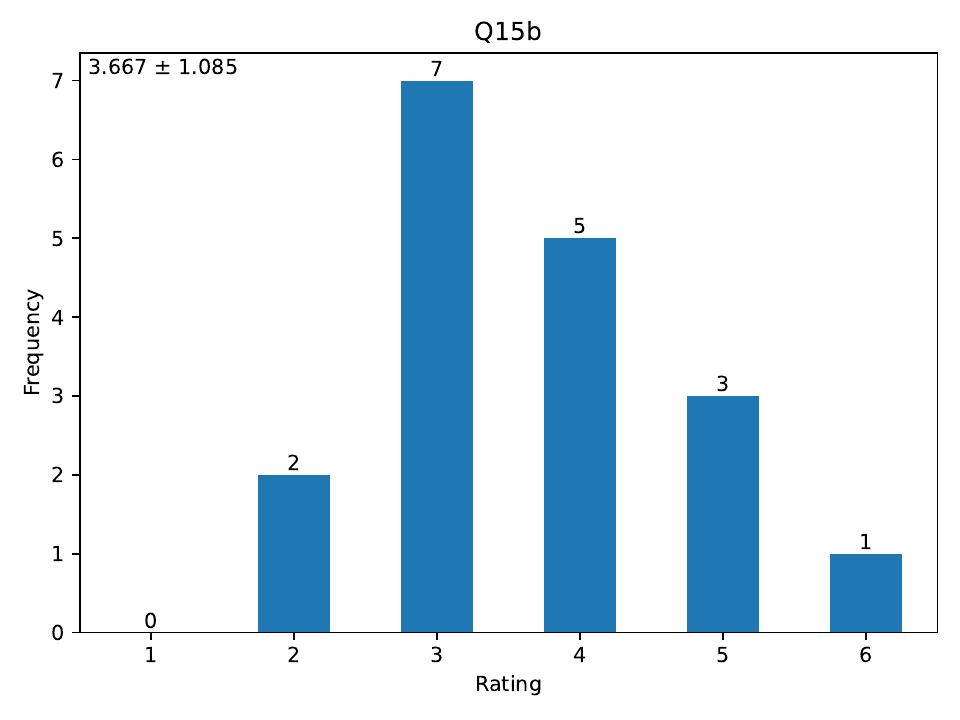}
\end{figure}
\FloatBarrier

\begin{figure}[!ht]
    \centering
    \includegraphics[width=0.45\textwidth]{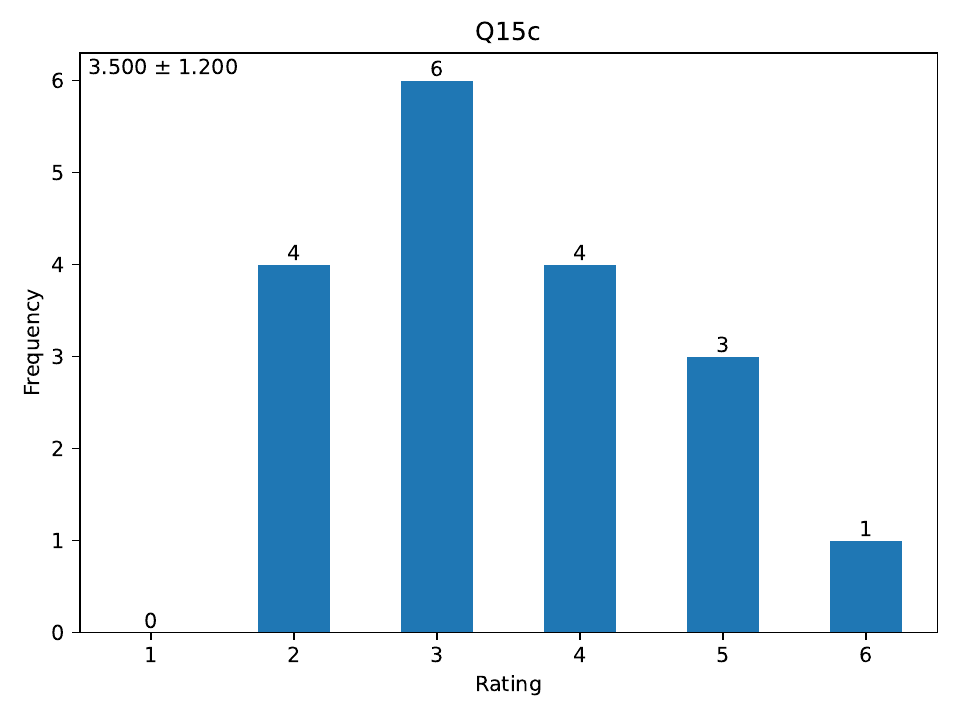}
    \includegraphics[width=0.45\textwidth]{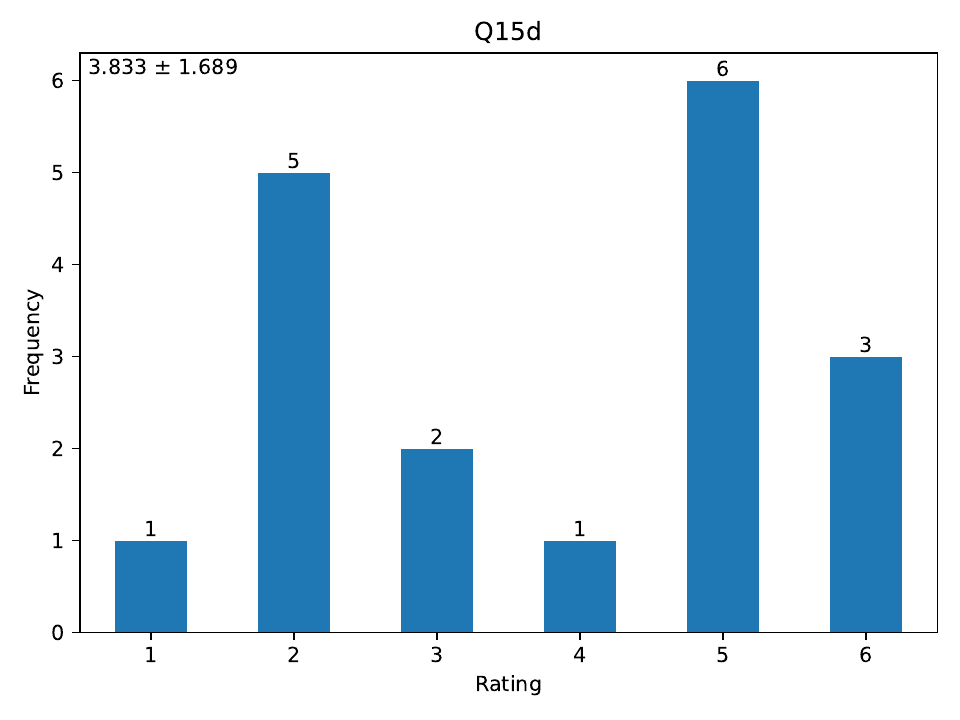}
\end{figure}
\FloatBarrier

\begin{figure}[!ht]
    \centering
    \includegraphics[width=0.45\textwidth]{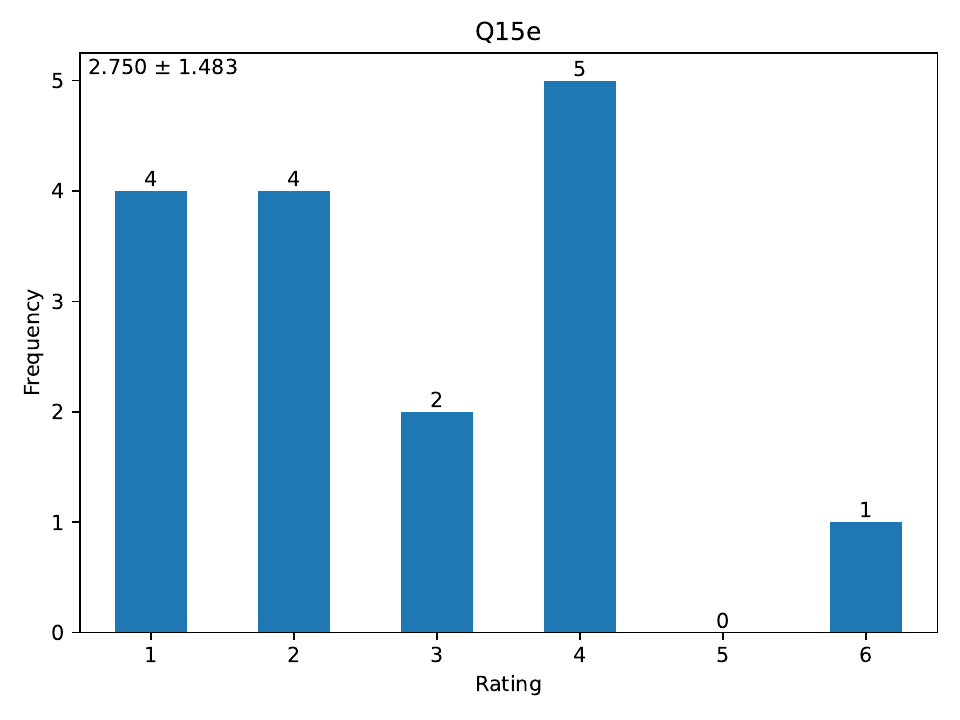}
    \includegraphics[width=0.45\textwidth]{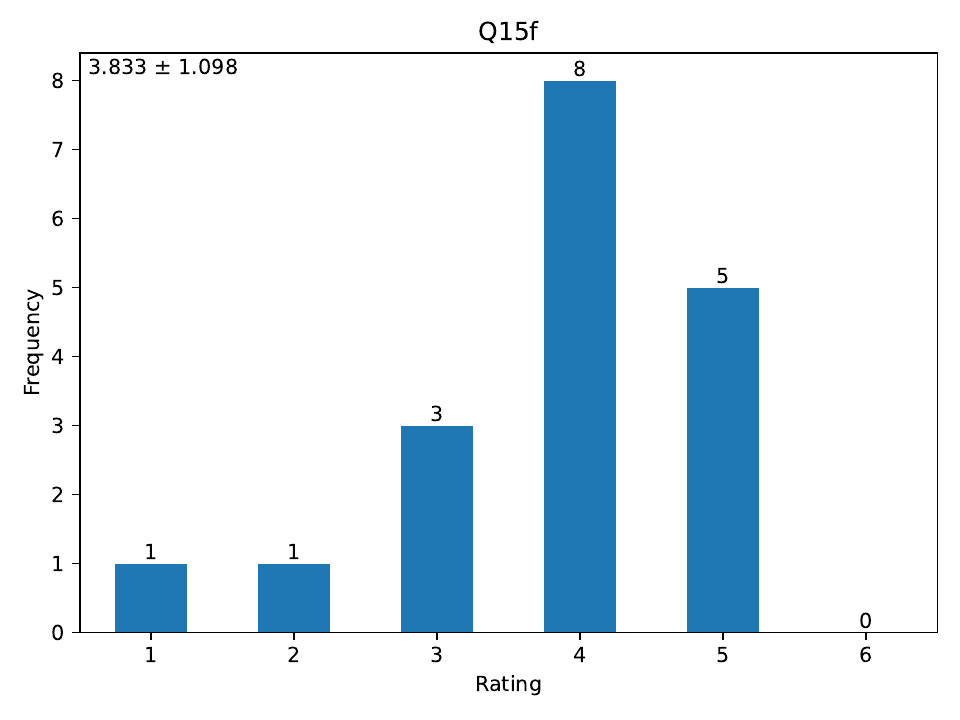}
\end{figure}
\FloatBarrier

\clearpage

\phantomsection\label{Q16}[Q16] Please list additional ways of measuring expressive power with which you are familiar (if any).

\begin{figure}[!ht]
    \centering
    \includegraphics[width=0.45\textwidth]{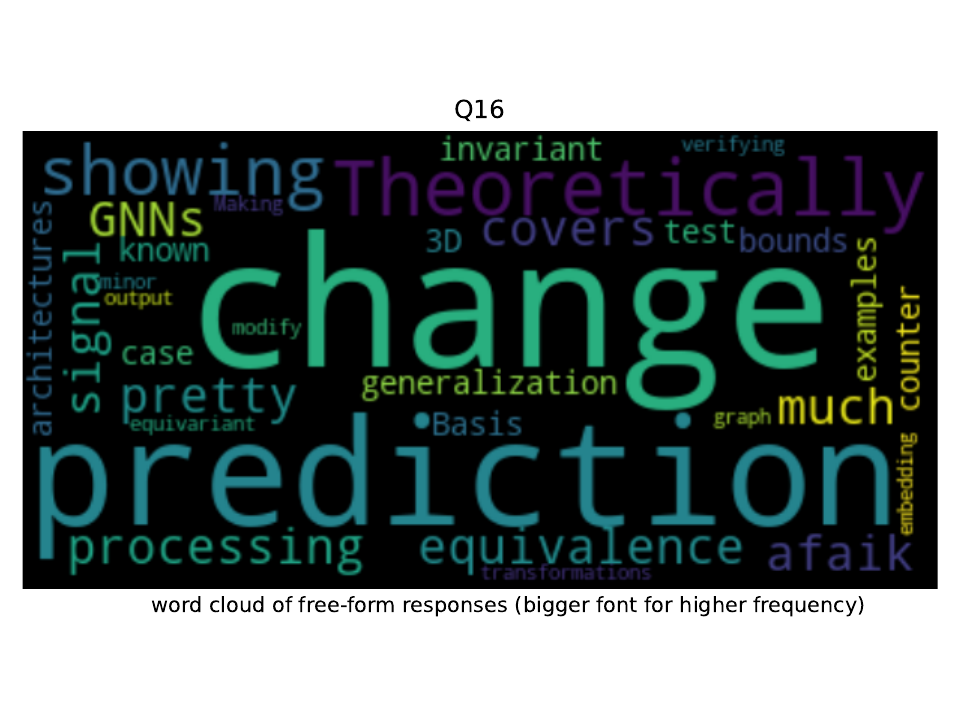}
\end{figure}
\FloatBarrier

\phantomsection\label{Q17}[Q17] How much do papers use real-world applications of graph machine learning to motivate methods of measuring expressive power? *

\textit{6 (always use) means all measurements of expressive power used by papers are inspired by real-world applications of graph machine learning. 1 (never use) means no measurements of expressive power used by papers are inspired by real-world applications.}

\begin{tabular}{ccccccc}
1 & 2 & 3 & 4 & 5 & 6  \\
$\ocircle$ & $\ocircle$ & $\ocircle$ & $\ocircle$ & $\ocircle$ & $\ocircle$
\end{tabular}

\begin{figure}[!ht]
    \centering
    \includegraphics[width=0.45\textwidth]{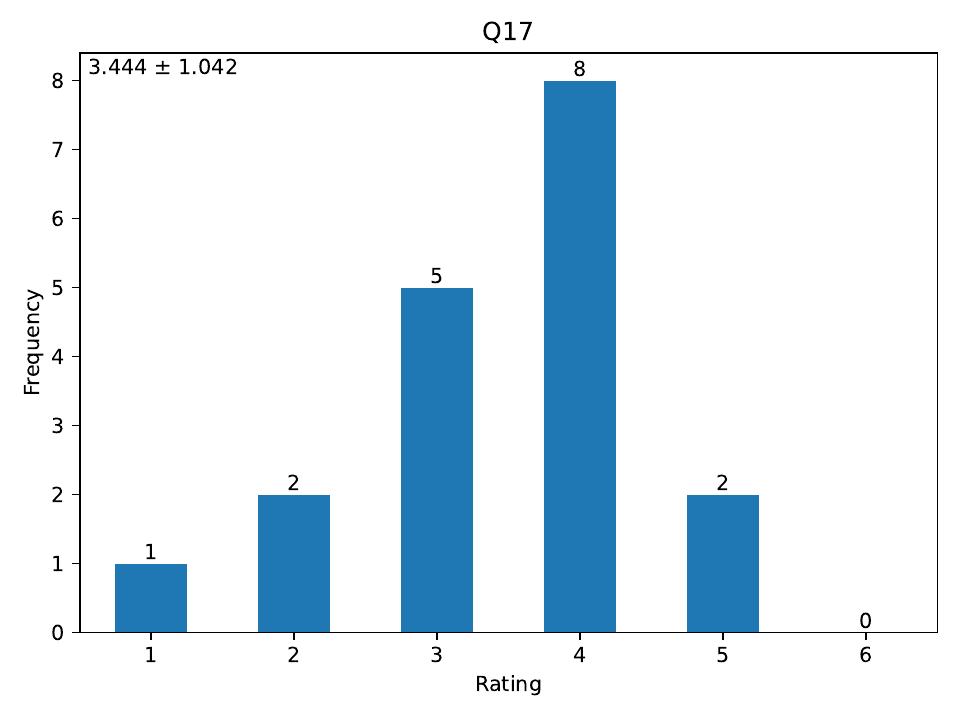}
\end{figure}
\FloatBarrier

\clearpage

\phantomsection\label{Q18}[Q18] How informative are different ways of measuring expressive power for predicting the performance of graph machine learning models on real-world applications? * 

\textit{6 (informative) means that measurements of expressive power are the most useful quantity for predicting model performance on real-world applications. In contrast, 1 (not informative) means that measurements of expressive power provide no useful information for predicting model performance on real-world applications.}

\begin{tabular}{ccccccc}
1 & 2 & 3 & 4 & 5 & 6  \\
$\ocircle$ & $\ocircle$ & $\ocircle$ & $\ocircle$ & $\ocircle$ & $\ocircle$
\end{tabular}

\begin{figure}[!ht]
    \centering
    \includegraphics[width=0.45\textwidth]{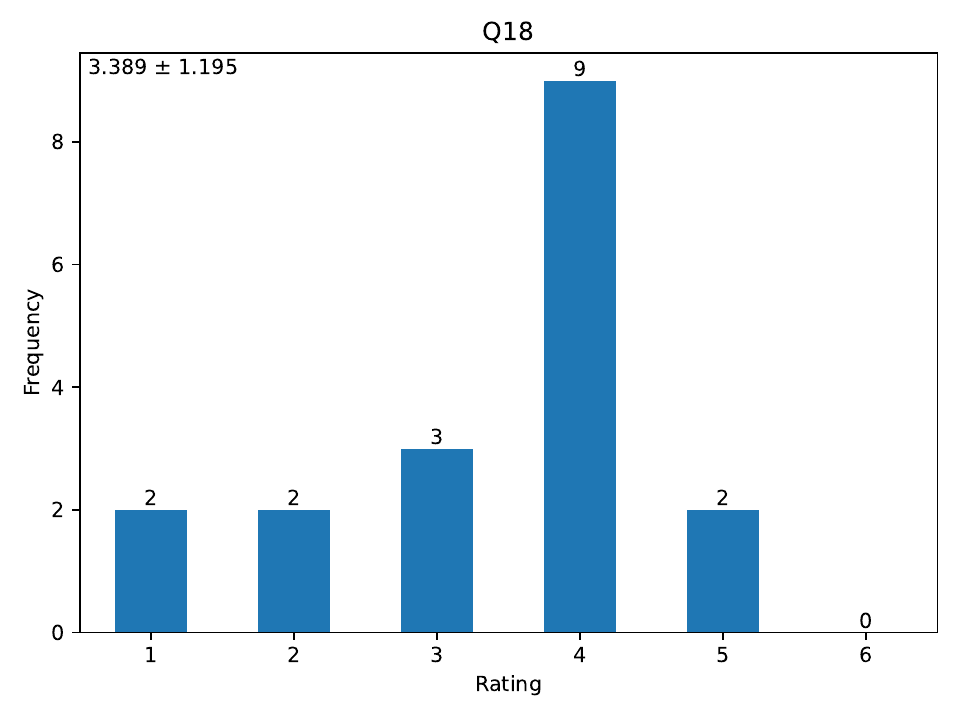}
\end{figure}
\FloatBarrier

\phantomsection\label{Q19}[Q19] Do you have additional thoughts in response to the questions in this section?

\textit{This can include justifications of your responses or a lack of clarity on any questions. Please be mindful not to bring up any identifying or sensitive information about yourself or third-parties.}

\begin{itemize}[noitemsep,topsep=0pt,parsep=0pt,partopsep=0pt]
    \item Open text field
\end{itemize}

\subsection{Weisfeiler-Leman Isomorphism Test Introduction}

We now focus on a specific method by which the measurement of expressive power is operationalized: via comparison to the WL test. In this method, we compare the non-isomorphic graphs that an architecture can theoretically distinguish to those that the WL test (or higher-order variants) can distinguish.

For example, we can say that Graph Convolutional Network (GCN) is at most as expressive as the 1-WL test (i.e., GCN can theoretically distinguish as many non-isomorphic graphs as the 1-WL test can).

\clearpage

\phantomsection\label{Q20}[Q20] Are you familiar with the WL test? * 
\begin{itemize}[noitemsep,topsep=0pt,parsep=0pt,partopsep=0pt]
    \item[$\ocircle$] Yes
    \item[$\ocircle$] No \textcolor{red}{[if selected, survey skips next section]}
\end{itemize}

\begin{figure}[!ht]
    \centering
    \includegraphics[width=0.45\textwidth]{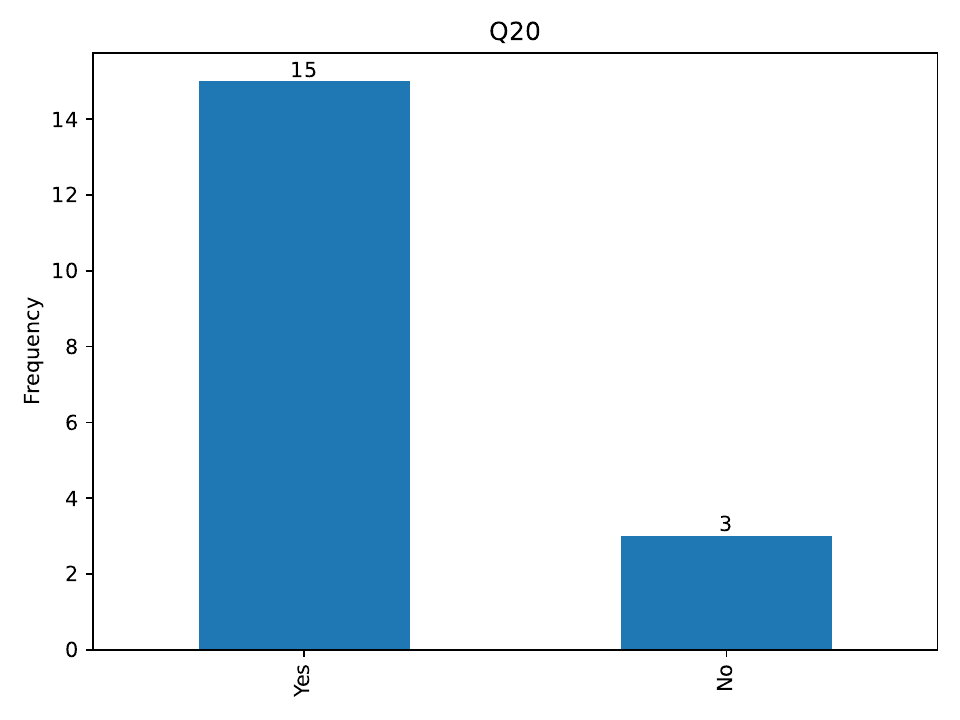}
\end{figure}
\FloatBarrier

\subsection{WL Test Questions}

\phantomsection\label{Q21}[Q21] How much do theoretical results of expressive power via comparison to the WL test (or higher order variants) appear convincing? * 

\textit{6 (convincing) means that such proofs generally employ realistic settings and assumptions about graph neural networks, as well as solid reasoning. In contrast, 1 (not convincing) means that such proofs generally use unrealistic or debatable settings or assumptions, or flawed reasoning.}

\begin{tabular}{ccccccc}
1 & 2 & 3 & 4 & 5 & 6  \\
$\ocircle$ & $\ocircle$ & $\ocircle$ & $\ocircle$ & $\ocircle$ & $\ocircle$
\end{tabular}

\begin{figure}[!ht]
    \centering
    \includegraphics[width=0.45\textwidth]{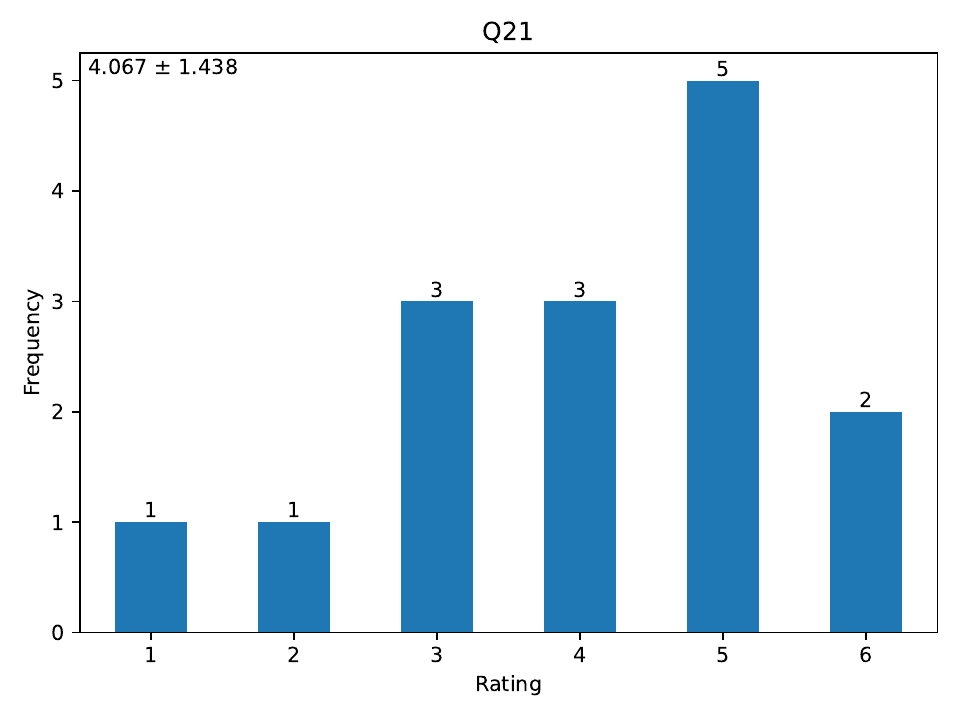}
\end{figure}
\FloatBarrier

\clearpage

\phantomsection\label{Q22}[Q22] Does the WL test help provide a useful upper bound on the accuracy of any task? * 

\begin{itemize}[noitemsep,topsep=0pt,parsep=0pt,partopsep=0pt]
    \item[$\ocircle$] 5 = Always
    \item[$\ocircle$] 4 = Almost always
    \item[$\ocircle$] 3 = Sometimes
    \item[$\ocircle$] 2 = Almost never
    \item[$\ocircle$] 1 = Never
\end{itemize}

\begin{figure}[!ht]
    \centering
    \includegraphics[width=0.45\textwidth]{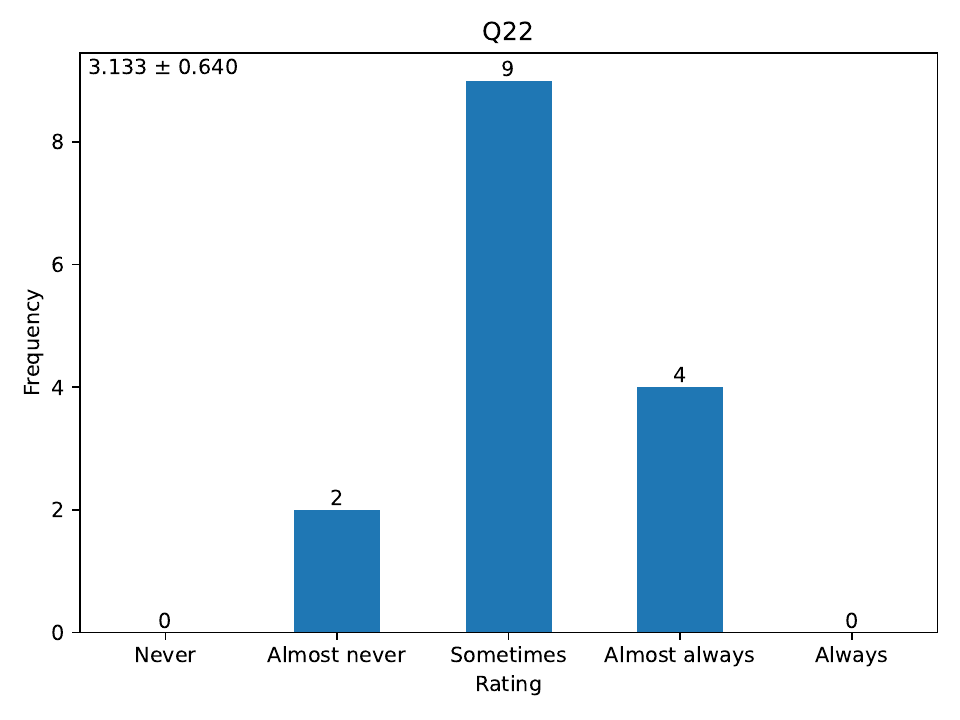}
\end{figure}
\FloatBarrier

\phantomsection\label{Q23}[Q23] Does the 1-WL test produce colorings or hashes that are sufficient to solve common graph learning tasks? * 

\begin{itemize}[noitemsep,topsep=0pt,parsep=0pt,partopsep=0pt]
    \item[$\ocircle$] 5 = Always
    \item[$\ocircle$] 4 = Almost always
    \item[$\ocircle$] 3 = Sometimes
    \item[$\ocircle$] 2 = Almost never
    \item[$\ocircle$] 1 = Never
\end{itemize}

\begin{figure}[!ht]
    \centering
    \includegraphics[width=0.45\textwidth]{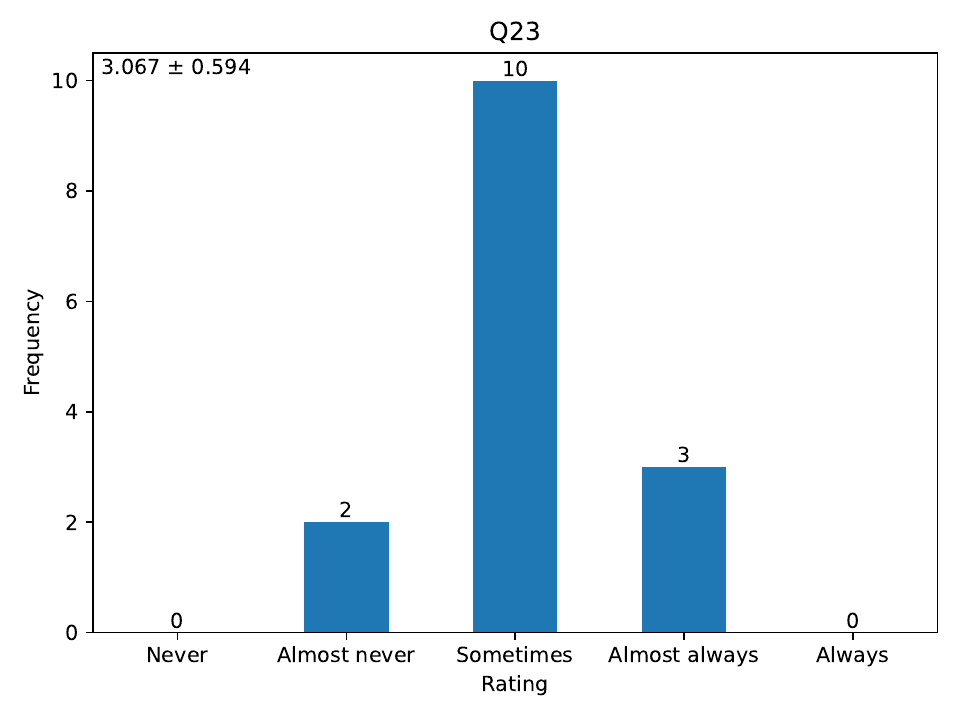}
\end{figure}
\FloatBarrier

\clearpage

\phantomsection\label{Q24}[Q24] \textbf{Do} graph neural networks in practice learn representations that align with the colorings or hashes produced by the WL test? * 

\begin{itemize}[noitemsep,topsep=0pt,parsep=0pt,partopsep=0pt]
    \item[$\ocircle$] 5 = Always
    \item[$\ocircle$] 4 = Almost always
    \item[$\ocircle$] 3 = Sometimes
    \item[$\ocircle$] 2 = Almost never
    \item[$\ocircle$] 1 = Never
\end{itemize}

\begin{figure}[!ht]
    \centering
    \includegraphics[width=0.45\textwidth]{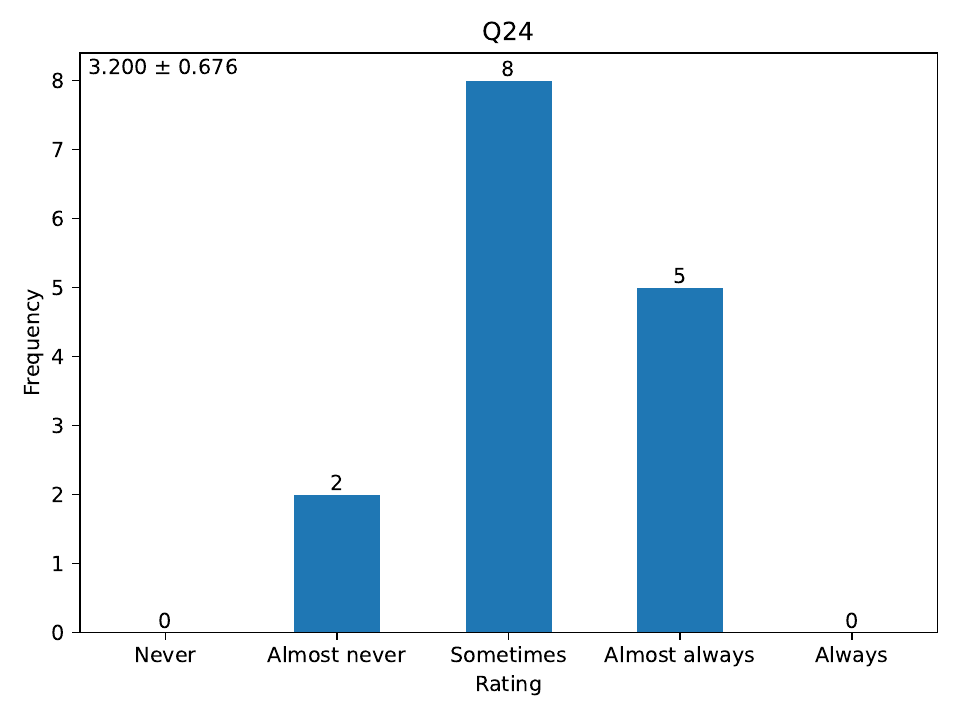}
\end{figure}
\FloatBarrier

\phantomsection\label{Q25}[Q25] \textbf{Can} graph neural networks ever learn representations that align with the colorings or hashes produced by the WL test? * 

\begin{itemize}[noitemsep,topsep=0pt,parsep=0pt,partopsep=0pt]
    \item[$\ocircle$] 5 = Always
    \item[$\ocircle$] 4 = Almost always
    \item[$\ocircle$] 3 = Sometimes
    \item[$\ocircle$] 2 = Almost never
    \item[$\ocircle$] 1 = Never
\end{itemize}

\begin{figure}[!ht]
    \centering
    \includegraphics[width=0.45\textwidth]{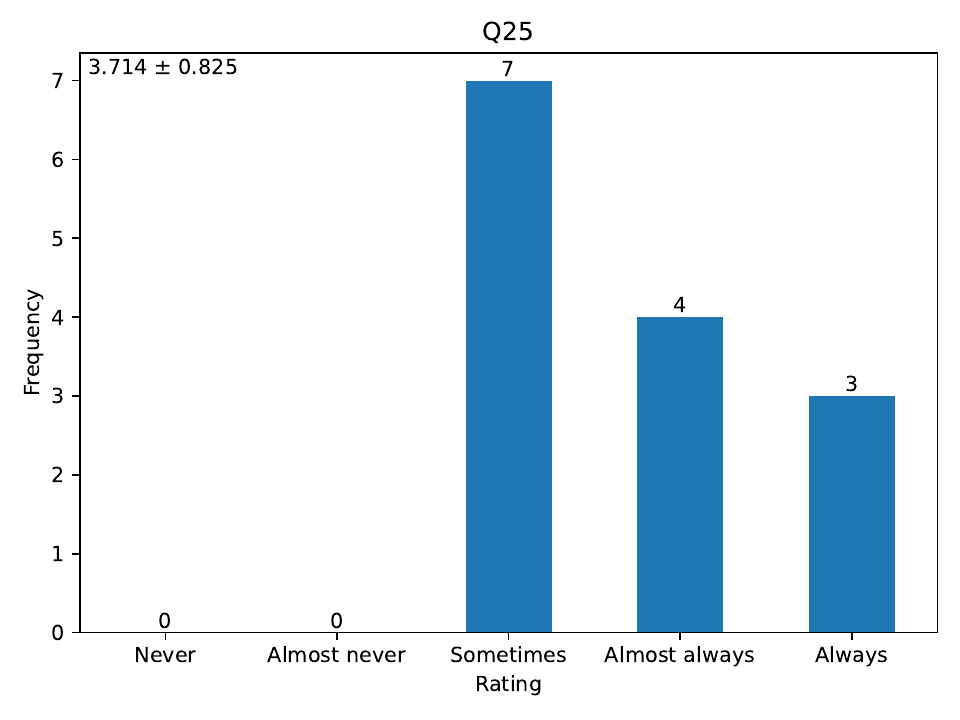}
\end{figure}
\FloatBarrier

\phantomsection\label{Q26}[Q26] Do you have additional thoughts in response to the questions in this section?

\textit{This can include justifications of your responses or a lack of clarity on any questions. Please be mindful not to bring up any identifying or sensitive information about yourself or third-parties.}

\begin{itemize}[noitemsep,topsep=0pt,parsep=0pt,partopsep=0pt]
    \item Open text field
\end{itemize}

\subsection{In-Practice Questions}

\phantomsection\label{Q27}[Q27] Please rank the following criteria for graph neural networks by how you prioritize them in practice. *

\textit{1 = least, 6 = most}

\begin{enumerate}[label=\alph*,noitemsep,topsep=0pt,parsep=0pt,partopsep=0pt]
    \item Expressive power
    \item Generalization
    \item Efficiency
    \item Fairness
    \item Privacy
    \item Robustness
\end{enumerate}

\begin{figure}[!ht]
    \centering
    \includegraphics[width=0.35\textwidth]{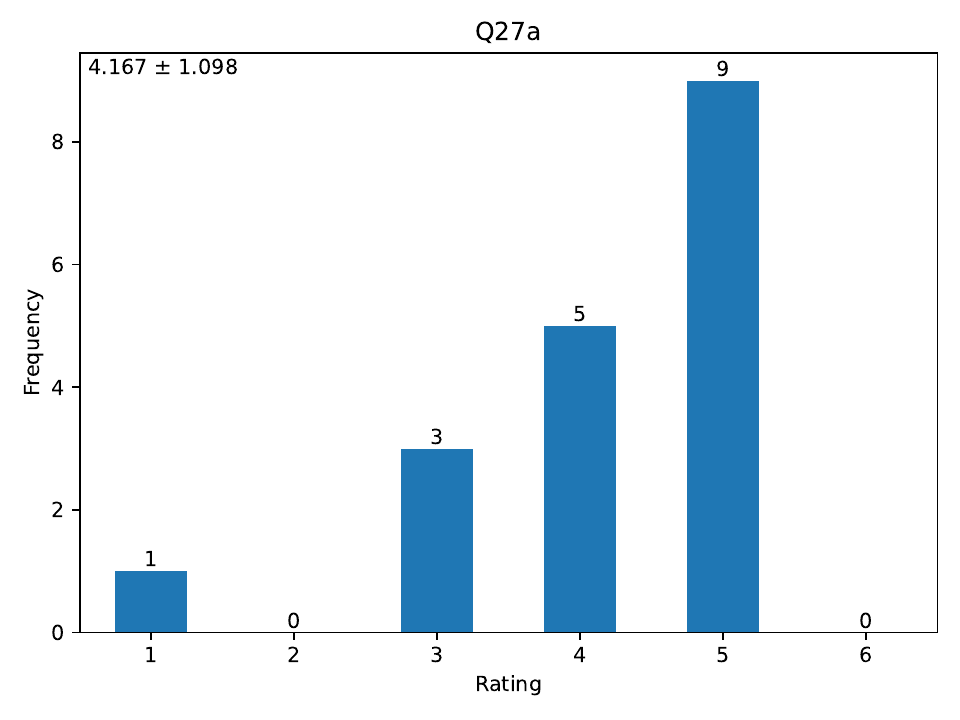}
    \includegraphics[width=0.35\textwidth]{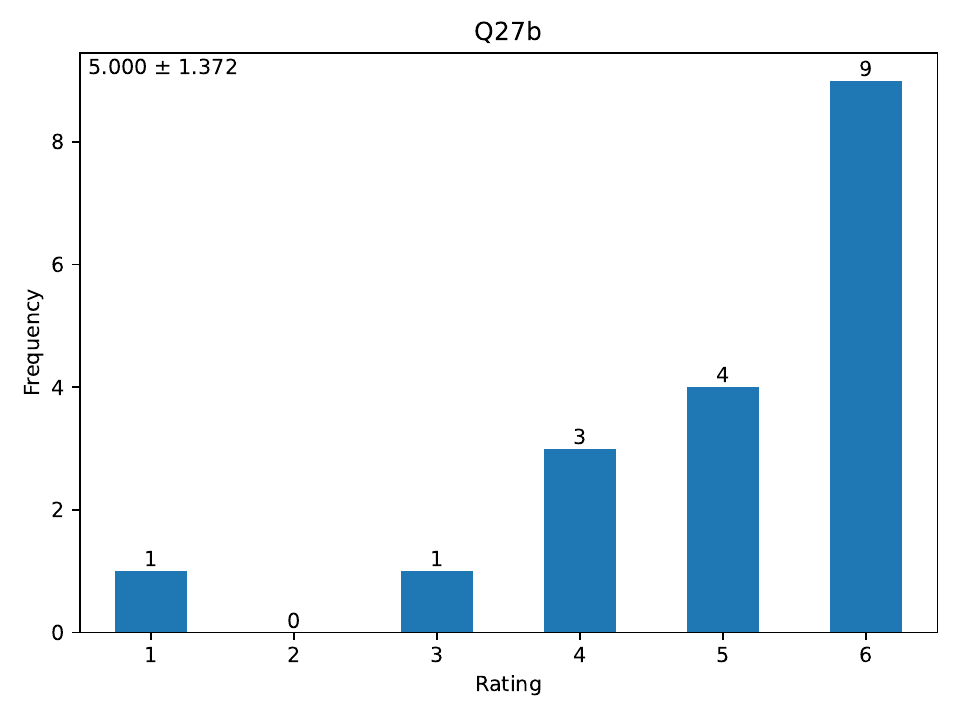}
\end{figure}
\FloatBarrier

\begin{figure}[!ht]
    \centering
    \includegraphics[width=0.35\textwidth]{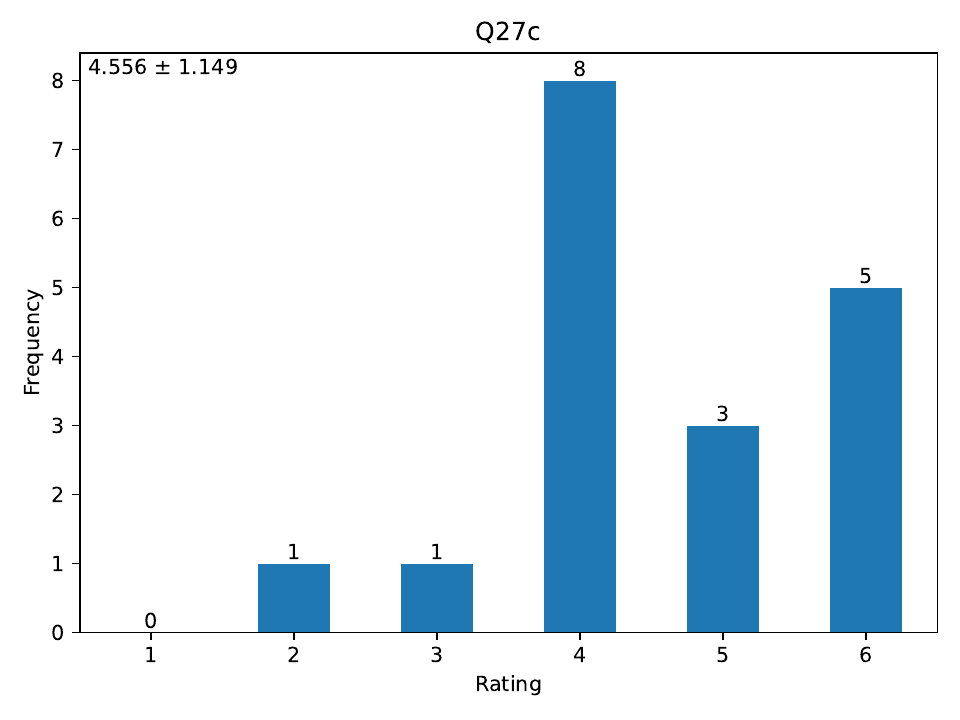}
    \includegraphics[width=0.35\textwidth]{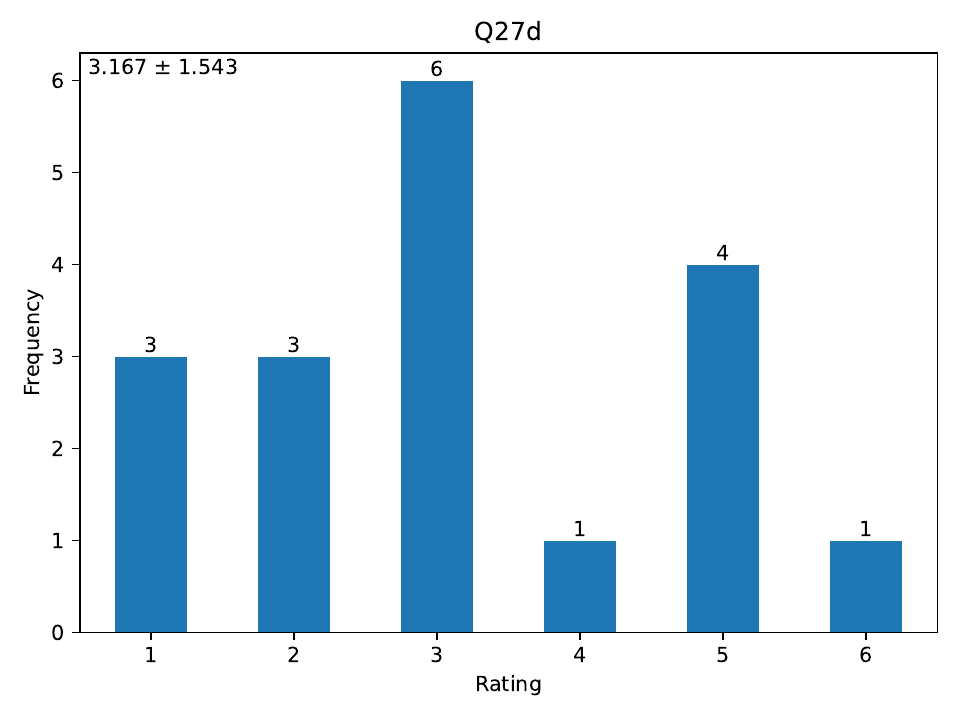}
\end{figure}
\FloatBarrier

\begin{figure}[!ht]
    \centering
    \includegraphics[width=0.35\textwidth]{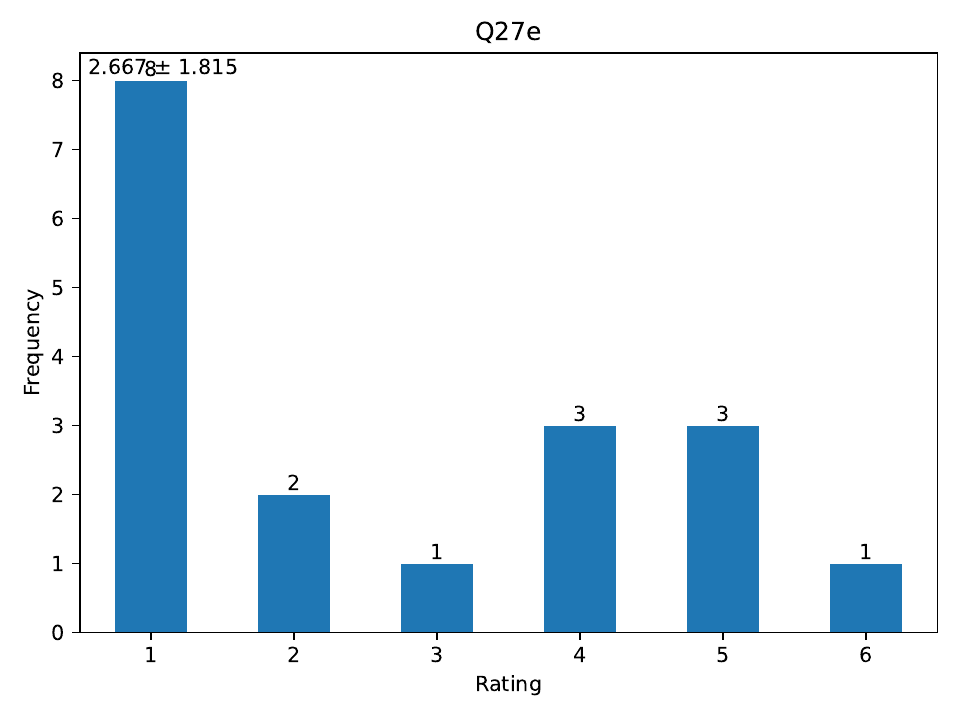}
    \includegraphics[width=0.35\textwidth]{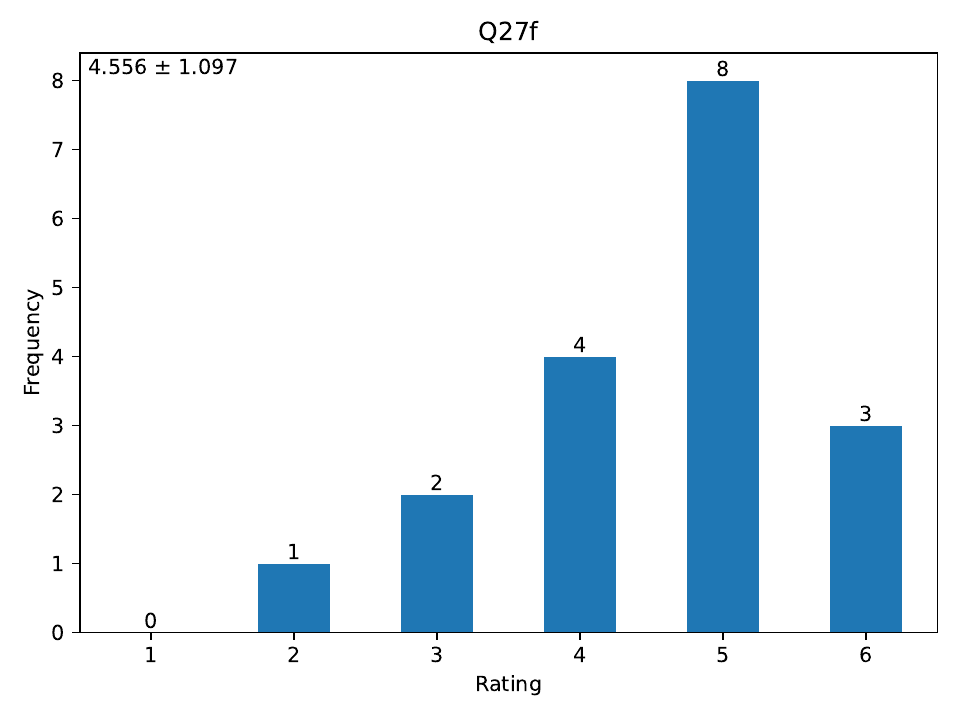}
\end{figure}
\FloatBarrier

\clearpage

\phantomsection\label{Q28}[Q28] How much do you use expressive power to
inform architecture design? *

\textit{6 (always) means that you always design new graph neural network architectures with expressive power in mind. 1 (never) means that you never consider expressive power when designing new architectures.}

\begin{tabular}{ccccccc}
1 & 2 & 3 & 4 & 5 & 6  \\
$\ocircle$ & $\ocircle$ & $\ocircle$ & $\ocircle$ & $\ocircle$ & $\ocircle$
\end{tabular}

\begin{figure}[!ht]
    \centering
    \includegraphics[width=0.45\textwidth]{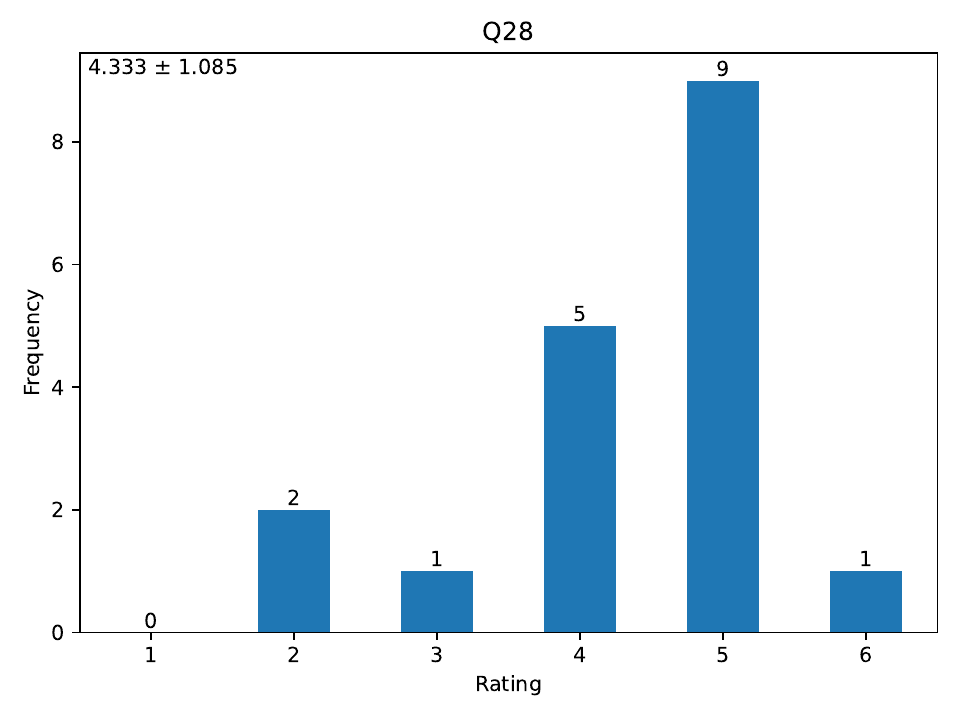}
\end{figure}
\FloatBarrier

\phantomsection\label{Q29}[Q29] How much do you use expressive power to inform benchmark design? *

\textit{6 (always) means you always design new benchmarks with expressive power in mind. 1 (never) means that you never consider expressive power when designing new benchmarks.}

\begin{tabular}{ccccccc}
1 & 2 & 3 & 4 & 5 & 6  \\
$\ocircle$ & $\ocircle$ & $\ocircle$ & $\ocircle$ & $\ocircle$ & $\ocircle$
\end{tabular}

\begin{figure}[!ht]
    \centering
    \includegraphics[width=0.45\textwidth]{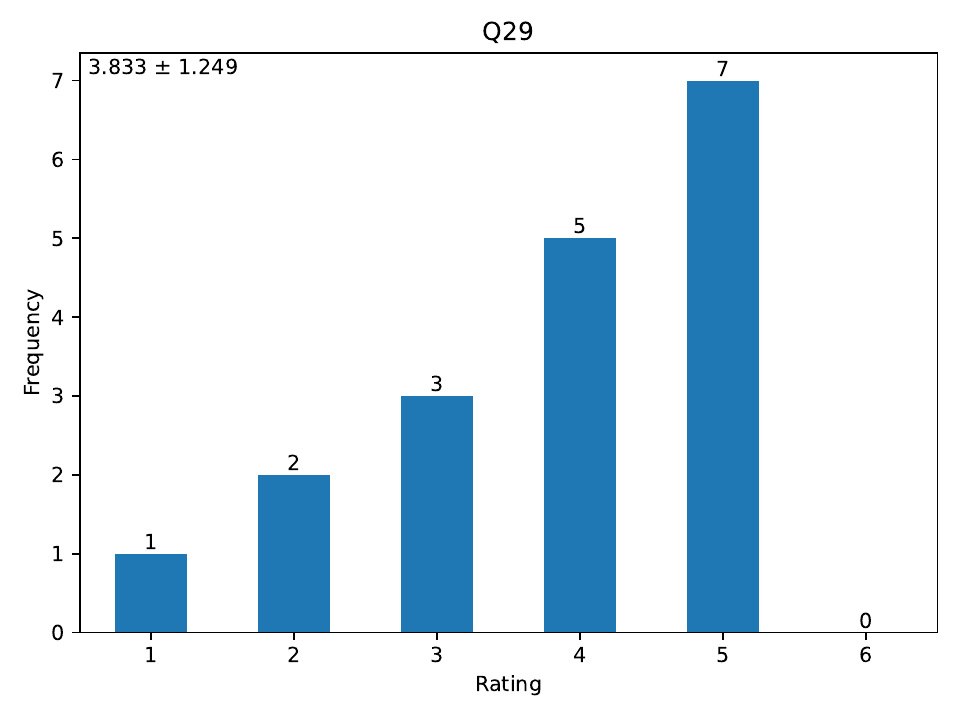}
\end{figure}
\FloatBarrier

\phantomsection\label{Q30}[Q30] Do you have additional thoughts in response to the questions in this section?

\textit{This can include justifications of your responses or a lack of clarity on any questions. Please be mindful not to bring up any identifying or sensitive information about yourself or third-parties.}

\begin{itemize}[noitemsep,topsep=0pt,parsep=0pt,partopsep=0pt]
    \item Open text field
\end{itemize}

\clearpage

\subsection{Feedback}

Thanks again for your participation in our survey!

\phantomsection\label{Q31}[Q31] Do you have any comments or feedback on the questions in this survey?

\textit{Please be mindful not to bring up any identifying or sensitive information about yourself or third-parties.}

\begin{itemize}[noitemsep,topsep=0pt,parsep=0pt,partopsep=0pt]
    \item Open text field
\end{itemize} 

\clearpage

\section{Graph tasks and benchmarks}
\label{sec:graph-tasks-benchmarks}

\paragraph{Tasks vs. benchmarks} A \textit{task} is a graph-related skill or competency that we want a GNN to demonstrate in the context of a specific input–output format. In contrast, a \textit{benchmark} attempts to measure performance on a task by grounding it in a graph domain and instantiating it with a concrete dataset and evaluation metric \cite{Bowman2021WhatWI}.

\paragraph{Benchmark validity} There are many aspects to benchmark validity (including but not limited to):
\begin{itemize}
    \item External validity \cite{Olteanu2019SocialDB}:
     \begin{itemize}
        \item Ecological validity: ``to what extent does an artificial situation (constrained social media platform [or synthetic datasets]) properly reflect a broader real-world phenomenon?'' (e.g.,  \cite{Palowitch2022GraphWorldFG})
        \item Temporal validity: ``to what extent do constructs change over time and invalidate previous conclusions''?
    \end{itemize}
    \item Construct validity \cite{jacobs2021}:
    \begin{itemize}
        \item Content validity: How well conceptualized is the task on which performance is being measured? How well-aligned is the benchmark with the task specification?
        \item Convergent validity: Does the benchmark actually measure what we care about?
        \item Discriminant validity: Does the benchmark measure things besides what we care about?
        \item Consequential validity: Considering societal implications, should the benchmark be used?
    \end{itemize} 
\end{itemize}

Benchmarks often have pitfalls that affect their measurement of real-world task performance \cite{gurukar2022benchmarking}. For example, benchmarks can have noise or poor data diversity. \cite{Ivanov2019UnderstandingIB} shows that popular graph ML benchmarks can consist of as low as 20\% non-isomorphic graphs. Benchmarks may also capture skewed substructure frequencies, and suffer from extreme sparsity and large numbers of isolated nodes due to lossy link observation. Benchmarks may also employ poorly-aligned or reductive performance metrics. Graph structure may not even be required to perform well on certain benchmarks. In many cases, we would like graph ML benchmarks to measure the extent to which GNNs can leverage meaningful combinations of node features and graph structure (e.g., to answer, are two graphs isomorphic?). To understand if benchmarks assess GNNs' ability to harness node features vs. graph structure, \cite{Liu2022TaxonomyOB} independently perturbs node features and graph structure, thereby extending partial-input baselines \cite{Poliak2018HypothesisOB} to the graph domain.

Benchmarks on which it is possible to achieve high accuracy without certain node features or parts of the graph structure (called ``reduced benchmarks'') are considered ``easy'' \cite{Feng2019MisleadingFO}; this is because reduced benchmarks suggest that a combination of node features and graph structure is not necessary to do well. However, for tasks where both features and structures are actually necessary, doing well on corresponding ``reduced benchmarks'' implies the existence of certain spurious correlations. Consider the task of natural language inference (NLI) (i.e., does a premise entail a hypothesis?): both the premise and hypothesis should ideally be necessary to solve the task, so model success on premise-only or hypothesis-only ``reduced benchmarks'' for NLI would suggest that the original benchmark has poor discriminant validity \cite{jacobs2021}. That is, it is unclear to what extent the benchmark measures the ability of a model to ``cheat'' using spurious correlations vs. actually perform the desired reasoning. But, just because a model performs well without graph structure, this doesn't entail that the model doesn't leverage graph structure when it is provided \cite{srikanth-rudinger-2022-partial}.

In contrast to ``easy'' benchmarks, benchmarks on which models fail without certain node features or parts of the graph structure are considered ``difficult.'' However, failures on the partial-input baselines proposed by \cite{Liu2022TaxonomyOB} do not indicate that models are not leveraging other spurious correlations (perhaps simple combinations of node features and graph structure). There are also infinitely many possible partial-input baselines. Consider the task of counting the number of size-4 maximal cliques comprising only purple nodes in a graph, as well as a benchmark for this task which has only size-4 maximal cliques. Furthermore, consider a GNN that can only count size-3 maximal cliques, i.e., triangles (but not 4-cliques or other larger cliques). With node color information (but without graph structure), the model may output the total number of purple nodes divided by 4 and perform poorly. With graph structure (but without node color information), the model may output 4 times the number of 3-cliques in the graph and perform poorly. However, with both node color information and graph structure, the model may output 4 times the number of 3-cliques comprising only purple nodes and achieve perfect performance on the benchmark, thereby deceiving us into thinking that it has learned to perform the task.

\clearpage

\section{Audited graph ML benchmarks}
\label{sec:benchmarks-overview}

\subsection{Benchmarks used in \S\ref{sec:benchmarking-expressive-power}}
\label{sec:gen-benchmarks-overview}

\begin{table}[!ht]
\small
\caption{Summary of benchmarks used in \S\ref{sec:benchmarking-expressive-power}.}
\label{tab:benchmarks}
\centering
\begin{adjustbox}{max width=\textwidth}
\begin{tabular}{llllr}
\toprule
{} &  Level &    Domain & Node features? &  $|{\mathcal E}_y|$ \\
\midrule
IMDB-BINARY   &  graph &    social &             No &               2 \\
IMDB-MULTI    &  graph &    social &             No &               3 \\
REDDIT-BINARY &  graph &    social &             No &               2 \\
PROTEINS      &  graph &       bio &            Yes &               2 \\
PTC\_MR        &  graph &       bio &            Yes &               2 \\
MUTAG         &  graph &       bio &            Yes &               2 \\
\midrule
Cora          &   node &  citation &            Yes &               7 \\
CiteSeer      &   node &  citation &            Yes &               6 \\
PubMed        &   node &  citation &            Yes &               3 \\
\bottomrule
\end{tabular}
\end{adjustbox}
\end{table}

\FloatBarrier

Our usage of these benchmarks complies with their license (where applicable); not all the benchmarks have a license. While these benchmarks are widely used, we did not obtain explicit consent from any data subjects whose data the benchmarks may contain. To the best of our knowledge (via manual sampling and inspection), the benchmarks do not contain any personally identifiable information or offensive content.

\subsection{Benchmarks used in \S\ref{sec:consequential-validity}}
\label{sec:ethics-benchmarks-overview}

The \textsc{Credit} network consists of 30,000 nodes representing individuals, with edges between them indicating similar spending and payment patterns.  Each node has 13 features (e.g.,  education, credit history), with an average degree of $95.79 \pm 85.88$. The corresponding task is to predict whether an individual will default on their credit card payment, and the sensitive groups are those 25 years old or younger and those above the age of 25. \textsc{German} network comprises 1,000 nodes representing clients in a German bank who are connected if they have similar credit accounts. Each node has 27 features (e.g.,  loan amount, account-related features), with an average degree of $44.48 \pm 26.51$. The corresponding task is to predict whether a client has good or bad credit risk, and the groups are men and women. For both benchmarks, we do not include group membership as a feature.

To the best of our knowledge (via manual sampling and inspection), neither benchmark contains personally identifiable information or offensive content. We use \cite{agarwal2021unified}'s data and data loading code\footnote{\url{https://github.com/chirag126/nifty}} in accordance with its MIT license. We refrain from using the \textsc{Recidivism} network from \cite{agarwal2021unified} so as not to support the development of carceral technology.

\clearpage

\section{Non-distinguishable graph pairs from MUTAG}
\label{sec:non-distinguishable-mutag}

\begin{figure}[!ht]
    \centering

\includegraphics[width=\textwidth]{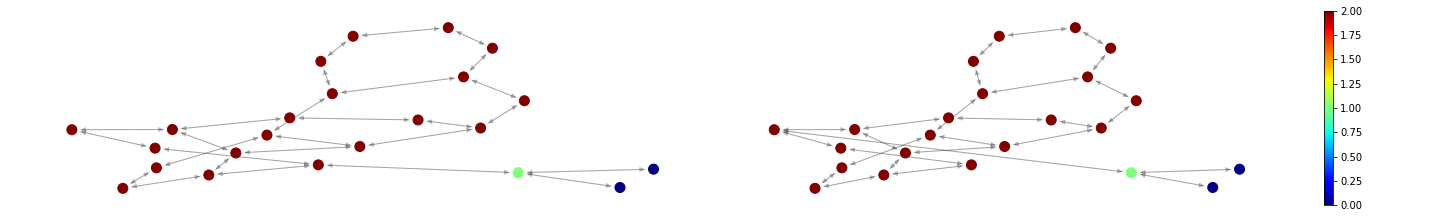}

\includegraphics[width=\textwidth]{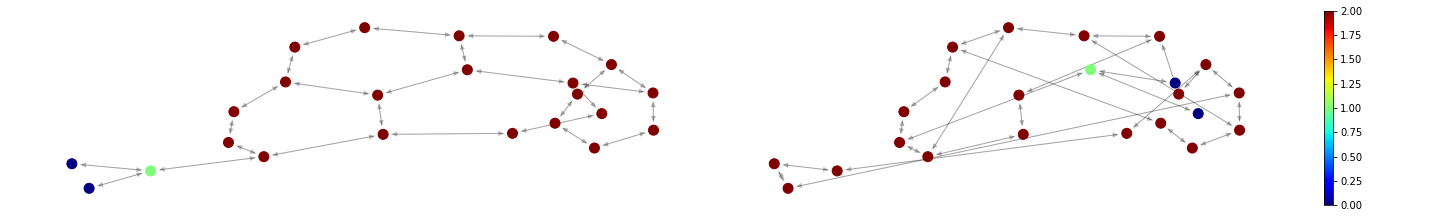}

\includegraphics[width=\textwidth]{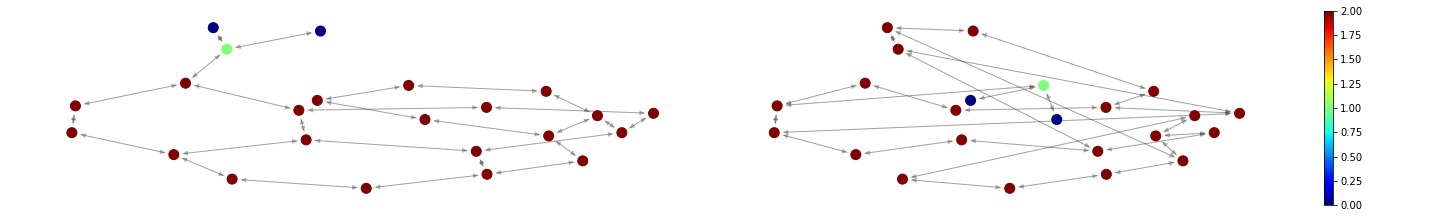}

\includegraphics[width=\textwidth]{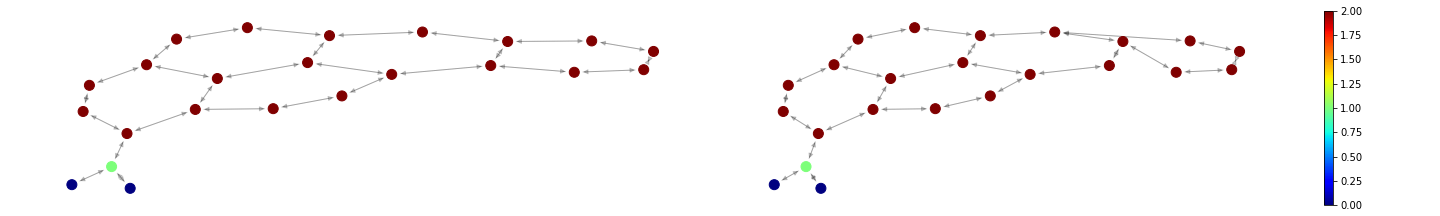}

\includegraphics[width=\textwidth]{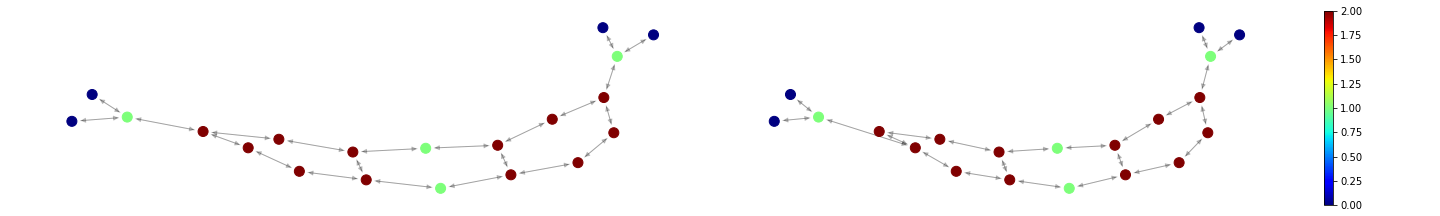}

\includegraphics[width=\textwidth]{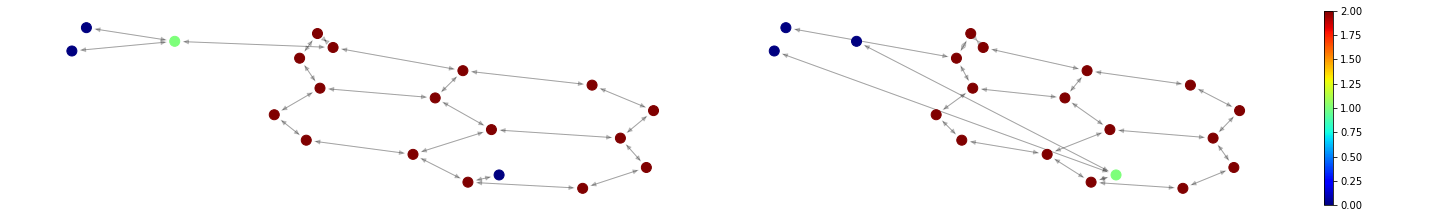}

\includegraphics[width=\textwidth]{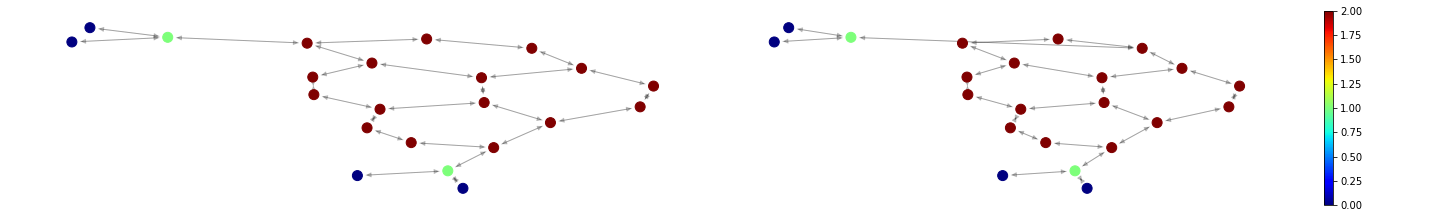}

    \caption{1-WL non-distinguishable graph pairs from the MUTAG benchmark (after three iterations).}
    \label{fig:non-distinguishable-mutag}
\end{figure}

\clearpage

\section{Adjusted mutual information (AMI)}
\label{sec:app-ami}

\begin{figure}[!ht]
    \centering

\includegraphics[width=0.35\textwidth]{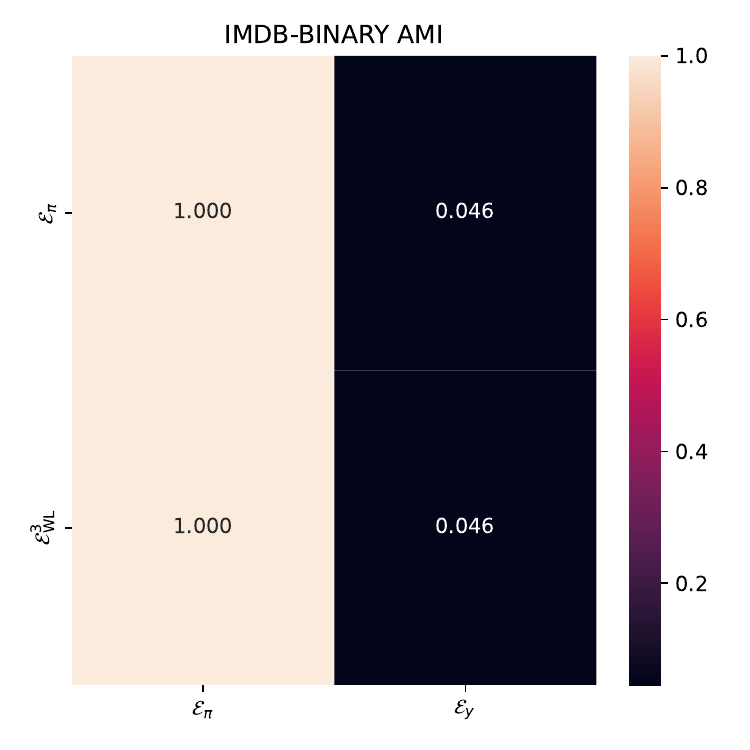}
\includegraphics[width=0.35\textwidth]{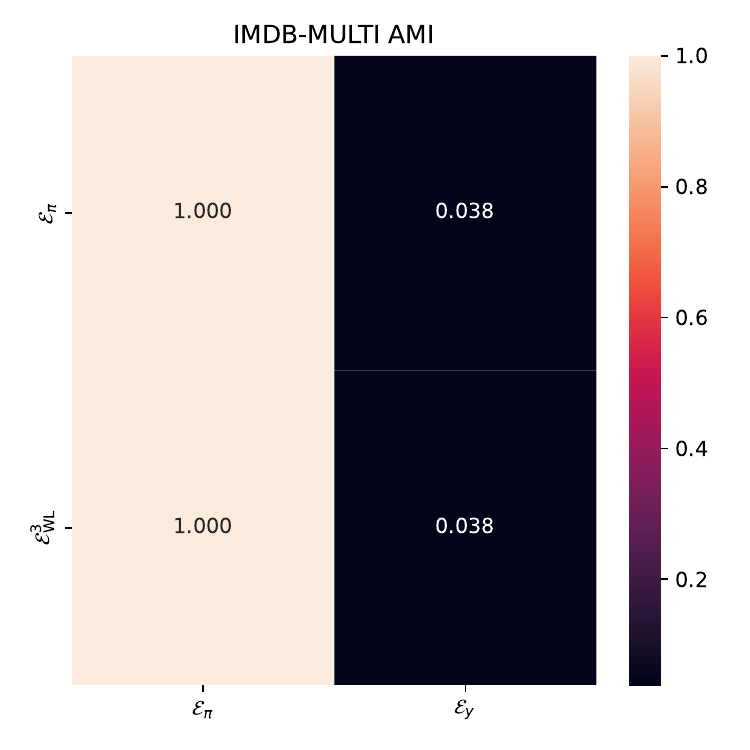}
\includegraphics[width=0.35\textwidth]{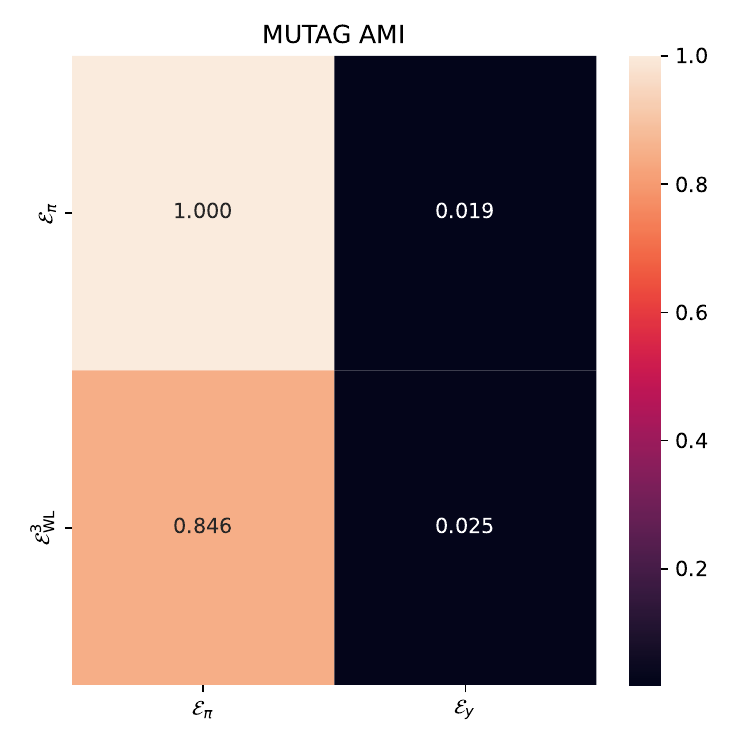}
\includegraphics[width=0.35\textwidth]{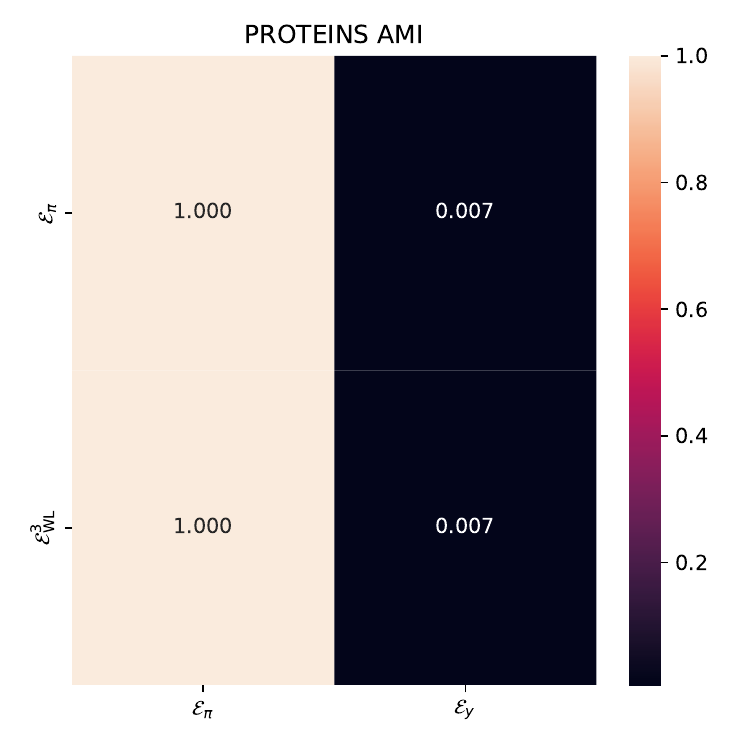}
\includegraphics[width=0.35\textwidth]{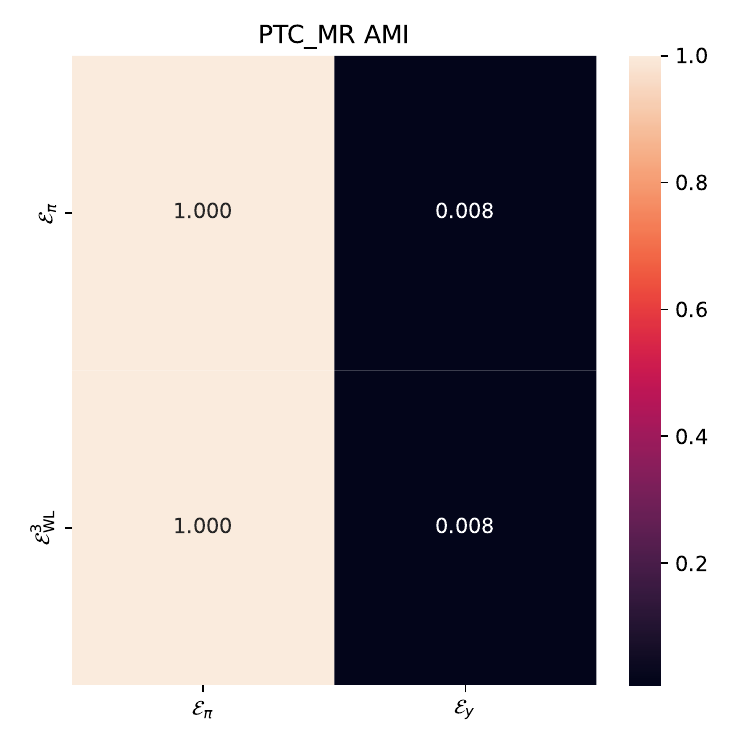}
\includegraphics[width=0.35\textwidth]{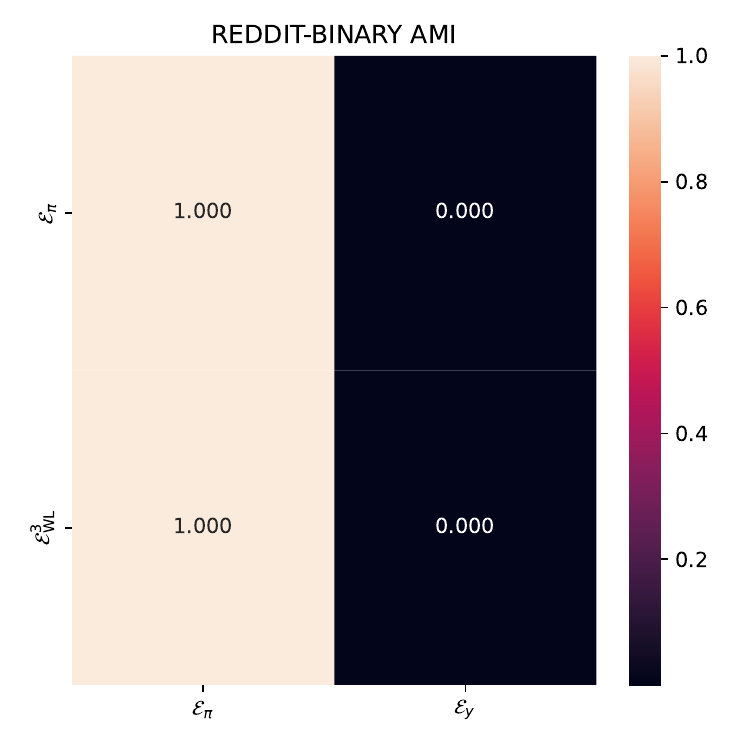}

\caption{Adjusted mutual information (AMI) between different benchmark partitions.}
\label{fig:other-benchmarks-ami}
\end{figure}

\clearpage

\section{GIN experimental settings}
\label{sec:experimental-settings}

Since GIN relies on node features, for IMDB-BINARY and IMDB-MULTI, we set the features as a one-hot encoding of the node degrees, and for REDDIT-BINARY, we set the features as a constant. For the encoder, we use 4 layers, 64 hidden units per layer, and global mean pooling; in each layer, $\epsilon$ is trainable and $h_\Theta$ is a 2-layer, 64-unit MLP with ReLU nonlinearities and BatchNorm. For the decoder, we use 2 layers, 64 hidden units per layer, and dropout with a ratio of 0.5. For training, we employ 100 epochs, a minibatch size of 128, a learning rate (LR) of 0.01, a LR decay factor of 0.5, a LR decay step size of 50, and no weight decay. We train on a random 90\% fold, and achieve comparable test accuracies to \cite{xu2018how}. Error bars are computed over 10 training runs with different random seeds. We train GIN on a single NVIDIA GeForce GTX Titan Xp Graphic Card with 12196MiB of space. All other experiments are run on a single CPU on an internal server.

\clearpage

\section{GIN and 1-WL alignment}
\label{sec:app-alignment}

\begin{figure}[!ht]
    \centering
    \includegraphics[width=0.75\textwidth]{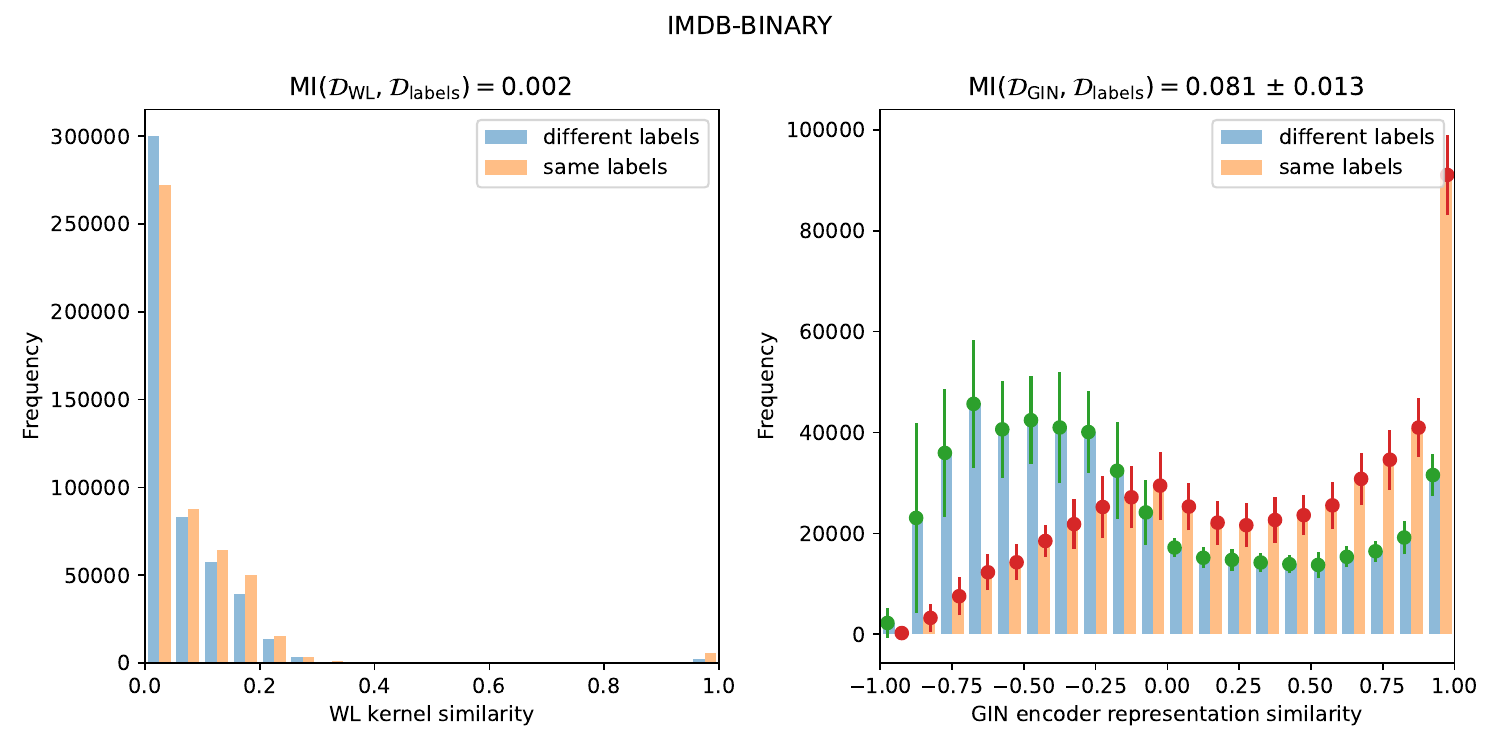}
    \includegraphics[width=0.75\textwidth]{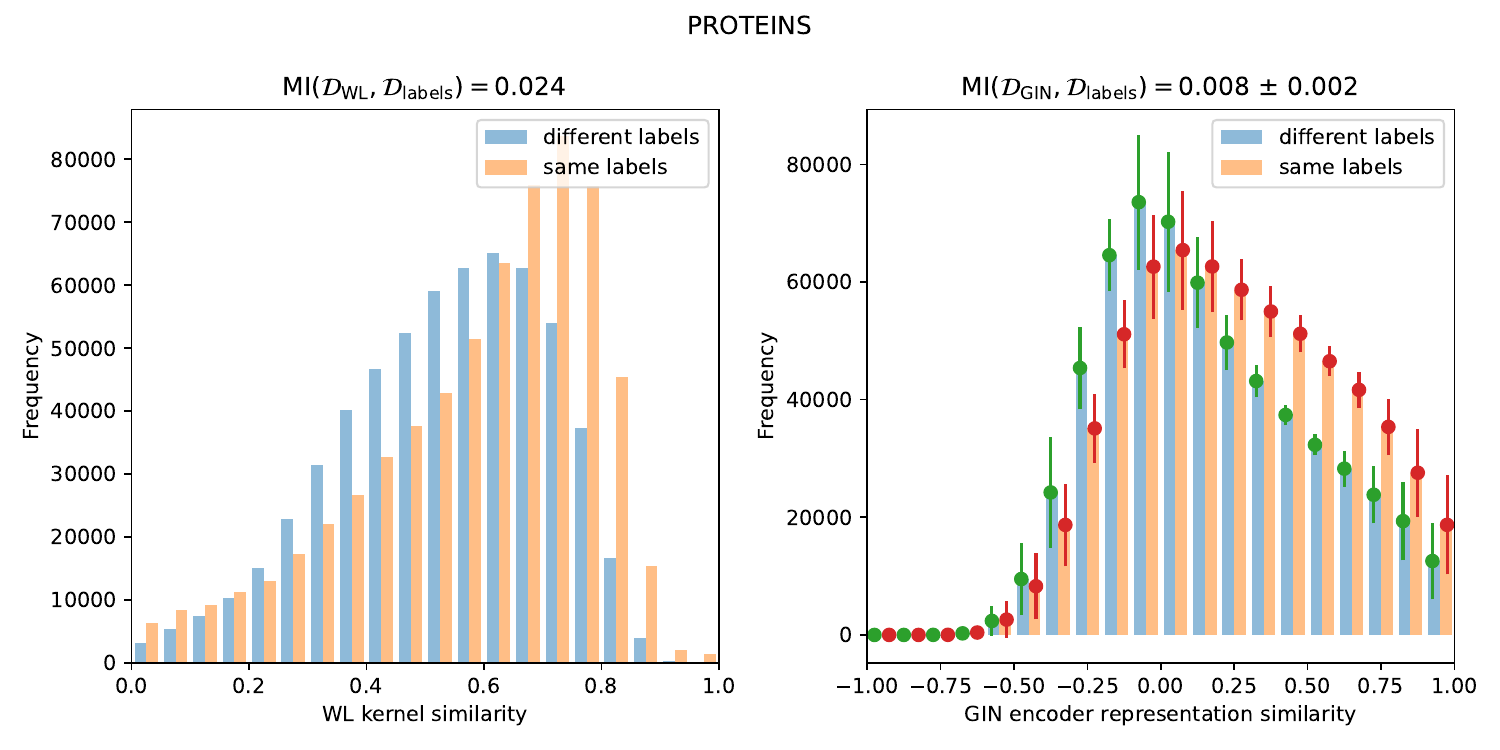}
    \includegraphics[width=0.75\textwidth]{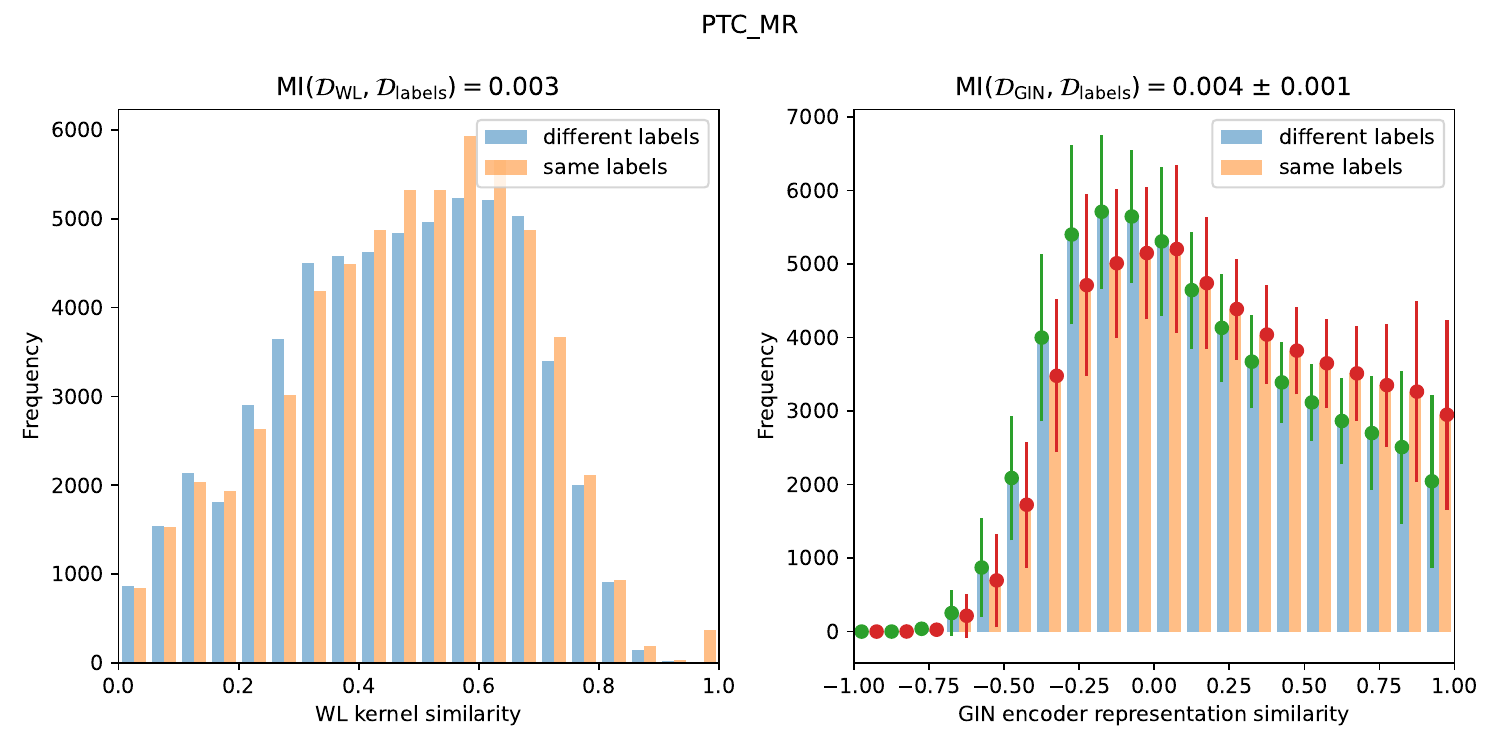}
    \caption{Distributions of WL kernel similarities and GIN encoder representation similarities of graph pairs with different vs. the same labels.}
    \label{fig:other-benchmarks-alignment}
\end{figure}
\FloatBarrier

\clearpage

\section{Limitations and future work}
\label{sec:limitations}

\paragraph{Survey limitations} Our survey sample is skewed towards English-speaking practitioners based in the United States (U.S.) given that our survey was administered in English and the graph ML community is U.S.-centric. (We did not collect demographic data for our sample.) Furthermore, our sample is relatively small; this could be because few practitioners feel qualified to comment on the topic and participants were not compensated. Moreover, our sample is skewed towards researchers (\hyperref[Q6]{Q6}). A possible reason for this is that expressive power largely remains a subject of \textit{theoretical}, rather than \textit{practical}, inquiry. There are likely only tens of academic theory researchers whose primary field of study is GNN expressive power; we believe that our sample reasonably captures the perspectives of these theoreticians. However, our survey results may not reflect the perspectives of applied graph ML practitioners. 

Nevertheless, our results still highlight that even within skewed samples, there exists weak agreement on how expressive power is conceptualized and differing beliefs about $k$-WL. We further note that $n = 18$ is not uncommon in psychology studies, and according to sample size calculators\footnote{e.g., \url{https://www.surveysystem.com/sscalc.htm}}, $n = 18$ is sufficient for a population of 70 with a confidence interval of 20\% and level of 95\%.

We emphasize that our survey results represent a snapshot in time. Practitioners' conceptualizations of expressive power are not static; they evolve over time and are often task-dependent. Furthermore, practitioners' predominant measurement of expressive power via comparison to $k$-WL has likely homogenized their conceptualizations of expressive power \cite{subramonian-etal-2023-tango}.

Although a few respondents had not previously theoretically characterized the expressive power of a GNN and were not familiar with the WL test, their perspectives are still necessary to uncover misalignments between theoretical results and practical concerns; they did not answer the survey questions about the WL test, so the conclusions drawn from the survey remain reliable. We cannot provide the raw responses to the open-ended survey questions for privacy reasons.

\paragraph{Theoretical analysis limitations} While our theoretical analysis primarily relies on the graphs in Figure \ref{fig:example}, they may be easily extended to numerous common classes of graphs (e.g., isoregular graphs \cite{bronstein2020isoregular}) on which 1-WL is known to fail. Furthermore, our theoretical analysis does not cover the potential for GNNs to \textit{approximate} isomorphism testing, substructure counting, etc. \cite{bronstein2020approximate}.

\paragraph{Benchmark auditing limitations} We do not provide AMI for the node-level benchmarks because their size makes computing it intractable, but we expect similar results. We further do not analyze the alignment of GIN node representations with 1-WL node colorings, as the WL subtree kernel is not applicable to node pairs.

\paragraph{Future work} We encourage researchers to run our experiments in \S\ref{sec:benchmarking-expressive-power} on graph ML benchmarks with: (1) even larger graphs (for node-level tasks) and more graphs (for graph-level tasks); (2) different GNN architectures (e.g., to check how the performance of different architectures in the same $k$-WL realm differ on the same task); and (3) varying strengths and versions of $k$-WL. We also encourage more work on the trustworthiness of expressive GNNs. As part of this, further efforts must be invested into the construction of graph benchmarks for evaluating trustworthiness; there currently exist limited ``natively''-graph real-world datasets with sensitive attributes available (often for privacy reasons) \cite{subramonian2022on}, which is evidenced by how none of the social networks in \S\ref{sec:benchmarking-expressive-power} have node features.

\clearpage

\section{Responsible Research Checklist}
\label{sec:responsible-research-checklist}

\begin{enumerate}

\item For all authors...
\begin{enumerate}
  \item Do the main claims made in the abstract and introduction accurately reflect the paper's contributions and scope?
    \answerYes
  \item Did you describe the limitations of your work?
    \answerYes{See Section~\ref{sec:limitations}.}
  \item Did you discuss any potential negative societal impacts of your work?
    \answerYes{See Section~\ref{sec:consequential-validity}.}
  \item Have you read the ethics review guidelines and ensured that your paper conforms to them?
    \answerYes{However, we were unable to compensate survey participants due to internal bureaucratic hurdles.}
\end{enumerate}

\item If you are including theoretical results...
\begin{enumerate}
  \item Did you state the full set of assumptions of all theoretical results?
    \answerYes{}
	\item Did you include complete proofs of all theoretical results?
    \answerYes{}
\end{enumerate}

\item If you ran experiments (e.g. for benchmarks)...
\begin{enumerate}
  \item Did you include the code, data, and instructions needed to reproduce the main experimental results (either in the supplemental material or as a URL)?
    \answerNo{The code will be made available after the review process.}
  \item Did you specify all the training details (e.g., data splits, hyperparameters, how they were chosen)?
    \answerYes{See Section~\ref{sec:experimental-settings}.}
	\item Did you report error bars (e.g., with respect to the random seed after running experiments multiple times)?
    \answerYes{We report error bars over 10 GIN training runs in Figures \ref{fig:mutag-alignment} and \ref{fig:other-benchmarks-alignment}.}
	\item Did you include the total amount of compute and the type of resources used (e.g., type of GPUs, internal cluster, or cloud provider)?
    \answerYes{We mention the compute used in Section~\ref{sec:experimental-settings}.}
\end{enumerate}

\item If you are using existing assets (e.g., code, data, models) or curating/releasing new assets...
\begin{enumerate}
  \item If your work uses existing assets, did you cite the creators?
    \answerYes{We cite the original creators of all the benchmarks we use in Table \ref{tab:equivalence-class-statistics} and Section \ref{sec:consequential-validity}.}
  \item Did you mention the license of the assets?
    \answerYes{We mention the license in Section~\ref{sec:benchmarks-overview}.}
  \item Did you include any new assets either in the supplemental material or as a URL?
    \answerYes{We include the code for our experiments (with dependencies and instructions) in the supplementary material.}
  \item Did you discuss whether and how consent was obtained from people whose data you're using/curating?
    \answerYes{We discuss consent in Section~\ref{sec:additional-details-survey} (survey participants) and Section~\ref{sec:benchmarks-overview} (benchmarks).}
  \item Did you discuss whether the data you are using/curating contains personally identifiable information or offensive content?
    \answerYes{We discuss this in Section~\ref{sec:benchmarks-overview}.}
\end{enumerate}

\item If you used crowdsourcing or conducted research with human subjects...
\begin{enumerate}
  \item Did you include the full text of instructions given to participants and screenshots, if applicable?
    \answerYes{We provide the full text of instructions in Section~\ref{sec:survey_questions}.}
  \item Did you describe any potential participant risks, with links to Institutional Review Board (IRB) approvals, if applicable?
    \answerNA{However, the survey was reviewed by an internal privacy team at a tech company.}
  \item Did you include the estimated hourly wage paid to participants and the total amount spent on participant compensation?
    \answerNo{We were unable to compensate survey participants due to internal bureaucratic hurdles.}
\end{enumerate}

\end{enumerate}

\end{document}